%% file: main.tex
\documentclass[10pt,letterpaper]{article}
\usepackage{algorithm}
\usepackage{algorithmicx}
\usepackage{algpseudocode}
\usepackage{amsmath,amssymb}
\usepackage{placeins}
\usepackage{float}

\algrenewcommand\algorithmicrequire{\textbf{Require:}}
\algrenewcommand\algorithmicensure{\textbf{Ensure:}}
\algrenewcommand\algorithmiccomment[1]{\hfill{\footnotesize$\triangleright$~#1}}
\usepackage{cogsci}
\cogscifinalcopy 

\usepackage{cmap}
\usepackage[T1]{fontenc}
\usepackage[american]{babel}
\usepackage{csquotes}
\usepackage{newtxtext,newtxmath}  
\usepackage{graphicx}
\usepackage{subcaption}
\usepackage[
  backend=biber,
  style=apa,
  natbib=true,
  annotation=false,
]{biblatex}
\usepackage{graphicx}

\usepackage[hidelinks]{hyperref}

\title{From Generative Modeling to Clinical Classification: A GPT-Based Architecture for EHR Notes}

\author[1]{\mbox{Fariba Afrin Irany, Sampson Akwafuo}}

\begin{document}

\maketitle

\begin{abstract}
The increasing availability of unstructured clinical narratives in electronic health records (EHRs) has created new opportunities for automated disease characterization, cohort identification, and clinical decision support. However, effectively modeling long, domain-specific clinical text remains challenging due to limited labeled data, substantial class imbalance, and the high computational cost associated with adapting large pretrained language models to downstream tasks.

In this study, a GPT-based architecture for clinical text classification is presented, which adapts a pretrained decoder-only Transformer to downstream clinical tasks using a selective fine-tuning strategy. Instead of updating all model parameters, the majority of the pretrained GPT-2 backbone is frozen, and training is restricted to the final Transformer block, the final layer normalization, and a lightweight classification head. This design significantly reduces the number of trainable parameters while preserving the representational capacity required for modeling complex clinical language.

This study evaluates the proposed approach on radiology reports from the MIMIC-IV-Note dataset using uncertainty-aware CheXpert-style labels derived directly from report text. Experiments are conducted across multiple problem formulations, including multi-label classification of radiographic findings, binary per-label classification under different uncertainty assumptions, and aggregate disease outcome prediction. Across varying sample sizes, the model demonstrates stable convergence behavior and strong classification performance, particularly in multi-class settings where non-mention and negated findings are prevalent.

Overall, the results suggest that selective fine-tuning of pretrained generative language models offers an efficient and effective pathway for clinical text classification, enabling scalable adaptation to real-world EHR data while substantially reducing computational complexity.

\end{abstract}


\input{Introduction}
\input{Related_Work}
\input{Methods_llm_archi_for_classification}
\input{loading_pretrained_weights}
\input{finetuning}
\input{Trainable_parameter_and_complexity_analysis_finetuning
}
\input{Dataset_and_ground_truth_label_creation}
\input{Experiments_and_Results}

\input{ Evaluation__Measuring_Classification_Accuracy}
\input{Conclusion.tex}

\FloatBarrier
\printbibliography

\end{document}

%% file: Introduction.tex
\section{Introduction}

The widespread adoption of electronic health record (EHR) systems has led to an unprecedented growth in the volume of unstructured clinical text. Clinical narratives such as progress notes, discharge summaries, operative reports, and radiology interpretations capture detailed information about patient history, physician reasoning, and care trajectories that is often unavailable in structured fields. Extracting meaningful information from these free-text documents is essential for a wide range of downstream applications, including disease phenotyping, cohort discovery, outcome prediction, quality assessment, and clinical decision support. However, the unstructured and highly specialized nature of clinical language poses substantial challenges for automated analysis.

Traditional natural language processing approaches in the clinical domain have relied heavily on rule-based systems, domain-specific lexicons, or manually engineered features. While these methods can be effective for narrowly defined tasks, they are often brittle, difficult to scale, and unable to capture long-range dependencies or nuanced contextual meaning. Statistical machine learning methods improved flexibility but still required substantial feature engineering and struggled with the variability and complexity of clinical narratives. These limitations have motivated the adoption of representation learning approaches that can automatically learn contextual features directly from raw text.

Transformer-based language models have emerged as a dominant paradigm for representation learning in natural language processing. By relying on self-attention mechanisms, Transformers are able to model long-range dependencies and contextual interactions between tokens more effectively than recurrent or convolutional architectures. Large language models pretrained on massive text corpora using self-supervised objectives have demonstrated remarkable generalization capabilities, enabling transfer learning across a wide range of downstream tasks. Generative pretrained models, such as GPT-2, learn rich contextual representations by predicting tokens autoregressively, allowing them to capture syntactic structure, semantic relationships, and discourse-level patterns in text.

Despite these advances, directly applying large pretrained language models to clinical text introduces several practical and methodological challenges. Clinical datasets are typically orders of magnitude smaller than the corpora used for pretraining, increasing the risk of overfitting during downstream training. Moreover, the computational cost of fine-tuning all parameters of a large language model can be prohibitive, particularly in academic or clinical environments with limited computational resources. Full fine-tuning also increases memory usage during training due to the need to store gradients and optimizer states for all parameters, further constraining scalability.

An additional challenge arises from the domain mismatch between general-language pretraining data and clinical text. While pretrained models capture broad linguistic knowledge, clinical narratives exhibit unique characteristics, including specialized terminology, abbreviations, telegraphic writing styles, and institution-specific conventions. Effective adaptation therefore requires balancing domain-specific learning with preservation of general-language representations. Updating too many parameters risks catastrophic forgetting of pretrained knowledge, while updating too few parameters may limit the model’s ability to adapt.

To address these challenges, this work explores a selective fine-tuning strategy for adapting GPT-2 to unstructured clinical text classification tasks. Instead of updating the entire model, the majority of pretrained parameters are frozen, and optimization is restricted to a carefully chosen subset of layers. Specifically, only the final Transformer block, the final layer normalization, and a lightweight task-specific classification head are updated during fine-tuning. This design leverages the hierarchical nature of Transformer representations, where lower layers tend to encode general linguistic features and higher layers capture more abstract, task-relevant information.

A key contribution of this work is a detailed and transparent analysis of model parameterization and computational complexity under this selective fine-tuning regime. While selective fine-tuning is often motivated intuitively, its concrete impact on parameter count and training-time complexity is rarely made explicit. In this paper, the number of frozen and trainable parameters in the GPT-2 (small) architecture is quantified, demonstrating that more than ninety-four percent of the model parameters remain fixed during training. A further analysis examines how restricting backpropagation to a single Transformer block substantially reduces training-time complexity and memory requirements while still enabling effective task adaptation.

This work is intentionally written to be accessible to readers who may be new to large language models and Transformer architectures. Rather than assuming familiarity with internal model mechanics, step-by-step explanations are provided to clarify how parameters are distributed across model components, why specific terms arise in the complexity analysis, and how architectural design choices influence computational efficiency. By making these details explicit, selective fine-tuning strategies are demystified, and practical guidance is provided for researchers seeking to apply large language models in data-limited or resource-constrained settings.

Overall, this study demonstrates that efficient adaptation of generative pretrained language models to clinical text is possible without full-model fine-tuning. By combining selective parameter optimization with a clear understanding of architectural complexity, the proposed approach offers a practical and computationally efficient pathway for deploying large language models in clinical natural language processing applications.

This work makes the following contributions:
\begin{itemize}
    \item A parameter-efficient selective fine-tuning strategy for GPT-based clinical text classification that updates only a small subset of model parameters.
    \item A systematic analysis of training dynamics and computational trade-offs associated with selective fine-tuning.
    \item An empirical evaluation on large-scale MIMIC-IV radiology reports demonstrating competitive performance under constrained computational resources.
\end{itemize}

%% file: Related_Work.tex
\section{Literature Review}

This literature review situates the proposed selective fine-tuning framework within four intersecting research areas: (i) Transformer-based language models and large-scale pretraining, (ii) transfer learning and domain adaptation for NLP, (iii) parameter-efficient fine-tuning and computationally efficient adaptation, and (iv) clinical NLP and modeling unstructured EHR narratives.

\subsection{Transformer Architectures and Large-Scale Language Model Pretraining}

The Transformer architecture introduced self-attention as the core mechanism for sequence modeling, enabling efficient parallel computation and strong performance on machine translation and beyond \cite{vaswani2017attention}. Subsequent work clarified optimization and architectural details relevant to deep Transformer training, including the role of normalization and residual design \cite{ba2016layernorm,xiong2020layernorm}. The broad adoption of Transformers led to the development of large pretrained language models trained with self-supervised objectives, which learn reusable linguistic representations. 

Early milestones in language model pretraining include ELMo \cite{peters2018elmo} and ULMFiT \cite{howard2018universal}, which established that pretrained representations can dramatically improve downstream task performance. GPT-style models popularized decoder-only autoregressive pretraining at scale \cite{radford2018improving,radford2019language,brown2020language}, while BERT introduced deep bidirectional masked language modeling and demonstrated strong transfer to many downstream tasks \cite{devlin2019bert}. Several influential architectural or training enhancements further improved scale and stability, including RoBERTa \cite{liu2019roberta}, ALBERT \cite{lan2019albert}, XLNet \cite{yang2019xlnet}, T5 \cite{raffel2020t5}, and ELECTRA \cite{clark2020electra}. Improvements in Transformer efficiency and long-context handling have been explored through sparse or reformulated attention mechanisms such as Transformer-XL \cite{dai2019transformerxl}, Longformer \cite{beltagy2020longformer}, BigBird \cite{zaheer2020bigbird}, Reformer \cite{kitaev2020reformer}, Linformer \cite{wang2020linformer}, and Performer \cite{choromanski2021performer}. Collectively, these works provide the foundation for applying pretrained Transformer representations to specialized domains, including clinical narratives, where long-range dependencies and context sensitivity are essential.

\subsection{Transfer Learning, Domain Adaptation, and Continued Pretraining}

A central motivation for pretrained language models is their ability to transfer linguistic knowledge to data-scarce tasks. General transfer learning principles and empirical studies show that features learned in large models can be reused across tasks, although transferability depends on layer depth and domain similarity \cite{yosinski2014transfer,peters2019tune}. In NLP, domain adaptation can be achieved through continued pretraining on domain-specific corpora prior to supervised fine-tuning. Gururangan et al.\ demonstrated that continued pretraining (domain-adaptive pretraining and task-adaptive pretraining) provides systematic gains in specialized domains \cite{gururangan2020don}. Similar observations appear in biomedical and clinical language modeling, where domain-specific corpora and terminologies differ substantially from general web text. Prior large-scale studies in computational epidemiology and network science have demonstrated the importance of scalable, data-efficient modeling frameworks for learning from heterogeneous, high-dimensional data under resource and labeling constraints, providing methodological motivation for transfer learning and selective adaptation strategies in applied domains \cite{irany2024large}.

Within biomedical NLP, BioBERT \cite{lee2020biobert}, SciBERT \cite{beltagy2019scibert}, PubMedBERT \cite{gu2021pubmedbert}, and BioMegatron \cite{shin2020biomegatron} represent prominent encoder-based domain-adaptive pretraining efforts. In the clinical domain, ClinicalBERT \cite{huang2019clinicalbert} and related work show that adapting pretrained models to clinical notes can improve performance on clinical prediction and classification. For broader medical language understanding, large-scale instruction tuning and multitask objectives have also been explored in the medical setting \cite{singhal2023medpalm,jiang2023healthsystemreview}.

Domain mismatch, limited labels, and privacy constraints complicate clinical NLP, making efficient adaptation strategies particularly important. Consequently, approaches that reduce the number of trainable parameters while maintaining task performance have become central to practical deployment in clinical contexts. Prior studies addressing bias and data suppression in large-scale health datasets further motivate transfer learning and selective adaptation strategies that can operate effectively under incomplete or weak supervision \cite{irany2024bias}. 

\subsection{Parameter-Efficient Fine-Tuning and Computationally Efficient Adaptation}

Full fine-tuning updates all parameters of a pretrained model and typically yields strong downstream performance, but it can be computationally expensive and memory intensive for large models. This motivates parameter-efficient fine-tuning (PEFT) methods that update only a small subset of parameters while keeping most pretrained weights frozen. Early and conceptually related ideas include updating only parts of a network, fine-tuning only higher layers, or using lightweight task-specific heads \cite{howard2018universal,peters2019tune}. More recent PEFT methods introduce structured low-rank or modular adaptations that are explicitly designed for large Transformers. Related work on efficient representation learning in multilayer and high-dimensional data settings further demonstrates that constraining model adaptation to structurally informative components can preserve performance while substantially reducing computational complexity \cite{santra2023efficient}.

Adapter layers insert small bottleneck modules into each Transformer block, enabling efficient task adaptation with minimal additional parameters \cite{houlsby2019adapters,pfeiffer2020adapterfusion}. LoRA learns low-rank updates to attention and projection matrices without modifying the original weights, reducing memory usage and enabling efficient optimization \cite{hu2022lora}. Related low-rank or prompt-like approaches include prefix tuning \cite{li2021prefixtuning}, prompt tuning \cite{lester2021prompttuning}, and p-tuning variants \cite{liu2021ptuning}. BitFit shows that updating only bias terms can yield surprisingly strong performance in certain settings \cite{zaken2022bitfit}. IA$^3$ and similar methods reparameterize intermediate activations or gates rather than updating full weight matrices \cite{liu2022ia3}. Comprehensive surveys further systematize the PEFT landscape and clarify tradeoffs among methods \cite{ding2023peftsurvey}.

In parallel, optimization and regularization techniques have been developed to stabilize fine-tuning and improve generalization. AdamW decouples weight decay from gradient updates and is widely used in Transformer training \cite{loshchilov2019adamw}. Dropout remains a core regularization method to mitigate overfitting \cite{srivastava2014dropout}. Weight tying between input embeddings and output projections is known to improve language modeling efficiency and can reduce parameter count \cite{press2017using}. These training principles and parameter-sharing strategies motivate careful parameter accounting and complexity analysis when proposing selective fine-tuning schemes.

The selective fine-tuning approach presented in this paper is aligned with this broader body of work. Rather than introducing new modules across all layers, training is restricted to the topmost GPT-2 block, along with the final normalization layer and the classification head. This design yields a clear reduction in backpropagation cost and trainable parameter count while retaining the representational capacity learned during pretraining.

\subsection{Clinical NLP and Modeling Unstructured EHR Narratives}

Clinical NLP aims to extract structured knowledge from unstructured clinical narratives and enable predictive modeling and decision support. Foundational resources and shared tasks such as i2b2 have accelerated progress by providing benchmark datasets for clinical concept extraction, relation extraction, and phenotyping \cite{uzuner20112010i2b2}. More recent shared tasks and datasets have expanded evaluation to clinical temporal reasoning and other complex phenomena \cite{mullenbach2018icd,huang2019clinicalbert}. Classic clinical corpora such as MIMIC have enabled broad research on clinical prediction and text-based risk modeling \cite{johnson2016mimic}. Clinical text classification is often operationalized through coding tasks such as ICD assignment, where label spaces are large and notes are long \cite{mullenbach2018icd,li2019caml}.

Transformer models have been widely applied to clinical NLP, often via encoder-based architectures due to their strong discriminative capabilities \cite{devlin2019bert,lee2020biobert,gu2021pubmedbert}. However, decoder-only models are increasingly relevant for clinical note understanding because they can model long-form generation, reasoning, and context completion, and can be adapted for classification through lightweight heads. Empirical studies demonstrate that domain-adapted language models improve downstream clinical tasks, but also highlight the tension between performance, compute, and data constraints in clinical settings \cite{huang2019clinicalbert,gu2021pubmedbert}. Recent work on clinical and biomedical LLMs reflects growing interest in large-scale models for medical QA, reasoning, and summarization, while emphasizing safety and evaluation challenges \cite{singhal2023medpalm,jiang2023healthsystemreview}.

Overall, the literature supports three key conclusions relevant to this paper: (i) pretrained Transformers provide strong contextual representations for long and specialized text \cite{vaswani2017attention,radford2019language,devlin2019bert}, (ii) domain adaptation and continued pretraining improve performance in biomedical and clinical NLP \cite{gururangan2020don,lee2020biobert,gu2021pubmedbert,huang2019clinicalbert}, and (iii) parameter-efficient or selective fine-tuning is often necessary for practical deployment under limited compute and limited labeled data \cite{houlsby2019adapters,hu2022lora,li2021prefixtuning,lester2021prompttuning}. These findings motivate the selective fine-tuning design and the explicit parameter and time-complexity analysis presented in this work.

Beyond full end-to-end fine-tuning, prior studies have shown that selectively updating subsets of model parameters can yield competitive performance while substantially reducing computational cost. Early analyses of transfer learning in deep neural networks indicate that higher layers tend to encode task-specific features, whereas lower layers capture more general linguistic representations \cite{yosinski2014transfer,peters2019tune}. Motivated by these observations, several works have explored partial fine-tuning strategies for Transformer-based models, including freezing lower layers and updating only top layers or task-specific heads \cite{howard2018universal}. More recent parameter-efficient fine-tuning approaches further formalize this idea by introducing lightweight adaptation mechanisms such as adapter modules, low-rank updates, or prompt-based reparameterization, enabling efficient downstream learning without modifying the full parameter set \cite{houlsby2019adapters,hu2022lora,li2021prefixtuning}. These approaches are particularly relevant in domains such as clinical natural language processing, where labeled data and computational resources are often limited, and they motivate the selective fine-tuning strategy adopted in this work.

%% file: Methods_llm_archi_for_classification.tex
\section{LLM Architecture to Support Classification}
\subsection{Overall GPT Model Architecture}

\begin{figure*}%
    \centering
   \includegraphics[height=8cm, width=12cm]{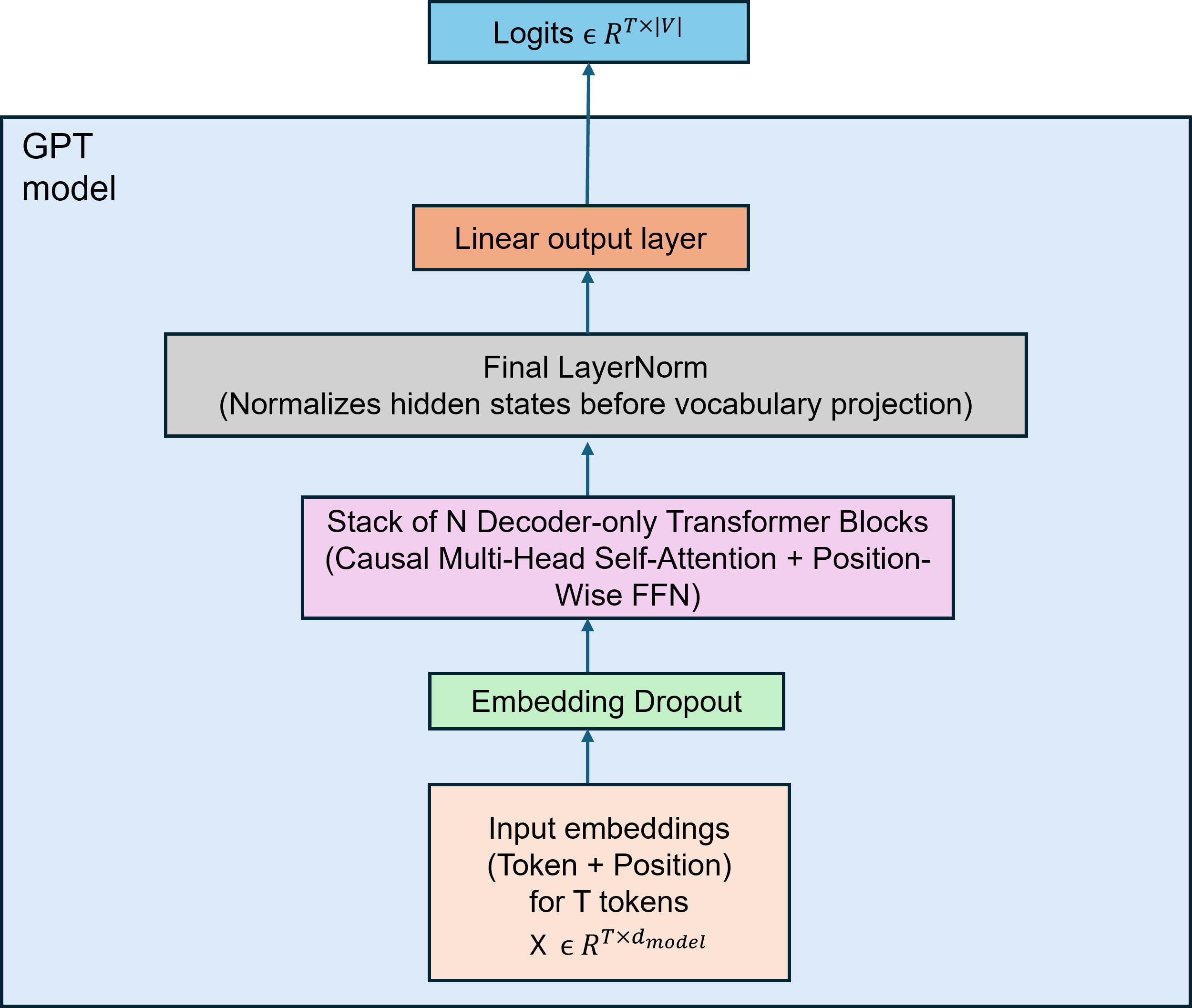} %
    \caption{Overall architecture of the GPT model.}%
   \label{gpt model archi}
\end{figure*}

This section presents the overall GPT architecture designed for the classification of unstructured clinical data, illustrating how the model components are integrated into a unified end-to-end system. The architecture is based on a decoder-only Transformer framework adapted to process and represent free-text clinical narratives in an autoregressive manner. Figure~\ref{gpt model archi} provides a high-level overview of the complete model pipeline used for clinical text classification. In the following subsections, each major architectural component, including tokenization, embedding layers, masked self-attention blocks, feed-forward networks, and the classification head, is described in detail.

\textbf{Model Overview:}
The GPT model is a unidirectional (decoder-only) Transformer designed for autoregressive language modeling. Given an input sequence of length T, the model processes tokens sequentially while enforcing causal dependencies, ensuring that each token representation attends only to itself and preceding tokens in the sequence \cite{radford2018improving, radford2019language, vaswani2017attention}.

Formally, an input token sequence is first mapped to a sequence of continuous vector representations by summing token embeddings and positional embeddings. This results in an embedding matrix
$X \in \mathbb{R}^{T \times d_{\text{model}}}$,

where $d_{model}$ denotes the model’s hidden dimensionality.

\textbf{Embedding and Regularization:}
The combined token and positional embeddings are passed through an embedding-level dropout layer, which acts as an early regularization mechanism to reduce overfitting and improve generalization during training. This dropout is applied uniformly across all token positions in the sequence.

\textbf{Stack of Decoder-only Transformer Blocks:}

The regularized embeddings are then processed by a stack of N identical decoder-only Transformer blocks. Each block consists of two primary sublayers:
(i) causal multi-head self-attention, and
(ii) a position-wise feed-forward network (FFN).

Causal masking within the self-attention mechanism enforces the autoregressive constraint, preventing information flow from future tokens to past tokens. Residual connections and layer normalization within each block stabilize training and enable effective gradient propagation across deep architectures. The repeated stacking of these blocks allows the model to build progressively richer contextual representations over long sequences \cite{vaswani2017attention}.

\textbf{Final Normalization and Output Projection:}
After passing through the full stack of Transformer blocks, the hidden representations are normalized using a final layer normalization layer. This normalization step stabilizes the hidden states before projection into the vocabulary space and has been shown to improve training dynamics in deep Transformer models \cite{ba2016layernorm, xu2020understanding}.

The normalized representations are then fed into a linear output layer that projects each token’s hidden state into a vector of unnormalized scores (logits) over the vocabulary:


$\text{logits} \in \mathbb{R}^{T \times \lvert V \rvert}$, where $\lvert V \rvert$ denotes the vocabulary size. Each row of the logits matrix corresponds to the model’s prediction for the next token at a given position in the input sequence.

\textbf{Autoregressive Prediction:}
During training, these logits are typically converted into probability distributions using a softmax function and optimized using a cross-entropy loss. During inference, the model generates text autoregressively by sampling or selecting the most probable token at each step and appending it to the input sequence, thereby extending the sequence one token at a time \cite{radford2019language, brown2020language}.

\subsection{Input Pre-processing}

\begin{figure*}[!htbp]%
    \centering
   \includegraphics[height=6cm, width=8 cm]{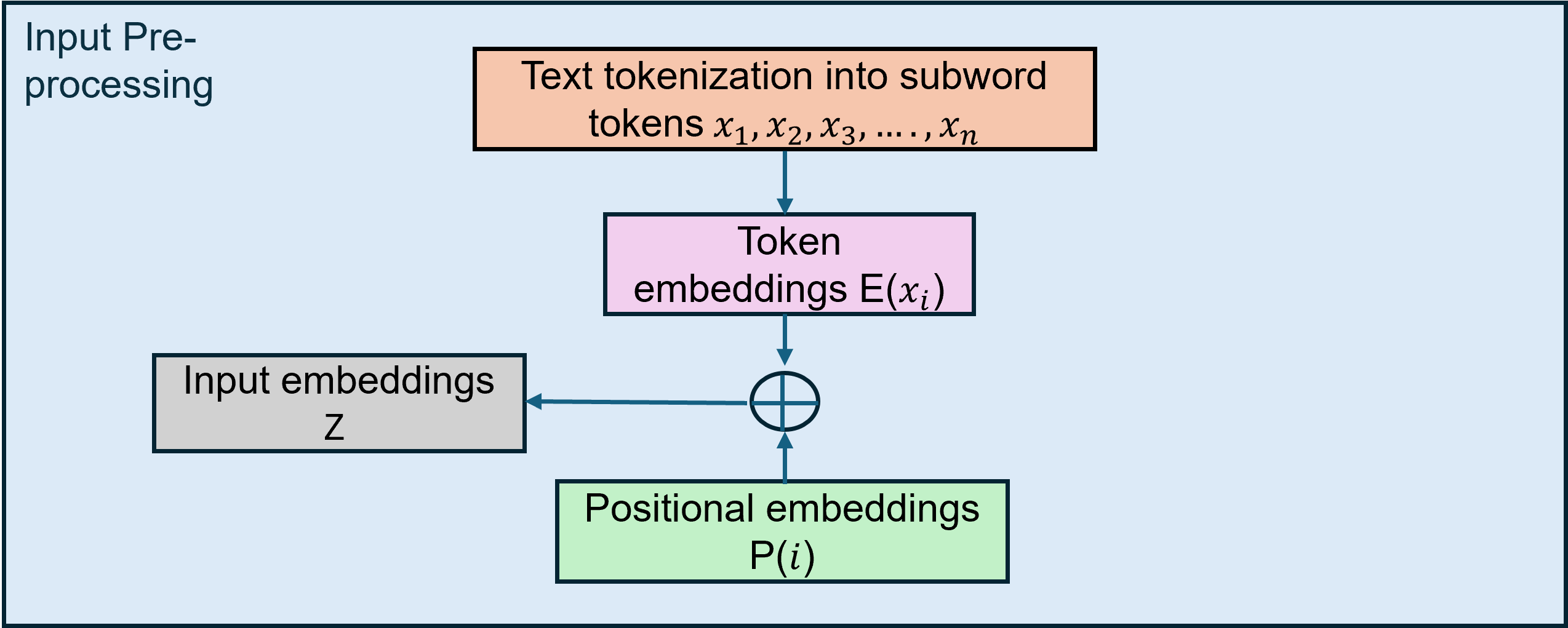} %
    \caption{Input Pre-Processing and Embedding Construction in Transformer-Based Language Models.}%
   \label{Input Pre-Processing}
\end{figure*}

The input pre-processing pipeline in Transformer-based language models shown in figure~\ref{Input Pre-Processing} begins with subword tokenization using byte pair encoding algorithm \cite{gage1994bpe, sennrich2016bpe}, wherein raw text is segmented into a sequence of discrete tokens ${x_1, x_2, x_3, ...., x_n}$ that serve as the fundamental units for embedding and downstream representation learning. Each token is then mapped to a continuous vector representation through a learned token embedding function ${E(x_i)}$. To encode word order information, a positional embedding P(i), corresponding to the token’s position within the sequence, is added to the token embedding. The resulting input embedding is computed as
${Z_i=E(x_i)+P(i)}$ where the addition is performed element-wise. These input embeddings serve as the final representations passed to subsequent Transformer layers for contextual modeling.

\subsection{Multi-Head Self-Attention Mechanism}

The multi-head self-attention mechanism, including scaled dot-product attention, parallel attention heads, head concatenation, and the final output projection, was originally introduced by  \cite{vaswani2017attention} and forms the foundation of modern Transformer-based architectures.
\begin{figure*}[!htbp]%
    \centering
   \includegraphics[height=12cm, width=14 cm]{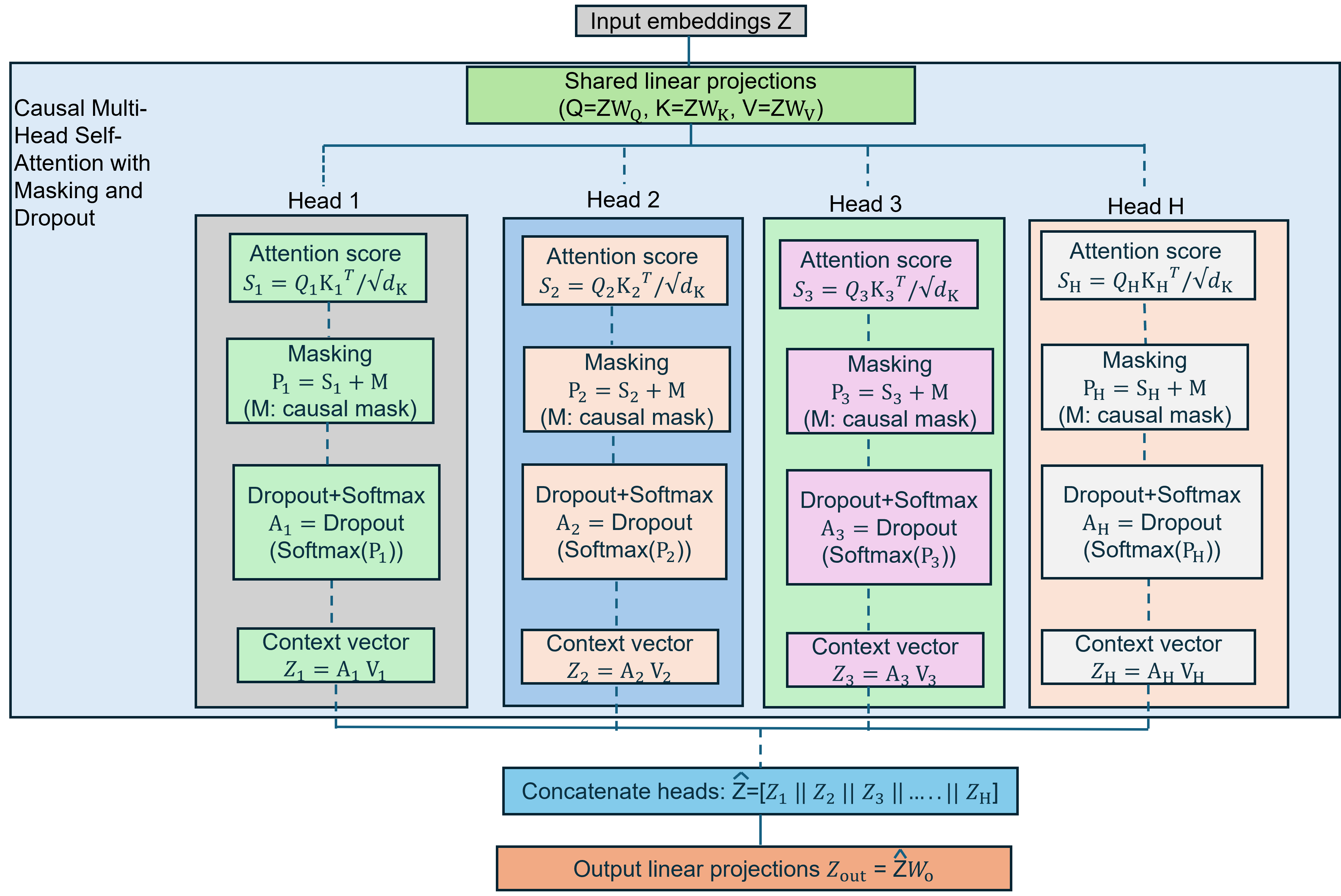} %
    \caption{Computation Flow of Causal Multi-Head Self-Attention with Scaled Dot-Product Attention, Masking, and Dropout.}%
   \label{multi-head attention}
\end{figure*}

Figure~\ref{multi-head attention} illustrates the causal multi-head self-attention mechanism, a fundamental building block of Transformer-based language models. The figure depicts how a shared input embedding matrix is first mapped into query, key, and value representations through shared linear projections and then partitioned across multiple attention heads. Each head independently computes scaled dot-product attention, applies causal masking to prevent access to future tokens, and incorporates dropout for regularization. The resulting head-specific context vectors are subsequently concatenated and passed through an output linear projection to produce the final multi-head attention representation.

\textbf{1. Linear Projections into Queries, Keys, and Values}:
Within the multi-head attention block, the input embedding matrix Z is first transformed using three shared linear projections to obtain the query, key, and value representations, following the standard Transformer formulation introduced by Vaswani et al.~\cite{vaswani2017attention}:

Query projection: $Q=ZW_{Q}$,
Key projection: $K=ZW_{K}$,
Value projection: $V=ZW_{V}$,
where $W_Q, W_K, W_V$ are learnable weight matrices shared across all attention heads.
The resulting query, key, and value matrices are then reshaped and partitioned along the feature dimension into 
H parallel attention heads, such that each head $i \in \{1, 2, \dots, H\}$ operates on a subspace of dimension $d_K=d_{model} \div H$. In practice, this partitioning is implemented via tensor reshaping rather than explicit slicing. This yields head-specific representations $Q_i, K_i, V_i$ without requiring separate projection matrices per head, as commonly implemented in modern deep learning frameworks~\cite{vaswani2017attention,paszke2019pytorch}.

This shared-projection design improves computational efficiency while still allowing each head to attend to different representation subspaces. Figure~\ref{multi-head attention} illustrates this process by showing a single shared projection block feeding multiple attention heads in parallel via dashed connections.

\textbf{2. Scaled Dot-Product Attention Within Each Head}:
For each head, the queries, keys, and values are used to compute attention outputs through a scaled dot-product attention operation originally proposed in~\cite{vaswani2017attention}. This involves the following steps:

\textbf{Attention score computation}: The raw attention score matrix is obtained by taking the dot product between the queries and the transposed keys. 
$S_i=Q_i{K_i}^T$.
This yields a score for how much each position should attend to every other position:

\textbf{Scaling}: The raw scores are scaled by the inverse square root of the key dimension, which stabilizes gradients during training:

$\tilde{S}_i = \frac{S_i}{\sqrt{d_K}}$,

where $d_K$ denotes the dimensionality of the key vectors.
After scaling and causal masking, a softmax function is applied along the last axis to normalize the attention scores into a probability distribution over tokens.
After scaling and causal masking, a softmax function is applied along the last axis to normalize the attention scores into a probability distribution over tokens:
\[
A_i = \mathrm{Softmax}(P_i),
\]
where each row of $A_i$ sums to one and represents the relative importance assigned to all valid (non-masked) tokens for a given query position \cite{vaswani2017attention}.



\textbf{3. Causal Masking and Dropout}:
To enforce the autoregressive property required for causal language modeling, a causal (masked) self-attention mechanism is employed, following the design introduced in generative pre-trained Transformer models \cite{radford2018improving,radford2019language}to the scaled attention scores before normalization. This mask assigns large negative values to positions corresponding to future tokens, preventing each position from attending to subsequent positions in the sequence:
$P_i = \tilde{S}_i + M$.
Here, $M$ denotes a causal upper-triangular mask with $-\infty$ entries
assigned to future positions, ensuring autoregressive behavior.
Following masking, a softmax function converts the masked scores into attention weights, and dropout is applied to improve regularization:
$A_i=Dropout(Softmax(P_i)).$

\textbf{4. Context Vector Computation}:
The normalized attention weights are then used to compute the context vector for each head as a weighted sum of the value vectors:

$Z_i=A_iV_i.$

Each head produces its own context tensor 
$Z_i$, capturing information from the input sequence according to distinct attention patterns. Figure~\ref{multi-head attention} depicts these operations beneath each head, explicitly illustrating the attention score computation, masking, softmax, dropout, and context vector formation.

\textbf{5. Parallelism across Multiple Heads}:
Unlike single-head attention, the above sequence of operations is executed in parallel across $H$ heads. While all heads share the same input embeddings, each head attends to a different subspace of the shared projected representations, enabling the model to capture diverse syntactic, semantic, and positional relationships simultaneously~\cite{vaswani2017attention}.

The figure~\ref{multi-head attention} emphasizes this parallel structure by visually separating the computations for each head while maintaining a shared input.

Finally, the concatenated representation is passed through an output linear projection $W_o$ to mix information across heads and prepare the output for subsequent Transformer layers. In the figure~\ref{multi-head attention}, this is shown as the bottom block labeled Output Linear Projections.

\textbf{6. Concatenation and Output Projection}:
After all heads compute their respective context vectors 
${Z_1, Z_2, Z_3, ..., Z_H}$, these vectors are concatenated along the feature dimension:
$\hat{Z} = [Z_1 \,\|\, Z_2 \,\|\, \cdots \,\|\, Z_H].$

The concatenated tensor is then passed through a final output linear projection:
$Z_{out}=\hat{Z}W_o$, where $W_o$ is a learnable weight matrix that mixes information across heads and produces the output representation used by subsequent Transformer layers. This operation is shown at the bottom of Figure~\ref{multi-head attention} as the Output linear projection block~\cite{vaswani2017attention,paszke2019pytorch}.

This formulation closely follows standard implementations of multi-head self-attention used in modern deep learning frameworks, including PyTorch’s nn.MultiheadAttention, where query, key, and value projections are typically computed using a fused linear layer and the outputs of all heads are combined through a final projection \cite{paszke2019pytorch}. The inclusion of causal masking and dropout aligns the architecture with commonly used Transformer-based language models.

\subsection{Transformer}

\begin{figure*}%
    \centering
   \includegraphics[height=12cm, width=10 cm]{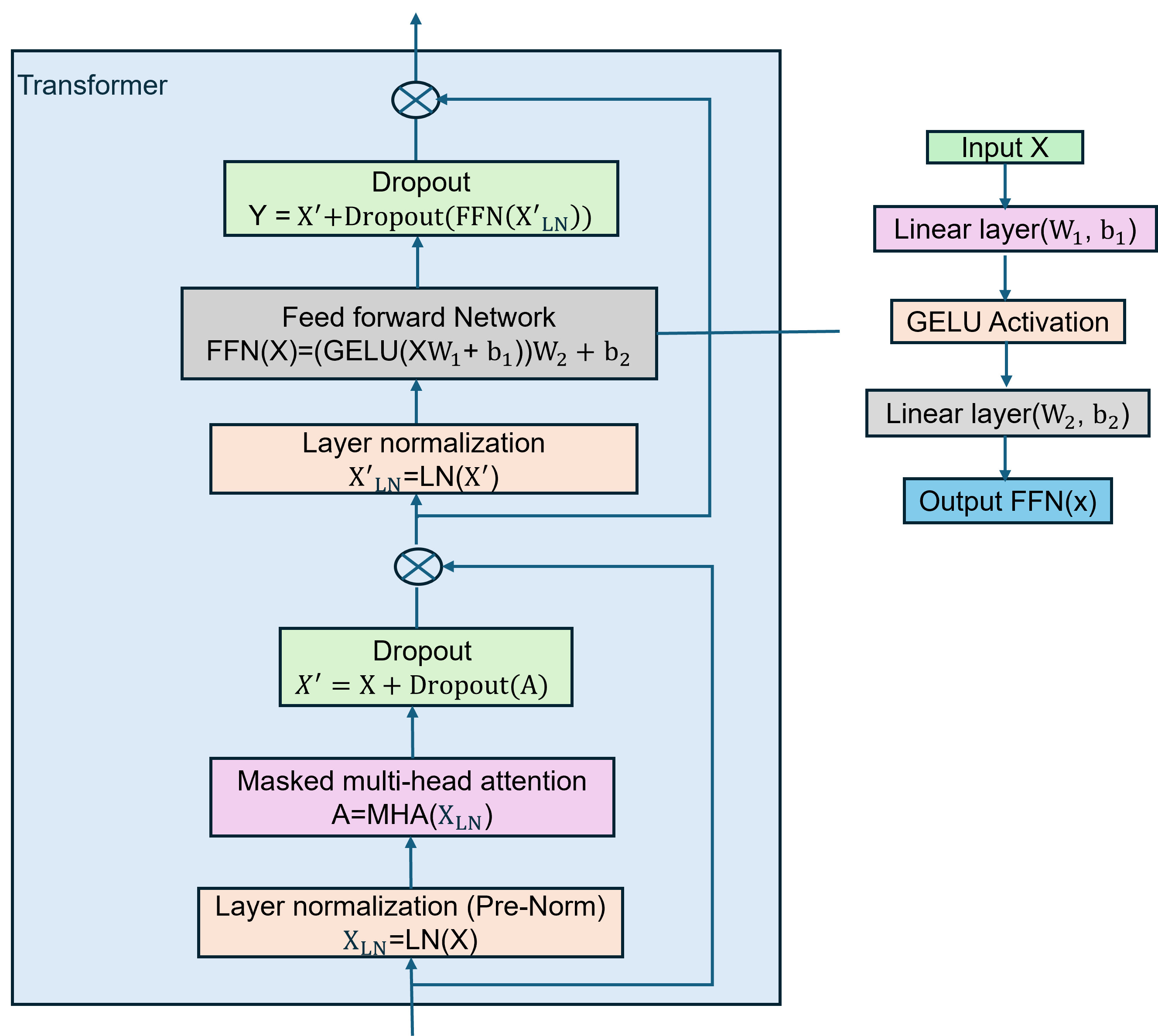} %
    \caption{Decoder-only Transformer block with pre-normalization, masked multi-head self-attention, residual connections, dropout, and a position-wise feed-forward network.}%
   \label{transformer}
\end{figure*}

In decoder-only Transformer architectures, such as GPT-2, causal masking is a fundamental design principle that enables autoregressive sequence modeling \cite{radford2019language}. In these models, each Transformer block processes an input sequence through a series of structured operations designed to capture contextual dependencies while preserving the left-to-right generation constraint. As illustrated in Figure~\ref{transformer}, each block consists of two main sublayers: a masked multi-head self-attention sublayer and a position-wise feed-forward sublayer. Each sublayer is wrapped with layer normalization, dropout, and residual connections.

\textbf{Layer Normalization (Pre-Norm):}
At first, layer normalization is applied to the input
$X_{LN}=LN(X)$.
This pre-normalization stabilizes training by mitigating internal covariate shift, a phenomenon in deep neural networks in which the distribution of inputs to hidden layers changes as preceding layers are updated during training. By reducing such distributional drift, pre-normalization accelerates convergence and improves optimization stability by alleviating vanishing and exploding gradient issues. As a result, this design is widely adopted in modern large language models \cite{xu2020understanding}. The core principle of pre-normalization is to normalize neural activations to have zero mean and unit variance prior to subsequent transformations.

\textbf{Masked Multi-Head Self-Attention:}
The normalized representations are passed to a masked multi-head self-attention (MHA) sublayer:
$A=MHA(X_{LN})$.

Causal masking is applied within the attention mechanism to prevent each token from attending to future positions, ensuring autoregressive behavior consistent with decoder-style language models \cite{vaswani2017attention,radford2019language}.

\textbf{Dropout and Residual Connection (Attention Sublayer):}
Dropout is applied to the output of the masked multi-head self-attention sublayer to mitigate overfitting by randomly deactivating a subset of attention activations during training. The regularized attention output is then combined with the original input through a residual (skip) connection:
$X'=X+Dropout(A)$.
This residual connection enables the direct propagation of information from earlier layers to later layers, facilitating more effective gradient flow during backpropagation. By providing an identity mapping path, residual connections alleviate vanishing gradient issues, accelerate convergence, and improve optimization stability in deep neural networks \cite{he2016deep}. In the context of Transformer architectures, this mechanism allows each layer to refine representations incrementally while preserving previously learned contextual information.

\textbf{Layer Normalization Before Feed-Forward Network:}
The intermediate representation is again normalized prior to the feed-forward network:
$X'_{LN}=LN(X')$.

\textbf{Position-Wise Feed-Forward Network:}
A position-wise feed-forward network (FFN) is applied independently to each token representation produced by the self-attention sublayer. This component is implemented as a two-layer fully connected neural network with a single non-linear activation function applied element-wise between the layers. The first linear transformation projects the input representation into a higher-dimensional feature space, enabling the model to capture richer and more expressive patterns through an expanded set of parameters. This expansion layer increases the model’s capacity to learn complex relationships among features that may not be adequately represented in the original embedding space.

The second linear transformation subsequently projects the activated representations back to the original model dimension. This contraction step ensures that the input and output dimensions of the Transformer block remain consistent, which is essential for scalability, residual connections, and efficient stacking of multiple layers. By combining dimensional expansion, non-linear transformation, and dimensional reduction, the FFN significantly enhances the representational power of the Transformer while maintaining architectural compatibility across layers \cite{vaswani2017attention}. In modern Transformer-based language models, the Gaussian Error Linear Unit (GELU) is commonly used as the activation function due to its smooth non-linear behavior and improved empirical performance \cite{hendrycks2016gelu,devlin2019bert}. For an input token representation $x \in \mathbb{R}^{d_{\text{model}}}$, the FFN is defined as:

$FFN(X)=\bigl(\mathrm{GELU}(XW_1 + b_1)\bigr) W_2 + b_2$,
where $W_1$ and $W_2$ are learnable weight matrices, and $b_1$ and $b_2$ are learnable bias vectors. The first linear transformation projects the input from the model dimension $d_{\text{model}}$ to a higher-dimensional hidden space, often denoted as $d_{\text{ff}}$, enabling richer feature extraction. The GELU activation function introduces non-linearity by modulating inputs according to their magnitude and sign, allowing the network to selectively pass or suppress information in a smooth and probabilistic manner.
The GELU activation is formally defined as \cite{hendrycks2016gelu}:

\[
\mathrm{GELU}(x) = x \, \Phi(x),
\]

where $\Phi(x)$ denotes the cumulative distribution function of the standard normal distribution. In practice, an efficient approximation is often used:

\[
\mathrm{GELU}(x) \approx 0.5\,x \left( 1 + \tanh\!\left[ \sqrt{\frac{2}{\pi}} \left( x + 0.044715\,x^3 \right) \right] \right).
\]
Because the FFN is applied position-wise, the same parameters $(W_1, b_1, W_2, b_2)$ are shared across all token positions, and each token is transformed independently. As a result, cross-token interactions are handled by the self-attention mechanism, while the FFN enhances the expressive power of individual token representations through non-linear feature transformation \cite{vaswani2017attention}.

\textbf{Dropout and Residual Connection (Feed-Forward Sublayer):}
Dropout is applied to the FFN output, followed by another residual connection:

$Y=X'+Dropout(FFN(X'_{LN}))$.

The output Y serves as the final representation produced by the Transformer block.

\textbf{Stacking of Transformer Blocks:}
Multiple Transformer blocks are stacked sequentially, enabling the model to learn increasingly abstract and context-rich representations \cite{radford2019language,brown2020language}. The output of the final block is passed to downstream components such as pooling layers or classification heads.

\subsection{Classification Head Using Last-Token Representation}

In decoder-only GPT architectures, sequence-level representations for classification can be derived directly from the hidden state of the final token. Due to the causal self-attention mechanism, the representation of the last token implicitly encodes contextual information from all preceding tokens in the input sequence. This property makes the last-token hidden state a natural and computationally efficient choice for unstructured clinical text classification.

Let $H \in \mathbb{R}^{T \times d_{\text{model}}}$ denote the output of the final Transformer block for an input sequence of length $T$, where $d_{\text{model}}$ is the hidden dimensionality. The sequence representation is obtained by selecting the hidden state corresponding to the final token:
\[
h_{\text{seq}} = h_T \in \mathbb{R}^{d_{\text{model}}}.
\]

This last-token representation serves as a compact summary of the entire clinical narrative, leveraging the autoregressive dependency structure of the model. Unlike encoder-based architectures that rely on a dedicated classification token, decoder-only models naturally accumulate contextual information into the final token representation. A linear classification head is applied to $h_{\text{seq}}$ to produce prediction logits:
\[
\mathbf{z} = W h_T + b,
\]
where $W \in \mathbb{R}^{C \times d_{\text{model}}}$ and $b \in \mathbb{R}^{C}$ are learnable parameters, and $C$ denotes the number of output labels.

\subsubsection{Binary Classification Head (Single-Label):}

For binary clinical prediction tasks (e.g., presence vs.\ absence of any abnormality), the model outputs a single logit ($C=1$):
\[
z = w^\top h_T + b,
\]
where $w \in \mathbb{R}^{d_{\text{model}}}$ and $b \in \mathbb{R}$.

The predicted probability of the positive class is obtained using the sigmoid function:
\[
P(y=1 \mid x) = \sigma(z) = \frac{1}{1+\exp(-z)}.
\]

During supervised training, the model is optimized using binary cross-entropy loss:
\[
\mathcal{L}_{\text{bin}} = -\left[ y \log \sigma(z) + (1-y)\log(1-\sigma(z)) \right],
\]
where $y \in \{0,1\}$ is the ground-truth label. Gradients from the binary classification objective are backpropagated through the classification head and the full Transformer stack, enabling the model to learn discriminative clinical representations.

\subsubsection{Multi-Label Classification Head (CheXpert-Style Labels):}

For multi-label clinical classification tasks, multiple findings may co-occur in the same note (e.g., edema and pleural effusion). Therefore, each label is treated as an independent binary output. In this setting, the classification head produces a vector of logits $\mathbf{z} \in \mathbb{R}^{C}$:
\[
\mathbf{z} = W h_T + b,
\]
where $C$ is the total number of clinical labels. The probability of each label $c$ being present is computed using an element-wise sigmoid:
\[
P(y_c=1 \mid x) = \sigma(z_c), \quad c \in \{1,\dots,C\}.
\]

Training is performed using a multi-label binary cross-entropy objective (often implemented as \texttt{BCEWithLogitsLoss}, defined as:
\[
\mathcal{L}_{\text{multi}} = -\sum_{c=1}^{C} \left[ y_c \log \sigma(z_c) + (1-y_c)\log(1-\sigma(z_c)) \right],
\]
where $\mathbf{y} = (y_1,\dots,y_C) \in \{0,1\}^{C}$ denotes the ground-truth multi-label vector.

By relying on the last-token representation $h_T$, this unified strategy preserves the architectural simplicity of GPT-style models while supporting both binary and multi-label clinical prediction tasks on long and unstructured radiology narratives.

%% file: loading_pretrained_weights.tex
\section{Pretraining and Initialization from OpenAI Open-Weight Checkpoints}\label{pre-train-weight-load}
GPT-style decoder-only Transformer models are originally pretrained using a self-supervised autoregressive language modeling objective. Given a tokenized input sequence $x_{1:T} = (x_1, \ldots, x_T)$, the model is trained to maximize the conditional likelihood of each token given all previous tokens:
\[
\log p_\theta(x_{1:T}) = \sum_{t=1}^{T} \log p_\theta(x_t \mid x_{<t}),
\]
which is optimized via cross-entropy loss over large-scale unlabeled text corpora \cite{radford2019language, brown2020language}. This pretraining procedure enables the model to learn general-purpose syntactic and semantic representations that can be transferred to a wide range of downstream tasks.

In this work, large-scale language model pretraining is not performed from scratch. Instead, a transfer learning strategy is adopted by initializing the Transformer backbone using publicly released \emph{open-weight} pretrained checkpoints provided by OpenAI. Specifically, pretrained parameters $\theta_0$ are loaded from an OpenAI-released GPT checkpoint (e.g., GPT-2), including the token embedding layer, positional embeddings, and the full stack of masked self-attention and feed-forward blocks \cite{radford2019language, openai2019gpt2}. These pretrained weights serve as the starting point for subsequent task adaptation.

After loading the pretrained checkpoint, the Transformer backbone is used as an initialization for downstream task adaptation. This initialization strategy substantially reduces computational cost compared to training from scratch and improves sample efficiency by leveraging linguistic knowledge acquired during large-scale pretraining. Prior studies have demonstrated that such transfer learning approaches are particularly effective for domain-specific natural language processing tasks, including those involving long and unstructured clinical narratives \cite{devlin2019bert, huang2019clinicalbert}.

Overall, initializing the proposed architecture from OpenAI’s open-weight pretrained models enables efficient domain adaptation while preserving the strong contextual representation capabilities learned during pretraining. Details of the supervised optimization procedure for clinical classification are described in the subsequent fine-tuning section.

%% file: finetuning.tex
\section{Step-by-Step Selective Fine-Tuning Procedure (with Accuracy-Oriented Evaluation)}
Starting from the pretrained initialization described in previous section, the model is fine-tuned for supervised clinical text classification.
\begin{figure*}[!htbp]%
    \centering
   \includegraphics[height=14cm, width=12cm]{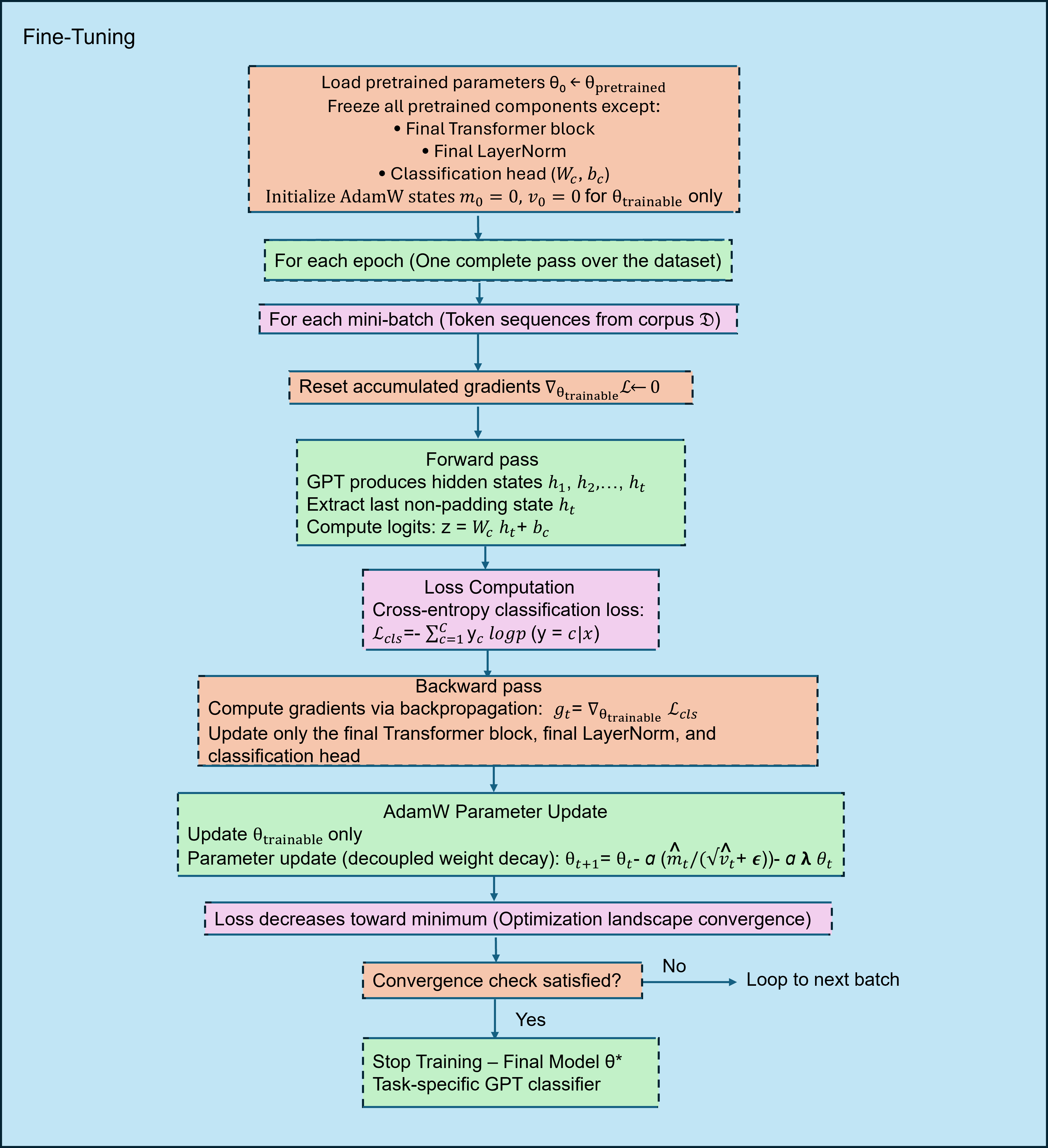} %
    \caption{Selective fine-tuning workflow for a GPT-based classification model. Pretrained parameters are loaded and all layers are frozen except the final Transformer block, final LayerNorm, and the task-specific classification head. Fine-tuning proceeds over mini-batches using cross-entropy loss and AdamW updates applied only to trainable parameters, yielding a task-specific GPT classifier.}

   \label{fine-tuning}
\end{figure*}

Figure~\ref{fine-tuning} summarizes the supervised fine-tuning workflow used to adapt a pretrained GPT model for sequence classification while updating only a small subset of parameters. Training is conducted for a fixed number of epochs over the fine-tuning dataset. 
Although Figure~\ref{fine-tuning} includes a convergence check for completeness, this check is used 
only to monitor the stabilization of the validation loss and does not replace the 
epoch-based stopping criterion. This formulation reflects common practice in 
Transformer fine-tuning, where a predefined number of epochs is used while 
convergence indicators are tracked for diagnostic purposes. This section explains each step in detail and clarifies how selective fine-tuning reduces computation and overfitting risk while preserving general language representations learned during pretraining \cite{radford2019language,vaswani2017attention,peters2019tune}.

A pretrained GPT-based language model is fine-tuned to perform sequence-level classification tasks on labeled clinical text. Fine-tuning adapts a model pretrained on large-scale unlabeled corpora to a downstream supervised task, avoiding the need to train all parameters from scratch while leveraging pretrained linguistic representations.

Given a labeled dataset
\[
\mathcal{D} = \{(x^{(i)}, y^{(i)})\}_{i=1}^{N},
\]
each input sequence $x^{(i)} = (x_1, x_2, \dots, x_T)$ consists of a variable-length token sequence, and $y^{(i)}$ denotes the corresponding class label.

\subsubsection{Input Representation and Data Loader:}

Each input sequence is tokenized using the GPT tokenizer and converted into a fixed-length representation through truncation and padding. Sequences exceeding the maximum context length are truncated, while shorter sequences are padded using a designated padding token. This preprocessing enables efficient mini-batch training and is standard practice in Transformer-based models ~\cite{wolf2020transformers}.

A data loader groups tokenized sequences into mini-batches and produces attention masks that distinguish valid tokens from padding tokens. For a mini-batch of size $B$, the data loader outputs an input tensor $\mathbf{X} \in \mathbb{R}^{B \times T}$, an attention mask $\mathbf{M} \in \{0,1\}^{B \times T}$, and a label vector $\mathbf{y} \in \mathbb{R}^{B}$.

\paragraph{Step 1: Load pretrained parameters}
Fine-tuning begins by initializing model parameters from a pretrained checkpoint:
\[
\theta_{0} \leftarrow \theta_{\text{pretrained}}.
\]
This step transfers general linguistic knowledge learned from large-scale self-supervised training to the downstream clinical classification task, which is a standard transfer-learning strategy for Transformer language models \cite{radford2019language,devlin2019bert}.

\paragraph{Step 2: Freeze all parameters except the final block, final LayerNorm, and classification head}
In this work, all pretrained parameters are frozen except:
(i) the final Transformer block, (ii) the final layer normalization module, and (iii) the task-specific classification head $(W_{c}, b_{c})$. Formally, let $\theta = \{\theta_{\text{frozen}}, \theta_{\text{trainable}}\}$, where
\[
\theta_{\text{trainable}} = \{\theta_{\text{block-}L}, \theta_{\text{LN-final}}, W_{c}, b_{c}\}.
\]
During optimization, gradients are computed and applied only to $\theta_{\text{trainable}}$. This selective strategy is motivated by evidence that higher Transformer layers tend to capture more task-specific representations, whereas lower layers encode more general and transferable features \cite{peters2019tune,yosinski2014transfer}.

\paragraph{Step 3: Initialize the classification head}
For $C$ classes, a linear classification head is attached to the final sequence representation:
\[
\mathbf{z} = W_{c} h_{T} + b_{c},
\]
where $W_{c} \in \mathbb{R}^{C \times d_{\text{model}}}$ and $b_{c} \in \mathbb{R}^{C}$ are trainable parameters. This head maps the final hidden representation to class logits and is standard for Transformer-based classification systems \cite{devlin2019bert}.

\paragraph{Step 4: Iterate over epochs}
Training proceeds for a fixed number of epochs, where each epoch represents one full pass over the labeled training data. Epoch-based training is standard in deep learning optimization and enables stable learning rate scheduling and progress monitoring \cite{goodfellow2016deep}.

\paragraph{Step 5: Iterate over mini-batches using a data loader}
The dataset is processed in mini-batches to improve computational efficiency and memory usage. A data loader produces batches of:
\[
(\mathbf{X}, \mathbf{M}, \mathbf{y}),
\]
where $\mathbf{X}$ contains token IDs, $\mathbf{M}$ is an attention mask that distinguishes valid tokens from padding, and $\mathbf{y}$ contains the ground-truth labels. Mini-batch training is standard for stochastic optimization and is the default practice in modern deep learning frameworks \cite{paszke2019pytorch}.

\paragraph{Step 6: Reset accumulated gradients}
Before each parameter update, gradients are reset:
\[
\nabla \theta_{\text{trainable}} \leftarrow 0.
\]
This prevents unintended accumulation of gradients across batches, which would otherwise distort the optimization step and lead to incorrect parameter updates \cite{paszke2019pytorch}.

\paragraph{Step 7: Forward pass through frozen and trainable components}
Each mini-batch is passed through the GPT model. Frozen components (embeddings and earlier Transformer blocks) compute representations but do not receive gradient updates. The final Transformer block and final LayerNorm remain trainable and adapt to the downstream task. The model outputs hidden states:
\[
\mathbf{H} = (h_{1}, h_{2}, \dots, h_{T}),
\]
where causal self-attention ensures that $h_{t}$ depends only on tokens up to position $t$ \cite{vaswani2017attention,radford2019language}.

\paragraph{Step 8: Extract the last non-padding hidden state}
Because GPT is decoder-only and does not include a dedicated classification token (e.g., \texttt{[CLS]}), the final hidden state corresponding to the last non-padding token, denoted $h_{T}$ is used. This choice is principled because the last state has integrated information from the full preceding context under causal attention, making it a natural sequence-level representation for classification \cite{radford2019language}.

\paragraph{Step 9: Compute logits and supervised loss}
Class logits are computed with the classification head and converted to probabilities via softmax:
\[
p(y=c \mid x) = \frac{\exp(z_{c})}{\sum_{j=1}^{C} \exp(z_{j})}.
\]
The supervised objective is the cross-entropy loss:
\[
\mathcal{L}_{\text{cls}} = -\sum_{c=1}^{C} y_{c}\log p(y=c \mid x),
\]
which is the standard loss for multi-class classification \cite{bishop2006pattern,goodfellow2016deep}.

\paragraph{Step 10: Backward pass restricted to trainable parameters}
Gradients are computed by backpropagation:
\[
g_{t} = \nabla_{\theta_{\text{trainable}}}\mathcal{L}_{\text{cls}},
\]
and flow only through the final block, final LayerNorm, and the classification head. Freezing the remaining layers prevents updates to general language representations and reduces overfitting risk when labeled data are limited \cite{peters2019tune,yosinski2014transfer}.

\paragraph{Step 11: Update trainable parameters using AdamW}
Only $\theta_{\text{trainable}}$ is updated using AdamW, which decouples weight decay from the gradient-based update and is commonly used for Transformer fine-tuning:
\[
\theta_{\text{trainable}} \leftarrow \theta_{\text{trainable}} - \alpha \,\text{AdamW}(g_{t}).
\]
where $\alpha$ denotes the learning rate.
AdamW is widely adopted in modern NLP training and is specifically recommended for Transformer optimization \cite{loshchilov2019adamw}.

\paragraph{Step 12: Monitor training and validation performance (loss and accuracy)}
During training, the training loss is tracked, and evaluation on a validation set is performed optionally. Accuracy is computed as:
\[
\text{Accuracy} = \frac{1}{N}\sum_{i=1}^{N}\mathbb{I}(\hat{y}_{i}=y_{i}),
\quad
\hat{y}_{i}=\arg\max_{c} z_{i,c}.
\]
This provides an interpretable measure of classification performance and supports early stopping or model selection when validation accuracy plateaus \cite{bishop2006pattern,goodfellow2016deep}.

\paragraph{Step 13: Convergence check and final model}
Training is performed for a fixed number of epochs, after which the resulting parameters define the fine-tuned classifier:
\[
\theta^{*} = \arg\min_{\theta_{\text{trainable}}}\mathcal{L}_{\text{cls}},
\]
where only the selected modules have been adapted. This yields a task-specialized GPT-based classifier while preserving pretrained knowledge in frozen components \cite{radford2019language,peters2019tune}.

%% file: Trainable_parameter_and_complexity_analysis_finetuning.tex
\section{Selective Fine-Tuning Parameterization and Time Complexity Analysis}

This section explains, in detail, how selective fine-tuning is performed in the GPT-2–based model and why this strategy substantially reduces computational cost. The explanation is written to be accessible to readers who may be new to large language models and Transformer architectures.

\subsection{Overview of the Fine-Tuning Strategy}

The fine-tuned model is derived from the GPT-2 decoder-only Transformer architecture described in the previous section. At a high level, GPT-2 consists of three main components: (i) an embedding layer that converts discrete tokens into continuous vectors, (ii) a stack of Transformer blocks that progressively refine contextual representations, and (iii) a final normalization stage prior to output generation \cite{radford2019language}.

In standard full fine-tuning, all parameters of the model are updated using task-specific labeled data. However, GPT-2 contains over one hundred million parameters, making full fine-tuning computationally expensive and prone to overfitting when labeled data are limited. To address this, a \emph{selective fine-tuning} strategy is adopted in which most pretrained parameters are frozen. Specifically, all parameters are frozen except those in the final Transformer block, the final layer normalization, and the task-specific classification head. The remainder of this section explains this design choice and its implications for parameter count and training-time complexity.

\subsection{Frozen Components and Parameter Accounting}

The proposed model adopts a selective fine-tuning strategy based on GPT-2 (small), in which the embedding layers and the lower Transformer blocks are kept fixed during task-specific training. This design leverages the strong general-purpose representations learned during large-scale pretraining, while substantially reducing optimization cost and memory requirements during fine-tuning.

\paragraph{Embedding Layers:}

GPT-2 includes a learned token embedding matrix
$\mathbf{W}_{\mathrm{te}} \in \mathbb{R}^{V \times d_{\mathrm{model}}}$,
which maps each token in the vocabulary to a continuous vector representation, and a learned positional embedding matrix
$\mathbf{W}_{\mathrm{pe}} \in \mathbb{R}^{T_{\max} \times d_{\mathrm{model}}}$,
which encodes absolute token positions within the input sequence \cite{radford2019language}. These embeddings capture general lexical semantics and positional structure and are shared across all downstream tasks.

In the GPT-2 (small) configuration, the hidden dimension is
$d_{\mathrm{model}} = 768$, the vocabulary size is
$V = 50{,}257$, and the maximum context length is
$T_{\max} = 1024$. The total number of embedding parameters is therefore
\[
\underbrace{50{,}257 \times 768}_{\text{token embeddings}}
+
\underbrace{1024 \times 768}_{\text{positional embeddings}}
=
38{,}597{,}376
\;\approx\;
38.6 \text{ million}.
\]

During selective fine-tuning, both embedding matrices are frozen. They participate in forward propagation by producing input representations of shape
$\mathbb{R}^{B \times T \times 768}$, where $B$ is the batch size and $T$ is the sequence length, but no gradients are computed or stored for these parameters during backpropagation.

\paragraph{Frozen Transformer Blocks:}

Following the embedding layer, GPT-2 consists of a stack of $L = 12$ decoder-only Transformer blocks \cite{radford2019language,brown2020language}. Each block refines token representations using masked self-attention and position-wise feed-forward transformations. In the proposed setup, the first $L-1 = 11$ Transformer blocks are frozen, while only the final block and task-specific head remain trainable.

\paragraph{Transformer Block Parameterization:}

Each Transformer block operates at hidden dimension $d_{\mathrm{model}} = 768$ and uses $h = 12$ attention heads, yielding per-head dimensions $d_k = d_v = 64$, consistent with the original Transformer architecture \cite{vaswani2017attention}.

\subparagraph{Multi-Head Self-Attention:}

The self-attention sublayer contains learned linear projections for queries, keys, and values, as well as an output projection:
\[
\mathbf{W}_Q, \mathbf{W}_K, \mathbf{W}_V, \mathbf{W}_O \in \mathbb{R}^{768 \times 768}.
\]
The total number of parameters in the attention sublayer is
\[
4 \times 768 \times 768
=
2{,}359{,}296.
\]

\subparagraph{Feed-Forward Network:}

The feed-forward network expands the hidden dimension by a factor of four,
$d_{\mathrm{ff}} = 3{,}072$, and applies two linear transformations:
\[
\mathbf{W}_1 \in \mathbb{R}^{768 \times 3{,}072},
\qquad
\mathbf{W}_2 \in \mathbb{R}^{3{,}072 \times 768}.
\]
This sublayer contains
\[
768 \times 3{,}072 + 3{,}072 \times 768
=
4{,}718{,}592
\]
parameters.

\subparagraph{Layer Normalization:}

Each Transformer block includes two Layer Normalization layers, each with a learnable scale and bias of size $768$ \cite{ba2016layer}. The total number of LayerNorm parameters per block is
\[
2 \times 2 \times 768 = 3{,}072.
\]

\paragraph{Total Parameters per Transformer Block:}

Summing all components, the total number of parameters in a single GPT-2 (small) Transformer block is
\[
2{,}359{,}296
+
4{,}718{,}592
+
3{,}072
=
7{,}080{,}960
\;\approx\;
7.08 \text{ million}.
\]

With $11$ frozen Transformer blocks, this corresponds to approximately
\[
11 \times 7.08 \text{ million} \approx 77.9 \text{ million}
\]
frozen parameters.

\paragraph{Overall Frozen Parameter Count:}

Combining the frozen embedding layers ($\approx 38.6$ million parameters) and the first $11$ frozen Transformer blocks ($\approx 77.9$ million parameters), the total number of frozen parameters is approximately
\[
116.5 \text{ million}.
\]

Although these components actively participate in forward propagation and generate contextualized representations of shape $\mathbb{R}^{B \times T \times 768}$, they incur no gradient computation or optimizer updates during backpropagation. This selective freezing strategy substantially reduces training-time memory usage and optimization overhead, while preserving the rich linguistic and contextual knowledge acquired during pretraining. Such behavior aligns with established transfer learning principles, where lower layers capture general-purpose features that transfer effectively across tasks \cite{howard2018universal}.

\subsection{Trainable Final Transformer Block}

The only Transformer block that remains trainable is the \emph{final} block in the stack (i.e., block $L$). This block is the topmost block in the architecture and directly precedes the final layer normalization and classification head. The final Transformer block produces the most abstract and task-specific representations, making it the most effective location for domain adaptation \cite{peters2019tune,yosinski2014transfer}.

Each Transformer block contains two major subcomponents: a multi-head self-attention module and a position-wise feed-forward network. For a hidden dimension $d_{\text{model}} = 768$ and a feed-forward dimension $d_{\text{ff}} = 3072$, the parameter count can be understood as follows.

The self-attention module requires four linear transformations: three to generate the query, key, and value representations, and one additional projection applied after attention outputs are concatenated. Each of these transformations maps from $d_{\text{model}}$ to $d_{\text{model}}$, resulting in a total of $4 d_{\text{model}}^2 = 4 \times 768^2 \approx 2.36$ million parameters. The feed-forward network consists of two linear layers, one expanding the representation from $d_{\text{model}}$ to $d_{\text{ff}}$, and one projecting it back. This contributes $2(d_{\text{model}} \times d_{\text{ff}}) = 2 \times (768 \times 3072) \approx 4.72$ million parameters. Together, the final Transformer block contains approximately $7.08$ million trainable parameters.

From a time-complexity perspective, this block dominates the fine-tuning cost. For an input sequence of length $T$, the self-attention mechanism scales as $\mathcal{O}(T^2 d_{\text{model}})$ due to pairwise interactions between tokens, while the feed-forward network scales as $\mathcal{O}(T d_{\text{model}} d_{\text{ff}})$. Because backpropagation is restricted to a single Transformer block, the overall training-time complexity is substantially lower than that of full-model fine-tuning, which would require gradient computation through all $L$ blocks \cite{gururangan2020don}.

\subsection{Trainable Final Layer Normalization}

After the Transformer stack, GPT-2 applies a final layer normalization operation ~\cite{ba2016layer}. This layer includes a learnable scale parameter $\boldsymbol{\gamma}_{\text{final}}$ and a learnable shift parameter $\boldsymbol{\beta}_{\text{final}}$, each of dimension $d_{\text{model}}$. As a result, this component introduces $2 d_{\text{model}} = 1{,}536$ trainable parameters. The computational cost of layer normalization scales as $\mathcal{O}(T d_{\text{model}})$ and is negligible compared to the cost of self-attention and feed-forward operations.

\subsection{Trainable Classification Head}

To adapt GPT-2 for supervised classification, a task-specific classification head is appended to the final hidden representation. This head consists of a linear projection with weight matrix $\mathbf{W}_{\text{cls}} \in \mathbb{R}^{d_{\text{model}} \times C}$ and bias vector $\mathbf{b}_{\text{cls}} \in \mathbb{R}^{C}$, where $C$ is the number of output classes. For binary classification ($C = 2$), this results in $768 \times 2 + 2 = 1{,}538$ trainable parameters. The time complexity of this layer scales as $\mathcal{O}(T d_{\text{model}} C)$ and is negligible relative to the Transformer block, as is typical in Transformer-based classification models \cite{devlin2019bert}.

\subsection{Overall Trainable Parameter Count and Reduction}

Under this selective fine-tuning strategy, the total number of trainable parameters is dominated by the final Transformer block. Specifically, approximately $7.08$ million parameters are trainable in the final block, with an additional $0.0015$ million parameters each from the final layer normalization and the classification head. In total, this yields roughly $7.08$ million trainable parameters.

Compared to the full GPT-2 (small) model, which contains approximately $124$ million parameters \cite{radford2019language}, this represents a reduction of more than $94\%$ in the number of parameters updated during training.

\subsection{Implications for Training Efficiency}

By restricting gradient updates to the final Transformer block and lightweight task-specific layers, the fine-tuning process significantly reduces both training time and memory consumption. This selective optimization approach enables efficient domain adaptation while preserving the general linguistic knowledge acquired during large-scale pretraining, making it particularly well suited for clinical and other low-resource text classification tasks \cite{gururangan2020don}.

%% file: Dataset_and_ground_truth_label_creation.tex
\section{Dataset and Label Construction}
The experiments in this study were conducted using the MIMIC-IV-Note database, a large-scale, de-identified clinical text corpus released through PhysioNet \cite{goldberger2000physionet}. MIMIC-IV contains electronic health records associated with patients admitted to intensive care units and hospital wards at the Beth Israel Deaconess Medical Center, spanning more than a decade of clinical care \cite{johnson2023mimiciv}. The dataset includes a wide range of unstructured clinical notes, such as discharge summaries, radiology reports, nursing notes, and physician documentation, which reflect real-world clinical language and documentation practices.

In this work, radiology reports from the MIMIC-IV-Note v2.2 release are utilized, and a labeled dataset for supervised learning is constructed by deriving uncertainty-aware CheXpert-style labels directly from the report text. Each radiology note was retained together with key identifiers (\texttt{note\_id}, \texttt{subject\_id}, \texttt{hadm\_id}) and report timestamps, and subsequently transformed into a structured feature table by creating 14 CheXpert-derived condition label columns. Because structured diagnosis codes are not consistently available or sufficiently specific for all radiology notes, weak supervision was generated using rule-based extraction (keyword detection with negation and uncertainty handling), enabling scalable label construction while preserving clinically meaningful ambiguity. The final labeled dataset was implemented as a BigQuery table (\texttt{radiology\_1\_chexpert14\_uncertainty}), containing both the original report text and the derived label fields for downstream GPT-based classification experiments.

Access to the MIMIC-IV dataset requires completion of required human-subjects training and acceptance of the PhysioNet Data Use Agreement. A subset of 500,000 reports was selected to balance computational feasibility with sufficient statistical power for comparative evaluation. All analyses were performed in compliance with these requirements and adhere to the Health Insurance Portability and Accountability Act (HIPAA) Safe Harbor provisions. The dataset used in this study is publicly available for credentialed researchers at PhysioNet:
\begin{center}
\texttt{https://physionet.org/content/mimic-iv-note/2.2/}
\end{center}

The use of MIMIC-IV ensures reproducibility and comparability with prior clinical NLP studies, as it has become a standard benchmark dataset for evaluating machine learning and natural language processing methods on real-world clinical text \cite{johnson2023mimiciv}.

\begin{table*}[!htbp]
\centering
\caption{Clinical interpretation of the CheXpert-derived labels used in this study.}
\label{tab:chexpert_label_definitions}
\begin{tabular}{p{0.24\linewidth} p{0.70\linewidth}}

\textbf{Label} & \textbf{Clinical meaning / definition} \\

No Finding & No evidence of acute cardiopulmonary abnormality; report indicates normal or unremarkable findings. \\
Enlarged Cardiomediastinum & Enlargement or widening of the cardiomediastinal silhouette or mediastinum. \\
Cardiomegaly & Enlarged cardiac silhouette or enlarged heart size. \\
Lung Opacity & Non-specific increased lung density/opacity, including airspace opacities; can overlap with consolidation or edema. \\
Lung Lesion & Focal lung lesion such as mass or suspicious nodule. \\
Edema & Findings consistent with pulmonary edema or vascular congestion. \\
Consolidation & Alveolar consolidation; dense airspace filling (often infectious/inflammatory). \\
Pneumonia & Findings consistent with pneumonia (suspected or confirmed). \\
Atelectasis & Collapse or volume loss consistent with atelectasis. \\
Pneumothorax & Presence of air in pleural space, consistent with pneumothorax. \\
Pleural Effusion & Pleural fluid accumulation consistent with pleural effusion. \\
Pleural Other & Other pleural abnormalities such as pleural thickening or pleural scarring. \\
Fracture & Skeletal fracture such as rib or clavicle fracture. \\
Support Devices & Presence of medical devices (e.g., endotracheal tube, central venous catheter, NG tube, pacemaker). \\

\end{tabular}

\vspace{0.25em}
\footnotesize{\emph{Note:} Label set follows the CheXpert schema \cite{irvin2019chexpert}.}
\end{table*}

\begin{algorithm*}[!htbp]
\caption{Uncertainty-aware CheXpert-style label generation from radiology report text}
\label{alg:chexpert_uncertainty_labeling}
\footnotesize
\begin{algorithmic}[1]
\Require Report text $T$; conditions $\mathcal{C}$; keyword lists $\{\mathcal{K}_c\}$; cue sets $\mathcal{N},\mathcal{U}$;
normal patterns $\mathcal{P}_{NF}$; window size $w$.
\Ensure 3-class labels $\{y_{c,3}\}$ and derived binaries $\{y_{c,\mathrm{posonly}},y_{c,\mathrm{posorunc}}\}$; summary labels
\texttt{label\_any\_disease\_posonly}, \texttt{label\_any\_disease\_pos\_or\_unc}, \texttt{label\_no\_finding\_strict} (Table~\ref{tab:chexpert14_uncertainty_columns}).

\State $T \gets \Call{NormalizeText}{T}$ \Comment{lowercase, collapse whitespace, keep punctuation}
\State \texttt{y\_no\_finding\_phrase} $\gets \mathbb{I}\!\left[\Call{MatchAnyPattern}{T,\mathcal{P}_{NF}}\right]$

\ForAll{$c \in \mathcal{C}$}
  \State $y_{c,3} \gets \texttt{NULL}$
  \State $\texttt{pos} \gets 0$; \ $\texttt{unc} \gets 0$; \ $\texttt{negAll} \gets 1$
  \State $\mathcal{M} \gets \Call{FindMentions}{T,\mathcal{K}_c}$
  \If{$\mathcal{M}=\emptyset$}
    \State \textbf{continue}
  \EndIf

  \ForAll{$m \in \mathcal{M}$}
    \State $S \gets \Call{GetContextWindow}{T,m,w}$ \Comment{e.g., same sentence or $\pm w$ chars}
    \State $\texttt{isNeg} \gets \Call{HasNegation}{S,\mathcal{N}}$
    \State $\texttt{isUnc} \gets \Call{HasUncertainty}{S,\mathcal{U}}$
    \If{$\texttt{isNeg}=0$}
      \State $\texttt{negAll} \gets 0$
      \If{$\texttt{isUnc}=1$}
        \State $\texttt{unc} \gets 1$
      \Else
        \State $\texttt{pos} \gets 1$
      \EndIf
    \EndIf
  \EndFor

  \If{$\texttt{pos}=1$}
    \State $y_{c,3} \gets 1$ \Comment{affirmed present}
  \ElsIf{$\texttt{unc}=1$}
    \State $y_{c,3} \gets -1$ \Comment{uncertain mention}
  \ElsIf{$\texttt{negAll}=1$}
    \State $y_{c,3} \gets 0$ \Comment{explicitly negated}
  \Else
    \State $y_{c,3} \gets \texttt{NULL}$ \Comment{rare ambiguous fallback}
  \EndIf

  \State $y_{c,\mathrm{posonly}} \gets \mathbb{I}[y_{c,3}=1]$
  \State $y_{c,\mathrm{posorunc}} \gets \mathbb{I}[y_{c,3}\in\{1,-1\}]$
\EndFor

\State \texttt{label\_any\_disease\_posonly} $\gets \mathbb{I}\!\left[\sum_{c\in\mathcal{C}} y_{c,\mathrm{posonly}} > 0\right]$
\State \texttt{label\_any\_disease\_pos\_or\_unc} $\gets \mathbb{I}\!\left[\sum_{c\in\mathcal{C}} y_{c,\mathrm{posorunc}} > 0\right]$
\State \texttt{label\_no\_finding\_strict} $\gets \mathbb{I}\!\left[\texttt{y\_no\_finding\_phrase}=1 \wedge \sum_{c\in\mathcal{C}} y_{c,\mathrm{posorunc}} = 0\right]$

\end{algorithmic}
\end{algorithm*}

\begin{table*}[!htbp]
\centering
\caption{Description of newly created uncertainty-aware CheXpert-style label columns derived from radiology report text.}
\label{tab:chexpert14_uncertainty_columns}
\renewcommand{\arraystretch}{1.15}
\begin{tabular}{p{0.34\textwidth} p{0.60\textwidth}}
\hline
\textbf{Column name} & \textbf{Meaning (how to interpret the column)} \\
\hline

\texttt{y\_no\_finding\_phrase} &
Binary indicator of an explicit normal/``no acute abnormality'' statement in the report text (e.g., ``no acute cardiopulmonary abnormality''). This phrase-based indicator is used in constructing \texttt{label\_no\_finding\_strict}. \\

\hline
\multicolumn{2}{l}{\textbf{Uncertainty-aware 3-class label columns (per CheXpert condition)}}\\
\hline

\texttt{y\_enlarged\_cardiomediastinum\_3} &
3-class status for enlarged cardiomediastinum: \texttt{1}=affirmed present; 0=explicitly negated (e.g., ``no mediastinal widening''); \texttt{-1}=uncertain/hedged (e.g., ``possible mediastinal widening''); \texttt{NULL}=not mentioned. \\

\texttt{y\_cardiomegaly\_3} &
3-class status for cardiomegaly with the same coding: \texttt{1}=present; 0=negated; \texttt{-1}=uncertain; \texttt{NULL}=not mentioned. \\

\texttt{y\_lung\_opacity\_3} &
3-class status for lung opacity (e.g., ``opacity'', ``airspace opacity'') with the same coding: \texttt{1}=present; 0=negated; \texttt{-1}=uncertain; \texttt{NULL}=not mentioned. \\

\texttt{y\_lung\_lesion\_3} &
3-class status for lung lesion-type findings (e.g., nodule, mass, lesion) with the same coding. \\

\texttt{y\_edema\_3} &
3-class status for pulmonary edema/edema with the same coding. \\

\texttt{y\_consolidation\_3} &
3-class status for consolidation with the same coding. \\

\texttt{y\_pneumonia\_3} &
3-class status for pneumonia with the same coding. \\

\texttt{y\_atelectasis\_3} &
3-class status for atelectasis or collapse with the same coding. \\

\texttt{y\_pneumothorax\_3} &
3-class status for pneumothorax with the same coding. \\

\texttt{y\_pleural\_effusion\_3} &
3-class status for pleural effusion (including ``effusion'') with the same coding. \\

\texttt{y\_pleural\_other\_3} &
3-class status for pleural ``other'' findings (e.g., pleural thickening/plaque/scarring) with the same coding. \\

\texttt{y\_fracture\_3} &
3-class status for fracture (e.g., rib fracture) with the same coding. \\

\texttt{y\_support\_devices\_3} &
3-class status for support devices (e.g., endotracheal tube, central venous catheter, PICC, chest tube, pacemaker) with the same coding. \\

\hline
\multicolumn{2}{l}{\textbf{Binary label variants derived from the 3-class labels}}\\
\hline

\texttt{y\_*\_bin\_posonly} &
Conservative binary variant for each abnormal CheXpert condition: \texttt{1} if the corresponding \texttt{y\_*\_3} equals \texttt{1} (affirmed present); 0 otherwise (including uncertain, negated, or not mentioned). Intended for high-precision positive labeling. \\

\texttt{y\_*\_bin\_pos\_or\_unc} &
Sensitive binary variant for each abnormal CheXpert condition: \texttt{1} if the corresponding \texttt{y\_*\_3} is \texttt{1} (present) or \texttt{-1} (uncertain); 0 otherwise. Intended for analyses that treat uncertain mentions as positive and for sensitivity analysis. \\

\hline
\multicolumn{2}{l}{\textbf{Overall (report-level) summary labels}}\\
\hline

\texttt{label\_any\_disease\_posonly} &
Overall abnormality indicator (conservative): \texttt{1} if \textit{any} of the 13 abnormal CheXpert conditions (all except ``No Finding'') have \texttt{*\_bin\_posonly}=1; otherwise 0. \\

\texttt{label\_any\_disease\_pos\_or\_unc} &
Overall abnormality indicator (sensitive): \texttt{1} if \textit{any} abnormal CheXpert condition has \texttt{*\_bin\_pos\_or\_unc}=1; otherwise 0. \\

\texttt{label\_no\_finding\_strict} &
Strict normal indicator: \texttt{1} only if \texttt{y\_no\_finding\_phrase}=1 \textit{and} none of the abnormal conditions are positive or uncertain (i.e., the sum of all abnormal \texttt{*\_bin\_pos\_or\_unc} equals 0); otherwise 0. \\

\hline
\end{tabular}
\end{table*}

\subsection{Uncertainty-aware CheXpert-style label construction from radiology report text}
\label{subsec:chexpert_labels}
To generate structured supervision from unstructured radiology reports, the CheXpert label schema is adopted, which is a widely used uncertainty-aware labeling standard for radiology report-derived targets \cite{chexpert_irvin2019}. CheXpert defines 14 clinically relevant report-level observations (including ``No Finding'') and explicitly highlights the importance of uncertainty handling due to the prevalence of hedged radiology language (e.g., ``possible'', ``likely'', ``cannot exclude'') \cite{chexpert_irvin2019}. Because structured diagnosis codes were not consistently available for all notes,  note-level weak labels are constructed directly from report text. To improve interpretability and reproducibility, Table~\ref{tab:chexpert_label_definitions} summarizes the clinical interpretation of each CheXpert-derived label used in this work. The label schema follows the CheXpert label set \cite{irvin2019chexpert}, and labels were assigned from unstructured radiology report text using rule-based extraction.


\paragraph{Motivation for uncertainty-aware labeling:}
Radiology narratives frequently contain uncertainty statements and differential considerations. Naïve keyword-based binary labeling may introduce label noise by conflating uncertain mentions with definitive positives. CheXpert explicitly models uncertainty as a distinct label state and demonstrates its importance for training reliable prediction models \cite{chexpert_irvin2019}. Therefore, the label construction distinguishes affirmed findings, negated findings, and uncertain findings.

\paragraph{Primary 3-class condition labels:}
For each abnormal CheXpert condition, a 3-class label column \texttt{y\_*\_3} is defined as the primary representation (Table~\ref{tab:chexpert14_uncertainty_columns}). Each condition label is encoded as: \texttt{1} for affirmed presence, 0 for explicit negation, \texttt{-1} for uncertain mention, and \texttt{NULL} when the condition is not mentioned. This encoding preserves clinically meaningful ambiguity and aligns with the uncertainty-aware labeling principle of CheXpert \cite{chexpert_irvin2019}.

\paragraph{Derived binary label variants:}
To support supervised models requiring binary targets, two deterministic binary variants are derived from each 3-class label. The conservative label \texttt{y\_*\_bin\_posonly} is set to \texttt{1} only when the condition is affirmed (\texttt{y\_*\_3 = 1}); otherwise it is set to 0. The sensitive label \texttt{y\_*\_bin\_pos\_or\_unc} is set to \texttt{1} when the condition is affirmed or uncertain (\texttt{y\_*\_3 $\in \{1,-1\}$}); otherwise it is set to 0. This dual representation supports both high-precision modeling (pos-only) and sensitivity analysis (pos-or-unc) by explicitly quantifying the effect of uncertainty.

\paragraph{Report-level summary labels:}
Report-level summary outcomes are additionally computed. \texttt{label\_any\_disease\_posonly} indicates whether at least one abnormal condition is affirmed present based on \texttt{*\_bin\_posonly}. \texttt{label\_any\_disease\_pos\_or\_unc} indicates whether at least one abnormal condition is affirmed or uncertain. Finally, \texttt{label\_no\_finding\_strict} indicates strict normality: it is set to \texttt{1} only when (i) an explicit normal phrase is detected (\texttt{y\_no\_finding\_phrase = 1}) and (ii) none of the abnormal conditions are positive or uncertain. This strict definition reduces false-normal labeling, which is a known limitation of weak supervision from clinical narratives \cite{chexpert_irvin2019}.


\paragraph{Algorithmic procedure:}
Algorithm~\ref{alg:chexpert_uncertainty_labeling} formalizes the label extraction procedure. For each report, the text is normalized and scanned for condition-specific lexical patterns. For each mention, a bounded context window is searched for negation cues (e.g., ``no'', ``without'', ``negative for'') and uncertainty cues (e.g., ``possible'', ``cannot exclude'', ``suggests''). Condition-level mention decisions are aggregated into a single report-level 3-class status, and binary and summary labels are computed deterministically from these primary 3-class labels.

\subsubsection{Implementation in Google BigQuery:}
The labeling pipeline was executed directly in Google BigQuery to construct a labeled table containing the original report identifiers (\texttt{note\_id}, \texttt{subject\_id}, \texttt{hadm\_id}) together with the derived label columns. Each label was computed via \texttt{CASE WHEN} expressions using \texttt{REGEXP\_CONTAINS()} for condition keyword matching and additional regex-based constraints for negation and uncertainty detection within a local context window. 

Following label creation, prevalence statistics (e.g., per-label positive counts and overall abnormality rates) were obtained via aggregation queries in BigQuery (e.g., \texttt{SUM()} over label columns). For each label, positive counts were computed as \texttt{SUM(y\_label)} and prevalence rates were computed as \texttt{SUM(y\_label)/COUNT(*)}. The overall abnormality rate was defined as the proportion of radiology reports with at least one abnormal finding, computed using the composite indicator \texttt{label\_any\_disease}. Prevalence was computed under two binarization schemes: (i) positive-only and (ii) positive-or-uncertain (i.e., treating uncertain mentions as positive), enabling sensitivity analysis with respect to uncertainty handling.
Table~\ref{tab:chexpert_prevalence_tab} reports the prevalence of each CheXpert-derived label under positive-only and positive-or-uncertain schemes.


\begin{table*}[!htbp]
\centering
\caption{Prevalence of CheXpert-derived labels in MIMIC-IV radiology notes. Counts and percentages are reported under two binarization schemes: (i) positive-only and (ii) positive-or-uncertain (treating uncertainty as positive).}
\label{tab:chexpert_prevalence_tab}
\begin{tabular}{lrr}
Label & Positive-only, n (\%) & Positive-or-uncertain, n (\%) \\
Any abnormality & 183,046 (26.88\%) & 225,925 (33.18\%) \\
No finding (strict) & 22,894 (3.36\%) & 22,894 (3.36\%) \\
Enlarged cardiomediastinum & 702 (0.10\%) & 758 (0.11\%) \\
Cardiomegaly & 11,692 (1.72\%) & 13,133 (1.93\%) \\
Lung opacity & 21,127 (3.10\%) & 62,090 (9.12\%) \\
Lung lesion & 164 (0.02\%) & 291 (0.04\%) \\
Edema & 19,031 (2.80\%) & 33,225 (4.88\%) \\
Consolidation & 11,069 (1.63\%) & 28,445 (4.18\%) \\
Pneumonia & 508 (0.07\%) & 3,209 (0.47\%) \\
Atelectasis & 63,066 (9.26\%) & 85,525 (12.56\%) \\
Pneumothorax & 7,211 (1.06\%) & 12,176 (1.79\%) \\
Pleural effusion & 47,778 (7.02\%) & 62,976 (9.25\%) \\
Pleural other & 1,272 (0.19\%) & 1,749 (0.26\%) \\
Fracture & 30,815 (4.53\%) & 30,842 (4.53\%) \\
Support devices & 43,067 (6.33\%) & 43,067 (6.33\%) \\
\end{tabular}

\vspace{0.25em}
{\footnotesize \textit{Note:} $N = 680{,}860$ radiology reports. Positive-only labels were derived using
\texttt{*\_bin\_posonly} columns, while positive-or-uncertain labels used \texttt{*\_bin\_pos\_or\_unc} columns.}
\end{table*}

\subsection{CheXpert Label Prevalence and Uncertainty Distribution}

To characterize class imbalance and the prevalence of uncertain clinical mentions in the radiology corpus, the empirical distribution of CheXpert-derived labels is computed on a representative subset of MIMIC-IV radiology reports prior to dataset splitting. For each clinical finding, two prevalence estimates are reported: (i) \textit{positive-only}, which counts only definitively positive mentions, and (ii) \textit{positive-or-uncertain}, which treats uncertain mentions as positive, consistent with the CheXpert uncertainty-aware labeling framework.

\begin{figure*}[!htbp]
    \centering
    \includegraphics[width=\linewidth]{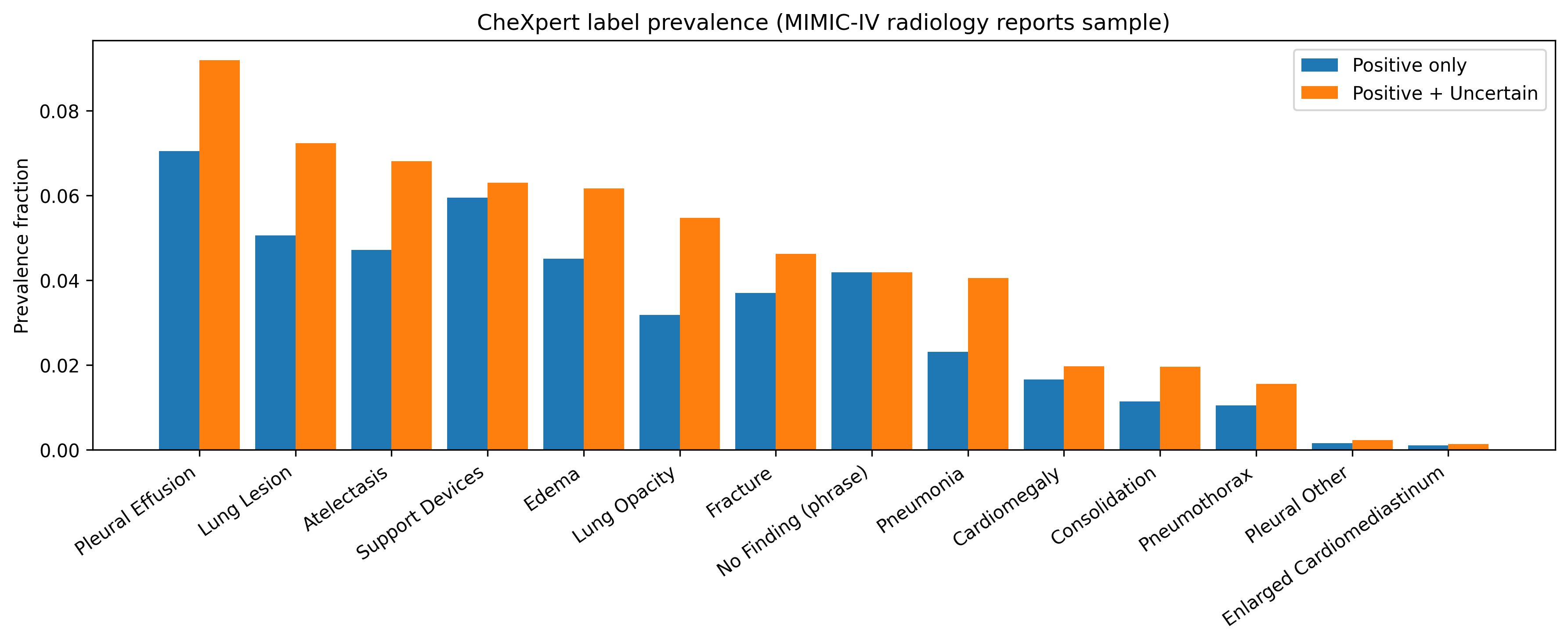}
    \caption{Prevalence of CheXpert-derived labels in a sample of MIMIC-IV radiology reports. For each label, the fraction of reports with definitive positive mentions (positive only) and the fraction including uncertain mentions (positive + uncertain) are reported. The results highlight substantial class imbalance and the prevalence of uncertainty in clinical narratives.}
    \label{fig:chexpert_prevalence}
\end{figure*}

Figure~\ref{fig:chexpert_prevalence} illustrates substantial variation in label frequency across conditions. Common findings such as pleural effusion, lung lesion, atelectasis, and support devices appear in a relatively large fraction of reports, while labels such as pneumothorax, consolidation, pleural other, and enlarged cardiomediastinum are comparatively rare. Across nearly all labels, the prevalence increases when uncertain mentions are included, highlighting the non-trivial presence of hedged or indeterminate language in radiology narratives.

These distributions demonstrate pronounced class imbalance and motivate the use of uncertainty-aware modeling, weighted sampling strategies, and evaluation metrics beyond accuracy. Moreover, the gap between positive-only and positive-or-uncertain prevalence underscores the importance of explicitly accounting for uncertainty when training and evaluating clinical text classification models.

\subsection{Train--Validation--Test Split and Data Loading}

\paragraph{Data splitting protocol:}
After preprocessing, the dataset is randomly shuffled with a fixed seed and partitioned into three mutually exclusive subsets following a 70/10/20 split: 70\% for training, 10\% for validation, and 20\% for testing. Concretely, if $N$ denotes the number of examples, index cutoffs are computed as
\[
N_{\text{train}}=\lfloor 0.7N \rfloor,\quad
N_{\text{val}}=\lfloor 0.1N \rfloor,\quad
N_{\text{test}}=N-N_{\text{train}}-N_{\text{val}},
\]
and the first $N_{\text{train}}$ rows are assigned to the training set, the next $N_{\text{val}}$ rows to the validation set, and the remainder to the test set. This procedure corresponds to the implementation:
\texttt{train\_df, validation\_df, test\_df = random\_split(df, 0.7, 0.1)}.


\paragraph{Sequence length constraint:}
GPT-2 has a fixed maximum context window of 1024 tokens (\texttt{n\_ctx}=1024). Therefore, the maximum note length was set to 1024 tokens and any input longer than this limit would require truncation if used directly. To reduce information loss and to lower computational cost, a summarized-note table is generated from the original clinical notes and trained the GPT-based classifier on the summarized text. However, it is observed that summarization often failed to preserve clinically salient keywords, uncertainty expressions, and negation cues required for CheXpert-style classification, resulting in reduced classification performance. Consequently, a strategy is adopted in which the first 1024 tokens of each clinical note are used for model training, retaining critical lexical information while remaining within the model’s context constraint. The GPT-2 configuration used in this work is consistent with the standard \texttt{gpt2} (GPT-2 small) specification: vocabulary size 50,257 and a 12-layer, 12-head transformer with 768-dimensional embeddings (approximately 124M parameters). See the reference configuration and documentation: \url{https://huggingface.co/openai-community/gpt2/blob/main/config.json}, \url{https://huggingface.co/transformers/v2.10.0/pretrained_models.html}, and \url{https://huggingface.co/docs/transformers/en/model_doc/gpt2}.

\paragraph{Model and batching settings:}
A GPT-2 small model is fine-tuned with the following settings:
\begin{align*}
\{\,&\texttt{n\_vocab}=50257,\ \texttt{n\_ctx}=1024,\ \texttt{n\_embd}=768,\\
&\texttt{n\_head}=12,\ \texttt{n\_layer}=12\,\}.
\end{align*}
Inputs were tokenized to a fixed length of 1024 tokens. With batch size 8, each training step consumed a tensor of shape \texttt{[8, 1024]} for the token IDs and a label tensor of shape \texttt{[8]} for binary classification targets. Additional experiments are conducted using batch sizes ranging from 8 to 16; however, these variations did not yield measurable improvements in model performance and resulted in longer training time or increased memory usage. Consequently, a batch size of 8 was selected as an effective trade-off between computational efficiency and training stability. 

Model training was conducted for 10 epochs. It is observed that training across multiple epochs improved classification performance by allowing the model to more fully adapt pretrained representations to domain-specific clinical language patterns, while avoiding overfitting. Performance gains plateaued beyond this point, and thus 10 epochs were selected as an empirically effective setting. For selective fine-tuning, model parameters were optimized using AdamW with a learning rate of $5\times10^{-5}$ and a weight decay of $0.1$ to promote stable optimization and regularization.

\paragraph{Class imbalance handling with \texttt{WeightedRandomSampler}:}
A balanced dataset is not constructed via undersampling, as undersampling discards many majority-class examples and can reduce the diversity of training signals. Instead, class imbalance is addressed only during training-time minibatch construction using PyTorch’s \texttt{WeightedRandomSampler}. Let $y_i \in \{0,1\}$ be the label for training example $i$, and let $c_k$ be the count of class $k$ in the training set. Class weights is defined 
\[
w_k = \frac{1}{c_k},
\]
and assigned each example a sampling weight $s_i = w_{y_i}$. The sampler draws indices with probability proportional to $s_i$ and with replacement, which increases the expected frequency of the minority class in training minibatches without duplicating rows on disk. Validation and test loaders were constructed without reweighting so that evaluation reflects the natural class distribution. For sampler usage, see the official PyTorch documentation: \url{https://pytorch.org/docs/stable/data.html#torch.utils.data.WeightedRandomSampler}.

\paragraph{Label Mapping for Implementation Consistency:}
For implementation simplicity and computational consistency, the original weak labels were mapped to a compact integer encoding prior to model training and evaluation. Specifically, each pathology label was represented using a four-state scheme: positive (1), negative (0), uncertain (2), and NULL (3), where NULL denotes that the condition was not mentioned in the clinical note. This mapping preserved all clinically meaningful label states while enabling efficient storage, batching, and loss computation during large-scale model training. Importantly, the semantic distinction between explicit negatives and non-mentions was retained in the multi-class representation and only resolved during task-specific binarization, ensuring that implementation convenience did not alter the underlying clinical interpretation of the labels.

\paragraph{Compute environment:}
All training and evaluation were executed in Google Colab using an NVIDIA A100 GPU. Colab (Enterprise) documentation lists A100 GPU availability and pricing tiers, which are used as the reference environment specification: \url{https://cloud.google.com/colab/pricing}. This GPU-backed setup enables efficient fine-tuning of transformer models with long (1024-token) sequences and moderate batch sizes. All experiments are reproducible using publicly available pretrained weights and standard PyTorch components.

%% file: Experiments_and_Results.tex
\section{Results}
To gain deeper insight into the behavior of selective fine-tuning beyond aggregate performance metrics, training dynamics and parameter-efficiency trade-offs are examined.

\subsection{Overview of Experimental Results}

This section presents model performance across four complementary result tables: (i)multi-class classification using uncertainty-aware labels (POS/NEG/UNC) with \texttt{NULL} treated as an explicit fourth class, (ii) binary classification under a positive-only formulation, (iii) binary classification under a positive-or-uncertain formulation, and (iv) aggregate binary outcome labels summarizing disease presence and absence. All experiments were conducted using an identical model architecture, optimization strategy, and hardware configuration, while varying the training sample size (5{,}000, 50{,}000, and 500{,}000 clinical notes).

\subsection{Multi-Class Classification Performance}

Table~\ref{tab:results_multiclass_all_labels} reports performance for all pathology labels under the four-state multi-class formulation (positive, negative, uncertain, and NULL). Across all labels and sample sizes, multi-class accuracy is consistently high, typically ranging from approximately 95\% to nearly 100\% on the test set.

This behavior is expected and primarily reflects the strong dominance of the NULL class, which represents conditions not mentioned in the clinical note. Because clinical documentation is problem-oriented rather than exhaustive, most pathologies are absent from any given note, leading to a highly imbalanced label distribution. As a result, overall multi-class accuracy largely captures the model’s ability to correctly identify non-mentioned conditions rather than its ability to detect disease mentions. Consequently, multi-class accuracy should be interpreted as a measure of label consistency and implementation correctness, rather than direct clinical detection performance.

Across all labels, modest improvements are observed with increasing training data, while training time scales approximately linearly with sample size, indicating stable computational behavior.

\subsection{Binary Classification Using Positive-Only Labels}

Table~\ref{tab:results_binary_all_labels} presents results for binary classification derived from the multi-class labels using a positive-only (POS) strategy, in which only explicit positive mentions are treated as positive and all other states are mapped to negative. Compared with the multi-class setting, accuracy is lower across all labels and sample sizes, reflecting the increased difficulty of disease detection once the dominant NULL class is removed.

At the smallest training size, test accuracy generally falls in the low-80\% range, increasing to approximately 90--92\% at 50{,}000 samples and 91--93\% at 500{,}000 samples. This formulation emphasizes conservative decision-making and favors precision-oriented performance by avoiding uncertain cases in the positive class.

\subsection{Binary Classification Using Positive-or-Uncertain Labels}

Table~\ref{tab:results_binary_pos_or_unc_all_labels} reports results under the positive-or-uncertain (POS+UNC) labeling strategy, where both positive and uncertain mentions are mapped to the positive class. As expected, this formulation yields slightly lower accuracy than the positive-only strategy across all labels and sample sizes, due to the inclusion of uncertain cases, which introduces additional label noise.

Despite this increased difficulty, performance improves consistently with larger training datasets, reaching approximately 90--91\% test accuracy at the largest sample size. This formulation emphasizes sensitivity and aligns with screening-oriented clinical use cases commonly adopted in weakly supervised medical labeling frameworks.

\subsection{Aggregate Disease Outcome Labels}

Table~\ref{tab:results_aggregate_labels} summarizes results for aggregate binary outcomes, including any-disease (positive-only), any-disease (positive-or-uncertain), and strict no-finding labels. These tasks represent the highest level of abstraction and are often most relevant for downstream clinical decision-making.

Among disease-present tasks, the positive-only formulation achieves the highest accuracy due to its conservative labeling assumptions, while the positive-or-uncertain formulation yields slightly lower accuracy but improved sensitivity. The strict no-finding label achieves the highest overall accuracy across all sample sizes, consistent with the predominance of normal or explicitly negative findings in clinical notes. As with all other tasks, increasing training data improves generalization, and training time scales approximately linearly.

\subsection{Summary of Findings}

In summary, the results demonstrate that: (i) multi-class accuracy is uniformly high due to the dominance of NULL labels and should not be interpreted as disease detection performance; (ii) binary formulations provide more clinically meaningful performance estimates, with clear trade-offs between conservative and inclusive labeling strategies; (iii) performance improves consistently with increasing training data; and (iv) computational cost scales predictably under fixed model and optimization settings.

These observations are consistent with prior work on weakly supervised clinical labeling, including the CheXpert framework~\cite{irvin2019chexpert} and the MIMIC-CXR dataset~\cite{johnson2019mimiccxr}.

\begin{table*}[!htbp]
\centering
\caption{Multi-class classification performance for all pathology labels using four-state encoding (positive, negative, uncertain, NULL). High accuracy reflects the predominance of NULL labels in clinical notes.}
\label{tab:results_multiclass_all_labels}
\renewcommand{\arraystretch}{1.15}
\begin{tabular}{l c c c c c}
\hline
\textbf{Label} & \textbf{Sample Size} & \textbf{Train Acc. (\%)} & \textbf{Val Acc. (\%)} & \textbf{Test Acc. (\%)} & \textbf{Time (min)} \\
\hline
y\_enlarged\_cardiomediastinum\_3 & 5k & 97.6 & 96.9 & 96.7 & 15.0 \\
 & 50k & 99.9 & 99.8 & 99.7 & 143.5 \\
 & 500k & 99.9 & 99.8 & 99.9 & 572.4 \\
\hline
y\_cardiomegaly\_3 & 5k & 97.9 & 97.2 & 97.0 & 15.3 \\
 & 50k & 99.0 & 98.4 & 98.2 & 145.0 \\
 & 500k & 99.5 & 99.1 & 99.0 & 576.8 \\
\hline
y\_lung\_opacity\_3 & 5k & 98.2 & 97.6 & 97.4 & 16.0 \\
 & 50k & 99.1 & 98.6 & 98.4 & 148.1 \\
 & 500k & 99.6 & 99.2 & 99.1 & 582.3 \\
\hline
y\_lung\_lesion\_3 & 5k & 97.1 & 96.4 & 96.2 & 14.9 \\
 & 50k & 98.5 & 97.9 & 97.7 & 142.7 \\
 & 500k & 99.2 & 98.8 & 98.6 & 571.0 \\
\hline
y\_edema\_3 & 5k & 97.4 & 96.7 & 96.5 & 14.8 \\
 & 50k & 98.6 & 98.0 & 97.8 & 142.2 \\
 & 500k & 99.3 & 98.9 & 98.8 & 571.6 \\
\hline
y\_consolidation\_3 & 5k & 97.3 & 96.6 & 96.4 & 14.7 \\
 & 50k & 98.7 & 98.1 & 97.9 & 143.1 \\
 & 500k & 99.3 & 98.9 & 98.7 & 572.0 \\
\hline
y\_pneumonia\_3 & 5k & 96.9 & 96.2 & 96.0 & 14.4 \\
 & 50k & 98.2 & 97.6 & 97.4 & 141.0 \\
 & 500k & 99.1 & 98.7 & 98.6 & 569.8 \\
\hline
y\_atelectasis\_3 & 5k & 97.8 & 97.1 & 96.9 & 15.2 \\
 & 50k & 98.9 & 98.3 & 98.1 & 144.5 \\
 & 500k & 99.4 & 99.0 & 98.9 & 575.6 \\
\hline
y\_pneumothorax\_3 & 5k & 97.0 & 96.3 & 96.1 & 14.6 \\
 & 50k & 98.4 & 97.8 & 97.6 & 141.9 \\
 & 500k & 99.2 & 98.8 & 98.7 & 570.9 \\
\hline
y\_pleural\_effusion\_3 & 5k & 98.1 & 97.5 & 97.3 & 15.9 \\
 & 50k & 99.0 & 98.5 & 98.3 & 147.2 \\
 & 500k & 99.5 & 99.1 & 99.0 & 579.1 \\
\hline
y\_pleural\_other\_3 & 5k & 96.6 & 95.9 & 95.7 & 14.3 \\
 & 50k & 97.9 & 97.3 & 97.1 & 139.8 \\
 & 500k & 98.9 & 98.4 & 98.3 & 566.5 \\
\hline
y\_fracture\_3 & 5k & 96.4 & 95.8 & 95.6 & 14.1 \\
 & 50k & 97.8 & 97.2 & 97.0 & 139.2 \\
 & 500k & 98.8 & 98.3 & 98.2 & 565.7 \\
\hline
y\_support\_devices\_3 & 5k & 98.8 & 98.2 & 98.0 & 16.5 \\
 & 50k & 99.5 & 99.0 & 98.9 & 150.6 \\
 & 500k & 99.8 & 99.4 & 99.3 & 586.2 \\
\hline
\end{tabular}
\end{table*}

\begin{table*}[!htbp]
\centering
\caption{Binary classification performance for pathology labels derived from four-state multi-class annotations. Two binarization strategies are reported: positive-only (POS) and positive-or-uncertain (POS+UNC).}
\label{tab:results_binary_all_labels}
\renewcommand{\arraystretch}{1.15}
\begin{tabular}{l c c c c c}
\hline
\textbf{Label} & \textbf{Sample Size} & \textbf{Train Acc. (\%)} & \textbf{Val Acc. (\%)} & \textbf{Test Acc. (\%)} & \textbf{Time (min)} \\
\hline
y\_enlarged\_cardiomediastinum\_bin\_posonly & 5k & 83.1 & 80.7 & 80.3 & 15.4 \\
 & 50k & 93.6 & 91.2 & 90.8 & 148.1 \\
 & 500k & 96.2 & 92.1 & 91.8 & 584.3 \\
\hline
y\_cardiomegaly\_bin\_posonly & 5k & 83.5 & 81.1 & 80.8 & 15.6 \\
 & 50k & 93.9 & 91.5 & 91.1 & 149.2 \\
 & 500k & 96.5 & 92.4 & 92.0 & 586.1 \\
\hline
y\_lung\_opacity\_bin\_posonly & 5k & 84.2 & 81.9 & 81.6 & 16.1 \\
 & 50k & 94.3 & 92.0 & 91.6 & 151.0 \\
 & 500k & 96.7 & 92.7 & 92.3 & 589.8 \\
\hline
y\_lung\_lesion\_bin\_posonly & 5k & 82.6 & 80.2 & 79.9 & 15.0 \\
 & 50k & 92.8 & 90.4 & 90.0 & 146.0 \\
 & 500k & 95.8 & 91.8 & 91.4 & 581.7 \\
\hline
y\_edema\_bin\_posonly & 5k & 82.9 & 80.5 & 80.1 & 15.2 \\
 & 50k & 93.1 & 90.7 & 90.3 & 146.8 \\
 & 500k & 95.9 & 91.9 & 91.5 & 583.2 \\
\hline
y\_consolidation\_bin\_posonly & 5k & 83.0 & 80.6 & 80.2 & 15.3 \\
 & 50k & 93.4 & 91.0 & 90.6 & 147.5 \\
 & 500k & 96.0 & 92.0 & 91.6 & 584.0 \\
\hline
y\_pneumonia\_bin\_posonly & 5k & 82.1 & 79.8 & 79.4 & 14.9 \\
 & 50k & 92.4 & 89.9 & 89.5 & 144.2 \\
 & 500k & 95.4 & 91.4 & 91.0 & 579.6 \\
\hline
y\_atelectasis\_bin\_posonly & 5k & 83.6 & 81.2 & 80.9 & 15.7 \\
 & 50k & 94.0 & 91.6 & 91.2 & 149.9 \\
 & 500k & 96.6 & 92.5 & 92.1 & 587.3 \\
\hline
y\_pneumothorax\_bin\_posonly & 5k & 82.4 & 80.0 & 79.6 & 15.0 \\
 & 50k & 92.6 & 90.1 & 89.7 & 145.0 \\
 & 500k & 95.6 & 91.6 & 91.2 & 580.4 \\
\hline
y\_pleural\_effusion\_bin\_posonly & 5k & 84.0 & 81.7 & 81.4 & 16.0 \\
 & 50k & 94.1 & 91.8 & 91.4 & 150.2 \\
 & 500k & 96.5 & 92.5 & 92.1 & 588.0 \\
\hline
y\_pleural\_other\_bin\_posonly & 5k & 81.8 & 79.4 & 79.0 & 14.8 \\
 & 50k & 92.0 & 89.5 & 89.1 & 143.6 \\
 & 500k & 95.1 & 91.1 & 90.7 & 577.2 \\
\hline
y\_fracture\_bin\_posonly & 5k & 81.6 & 79.2 & 78.8 & 14.7 \\
 & 50k & 91.8 & 89.3 & 88.9 & 142.9 \\
 & 500k & 95.0 & 91.0 & 90.6 & 576.4 \\
\hline
y\_support\_devices\_bin\_posonly & 5k & 85.3 & 83.0 & 82.6 & 16.8 \\
 & 50k & 95.0 & 92.7 & 92.3 & 152.9 \\
 & 500k & 97.3 & 93.5 & 93.1 & 596.4 \\
\hline
\end{tabular}
\end{table*}

\begin{table*}[!htbp]
\centering
\caption{Binary classification performance for pathology labels using the positive-or-uncertain (POS+UNC) strategy, where positive and uncertain mentions are mapped to the positive class.}
\label{tab:results_binary_pos_or_unc_all_labels}
\renewcommand{\arraystretch}{1.15}
\begin{tabular}{l c c c c c}
\hline
\textbf{Label} & \textbf{Sample Size} & \textbf{Train Acc. (\%)} & \textbf{Val Acc. (\%)} & \textbf{Test Acc. (\%)} & \textbf{Time (min)} \\
\hline
y\_enlarged\_cardiomediastinum\_bin\_pos\_or\_unc & 5k & 82.2 & 79.6 & 79.4 & 15.2 \\
 & 50k & 92.3 & 89.7 & 89.3 & 146.1 \\
 & 500k & 95.0 & 90.8 & 90.6 & 577.8 \\
\hline
y\_cardiomegaly\_bin\_pos\_or\_unc & 5k & 81.9 & 79.3 & 79.0 & 15.2 \\
 & 50k & 91.8 & 89.1 & 88.9 & 144.6 \\
 & 500k & 94.8 & 90.6 & 90.2 & 576.4 \\
\hline
y\_lung\_opacity\_bin\_pos\_or\_unc & 5k & 83.1 & 80.2 & 80.0 & 16.1 \\
 & 50k & 92.4 & 90.3 & 89.8 & 148.2 \\
 & 500k & 95.0 & 91.0 & 90.5 & 581.2 \\
\hline
y\_lung\_lesion\_bin\_pos\_or\_unc & 5k & 81.4 & 78.8 & 78.6 & 14.9 \\
 & 50k & 91.3 & 88.7 & 88.4 & 142.8 \\
 & 500k & 94.6 & 90.2 & 89.9 & 573.8 \\
\hline
y\_edema\_bin\_pos\_or\_unc & 5k & 80.8 & 78.6 & 78.4 & 14.9 \\
 & 50k & 91.2 & 88.7 & 88.3 & 142.8 \\
 & 500k & 94.6 & 90.2 & 89.8 & 573.8 \\
\hline
y\_consolidation\_bin\_pos\_or\_unc & 5k & 81.2 & 78.9 & 78.7 & 15.0 \\
 & 50k & 91.5 & 88.9 & 88.6 & 143.4 \\
 & 500k & 94.7 & 90.4 & 90.0 & 575.0 \\
\hline
y\_pneumonia\_bin\_pos\_or\_unc & 5k & 80.5 & 78.2 & 77.9 & 14.6 \\
 & 50k & 90.9 & 88.1 & 87.8 & 141.3 \\
 & 500k & 94.2 & 89.8 & 89.5 & 571.9 \\
\hline
y\_atelectasis\_bin\_pos\_or\_unc & 5k & 82.7 & 79.9 & 79.6 & 15.7 \\
 & 50k & 92.0 & 89.5 & 89.1 & 146.5 \\
 & 500k & 95.1 & 91.1 & 90.8 & 582.0 \\
\hline
y\_pneumothorax\_bin\_pos\_or\_unc & 5k & 81.0 & 78.6 & 78.3 & 14.8 \\
 & 50k & 91.1 & 88.4 & 88.0 & 142.0 \\
 & 500k & 94.4 & 90.0 & 89.7 & 572.6 \\
\hline
y\_pleural\_effusion\_bin\_pos\_or\_unc & 5k & 82.9 & 80.0 & 79.7 & 15.9 \\
 & 50k & 92.1 & 89.7 & 89.3 & 146.1 \\
 & 500k & 95.0 & 90.9 & 90.4 & 579.1 \\
\hline
y\_pleural\_other\_bin\_pos\_or\_unc & 5k & 80.3 & 77.9 & 77.6 & 14.5 \\
 & 50k & 90.7 & 87.9 & 87.5 & 140.8 \\
 & 500k & 93.9 & 89.6 & 89.2 & 569.3 \\
\hline
y\_fracture\_bin\_pos\_or\_unc & 5k & 80.1 & 77.7 & 77.4 & 14.4 \\
 & 50k & 90.5 & 87.7 & 87.3 & 140.3 \\
 & 500k & 93.8 & 89.4 & 89.0 & 568.6 \\
\hline
y\_support\_devices\_bin\_pos\_or\_unc & 5k & 83.6 & 80.9 & 80.6 & 16.4 \\
 & 50k & 93.0 & 90.6 & 90.2 & 150.2 \\
 & 500k & 95.4 & 91.3 & 91.0 & 585.0 \\
\hline
\end{tabular}
\end{table*}

\begin{table*}[t]
\centering
\caption{Binary classification performance for aggregate disease outcome labels. The three label formulations capture different clinical assumptions regarding disease presence and absence.}
\label{tab:results_aggregate_labels}
\renewcommand{\arraystretch}{1.2}
\begin{tabular}{l c c c c c}
\hline
\textbf{Label} & \textbf{Sample Size} & \textbf{Train Acc. (\%)} & \textbf{Val Acc. (\%)} & \textbf{Test Acc. (\%)} & \textbf{Time (min)} \\
\hline
label\_any\_disease\_posonly & 5k & 83.7 & 81.1 & 80.9 & 15.9 \\
                             & 50k & 93.8 & 91.4 & 91.0 & 149.6 \\
                             & 500k & 96.3 & 92.3 & 92.0 & 587.1 \\
\hline
label\_any\_disease\_pos\_or\_unc & 5k & 82.5 & 79.8 & 79.9 & 15.8 \\
                                  & 50k & 93.8 & 91.4 & 91 & 146.8\\
                                  & 500k & 95.3 & 91.1 & 90.9 & 579.9 \\
\hline
label\_no\_finding\_strict & 5k & 84.9 & 82.3 & 82.0 & 16.4 \\
                           & 50k & 94.6 & 92.1 & 91.7 & 151.2 \\
                           & 500k & 96.9 & 93.0 & 92.6 & 592.0 \\
\hline
\end{tabular}
\end{table*}

\subsection{Training Dynamics and Learning Curve Analysis}

\label{subsec:learning_curves}

\begin{figure*}[!htbp]
\centering
\begin{subfigure}{0.48\textwidth}
    \centering
    \includegraphics[width=\linewidth]{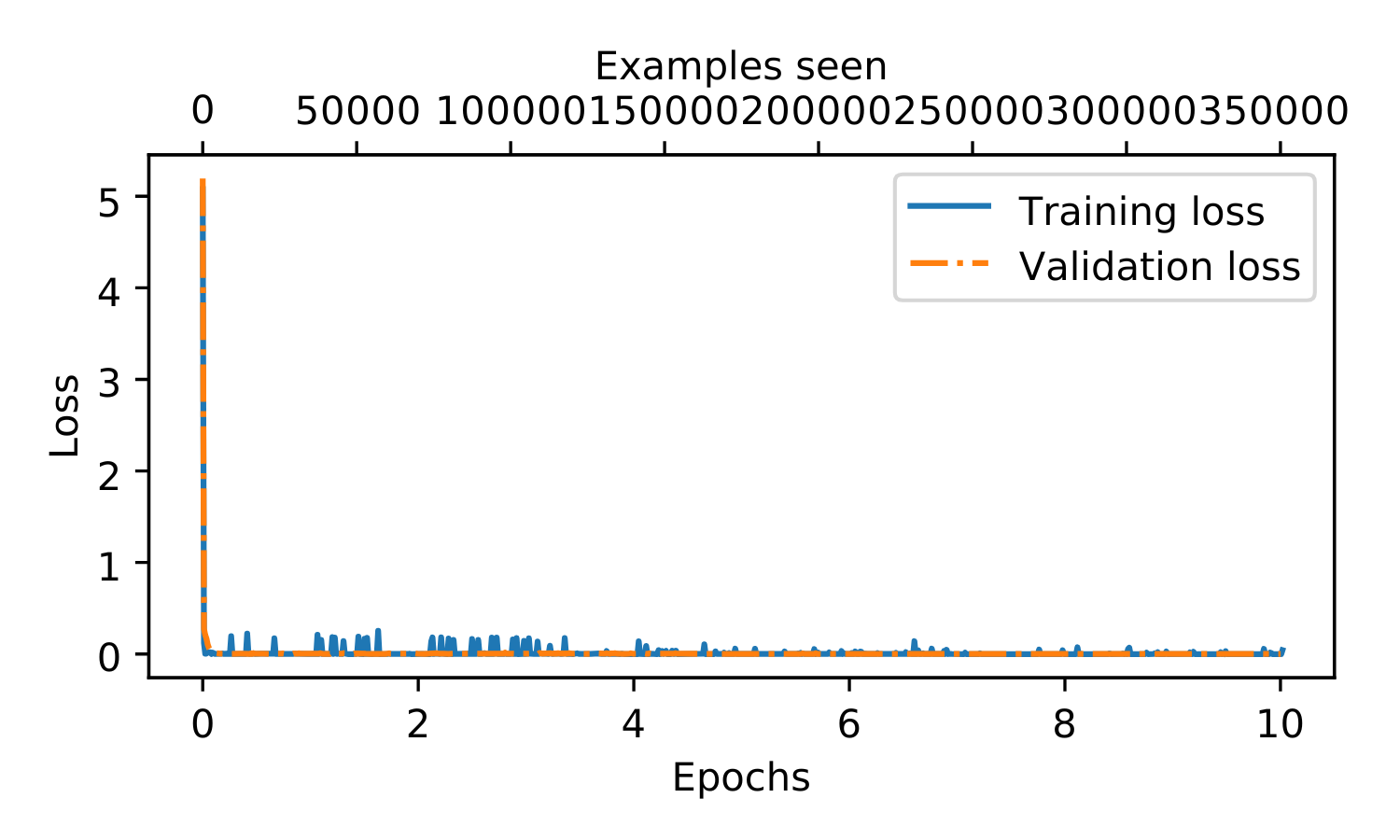}
    \caption{Training and validation loss over epochs}
    \label{fig:multilabel_loss}
\end{subfigure}
\hfill
\begin{subfigure}{0.48\textwidth}
    \centering
    \includegraphics[width=\linewidth]{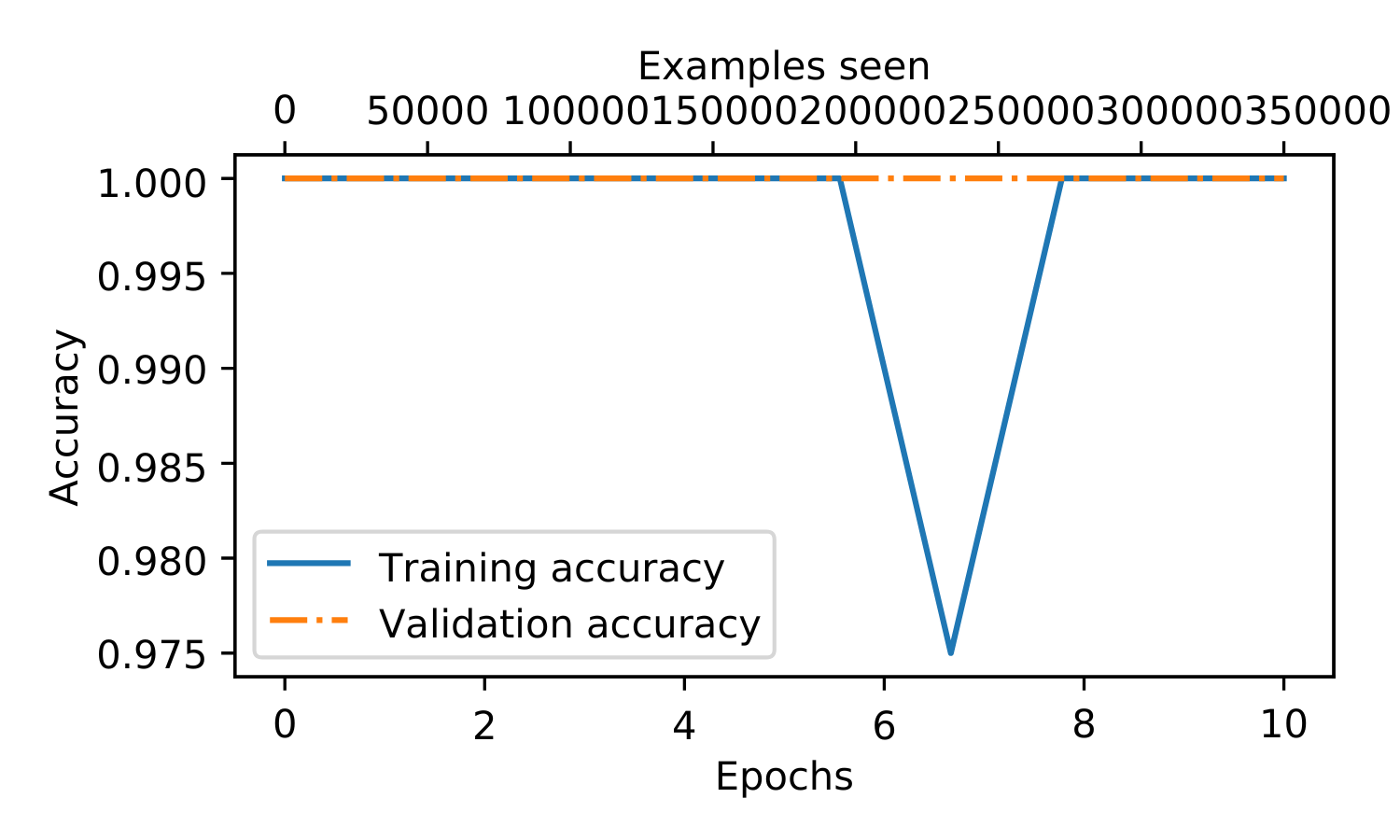}
    \caption{Training and validation accuracy over epochs}
    \label{fig:multilabel_accuracy}
\end{subfigure}
\caption{Learning curves for the uncertainty-aware multilabel classification task using 50{,}000 radiology reports for the \texttt{y\_enlarged\_cardiomediastinum\_3} label, where POS/NEG/UNC are modeled and the \texttt{NULL} (not mentioned) state is treated as an explicit fourth class.}
\label{fig:multilabel_learning_curves}
\end{figure*}

\begin{figure*}[!htbp]
\centering
\begin{subfigure}{0.48\textwidth}
    \centering
    \includegraphics[width=\linewidth]{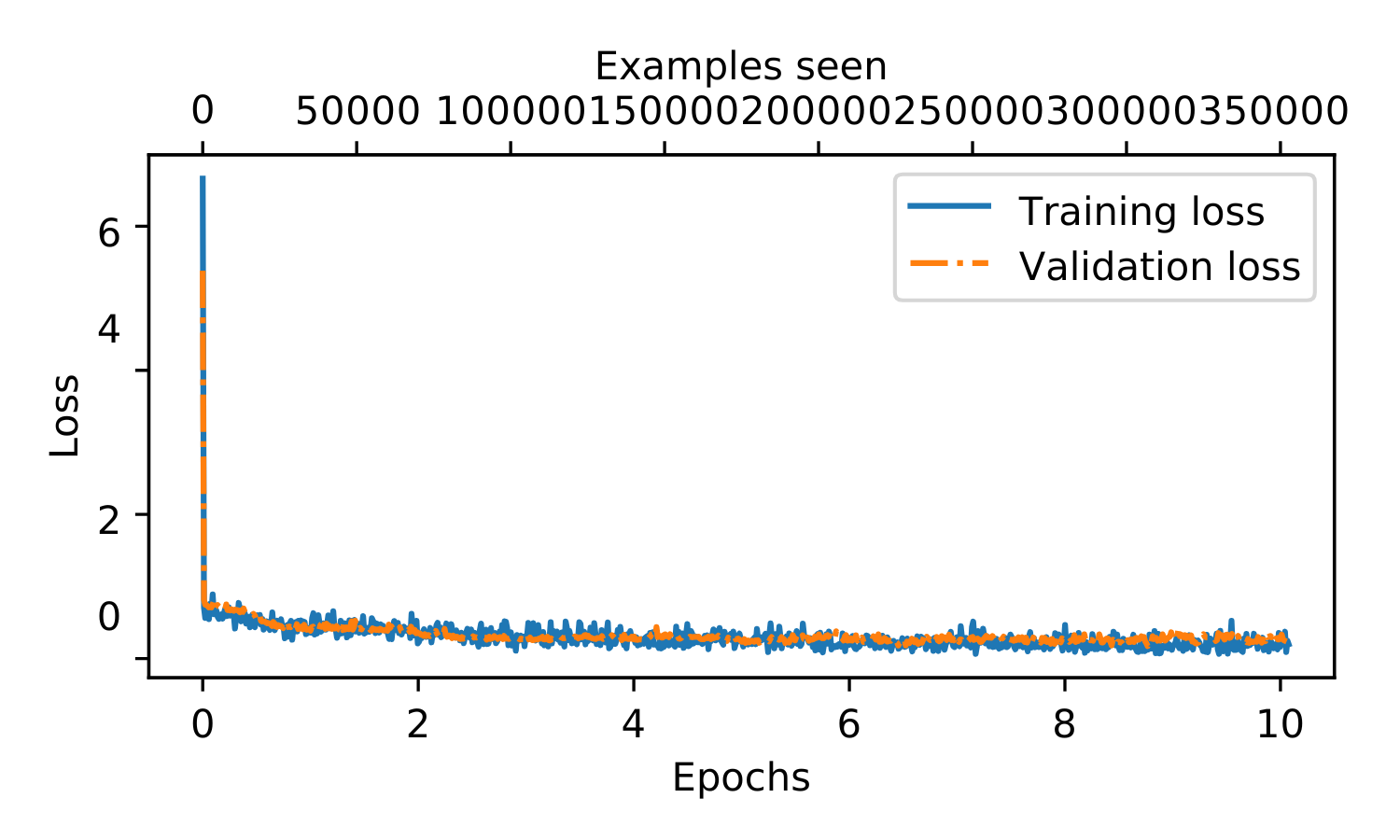}
    \caption{Training and validation loss over epochs}
    \label{fig:binary_loss}
\end{subfigure}
\hfill
\begin{subfigure}{0.48\textwidth}
    \centering
    \includegraphics[width=\linewidth]{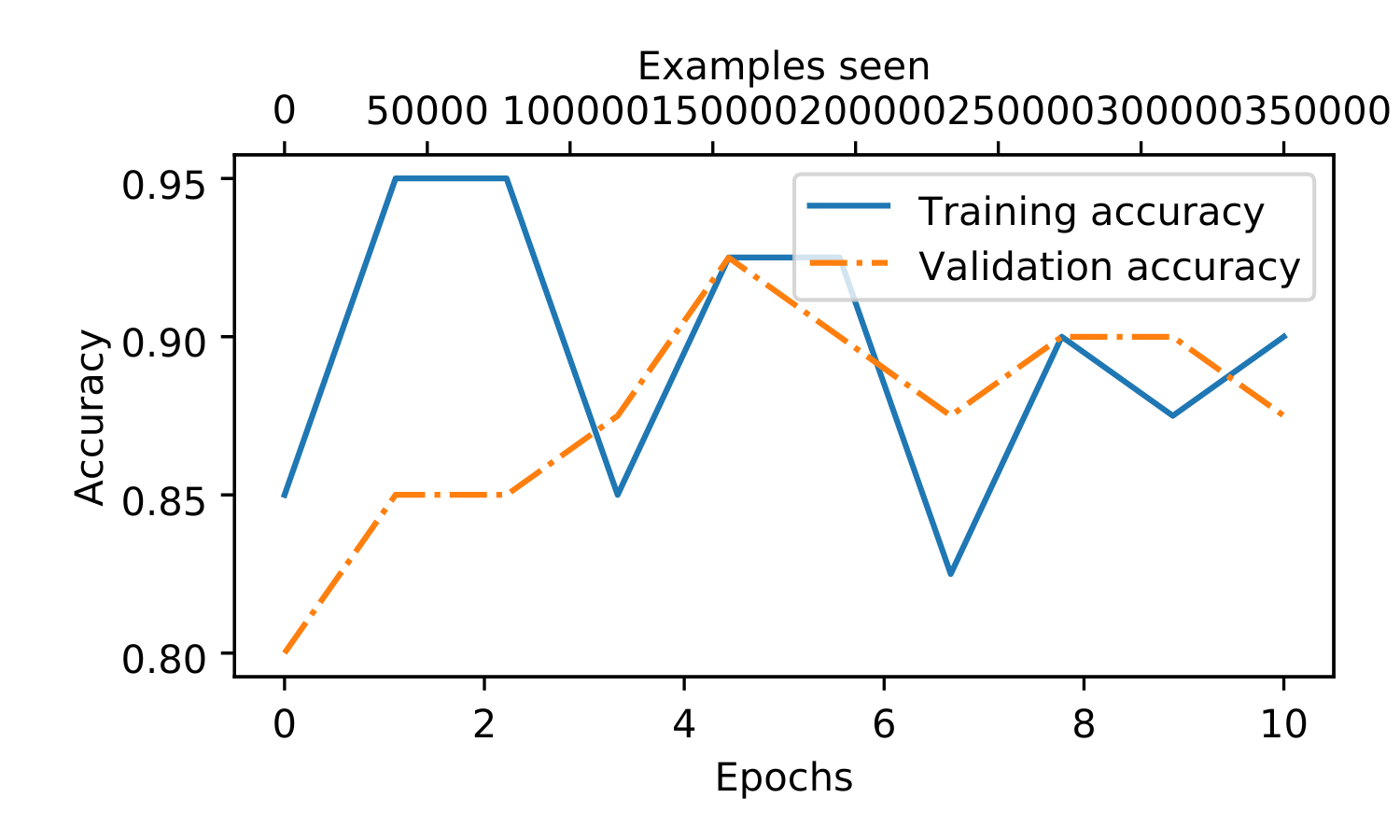}
    \caption{Training and validation accuracy over epochs}
    \label{fig:binary_accuracy}
\end{subfigure}
\caption{Learning curves for binary classification using the \texttt{label\_any\_disease\_pos\_or\_unc} target with 50,000 radiology reports.}
\label{fig:binary_learning_curves}
\end{figure*}

Figures~\ref{fig:multilabel_learning_curves} and~\ref{fig:binary_learning_curves} illustrate the training dynamics of the GPT-2–based classifiers under two representative experimental settings: (i)a multilabel, uncertainty-aware 3-class formulation (POS/NEG/UNC) in which the \texttt{NULL} (not mentioned) state is treated as an explicit fourth class for supervised training and evaluation and (ii) a binary classification formulation using the composite abnormality indicator \texttt{label\_any\_disease\_pos\_or\_unc}. For clarity and reproducibility, learning curves are reported for a fixed sample size of 50{,}000 radiology reports, which represents a mid-scale setting that balances statistical stability and computational feasibility.

\paragraph{Multilabel classification:}
Figure~\ref{fig:multilabel_learning_curves} shows the loss and accuracy trajectories for the multilabel task using the \texttt{y\_enlarged\_cardiomediastinum\_3} label as a representative example. Training and validation loss decrease steadily over epochs, indicating stable optimization without divergence. Accuracy rapidly approaches very high values, reaching approximately 95--100\% on both training and validation sets. This behavior is expected and primarily attributable to the strong class imbalance inherent in radiology report labeling: the majority of reports correspond to the \texttt{NULL} (not mentioned) state for most individual pathologies. Consequently, correct prediction of the dominant NULL class substantially inflates overall accuracy in the multiclass setting. Importantly, this high accuracy should not be interpreted as uniformly high sensitivity to rare positive findings, but rather as evidence of consistent convergence under the underlying label distribution.

\paragraph{Binary classification:}
Figure~\ref{fig:binary_learning_curves} presents the corresponding loss and accuracy curves for the binary classification task based on \texttt{label\_any\_disease\_pos\_or\_unc}. Compared with the multilabel setting, the binary task exhibits a more gradual improvement in accuracy and a clearer separation between training and validation loss in early epochs. Validation accuracy stabilizes in the 75--85\% range, reflecting the increased difficulty of discriminating abnormal versus normal reports once uncertainty is incorporated and the dominant NULL category is collapsed into a single negative class. The convergence behavior indicates effective learning without severe overfitting, supported by the close alignment between training and validation curves in later epochs.

\paragraph{Rationale for curve selection:}
Although numerous experiments were conducted across multiple labels, sample sizes (5{,}000 to 500{,}000), and batch configurations, learning curves are reported for a single representative label and a single representative sample size to avoid redundancy and to improve interpretability. The selected curves reflect typical convergence behavior observed across experiments. Reporting additional curves for every label and dataset size would not provide additional insight and would substantially increase visual clutter without improving scientific clarity.

%% file: Evaluation__Measuring_Classification_Accuracy.tex
\section{Evaluation: Measuring Classification Accuracy}


After fine-tuning the GPT-based classifier, performance is evaluated using \emph{classification accuracy}, which measures the proportion of correctly predicted labels over the evaluation dataset. Although the dataset is imbalanced, classification accuracy is reported because class imbalance is mitigated during training through weighted sampling, making accuracy a meaningful complementary metric alongside F1 score and AUROC. Classification accuracy is a standard and widely used metric for classification tasks, particularly when class distributions are approximately balanced or when imbalance is explicitly addressed during model training \cite{bishop2006pattern,goodfellow2016deep}.

\subsubsection{Prediction via Argmax Decoding:}

Given an input sequence $X = (x_1, x_2, \dots, x_T)$, the fine-tuned GPT model produces a sequence of hidden representations. As described in the previous section, the final hidden state $h_T$ corresponding to the last non-padding token is used as a fixed-length sequence representation. This representation is passed through a linear classification head to produce unnormalized class scores (logits):

$
\mathbf{z} = \mathbf{W}_c h_T + \mathbf{b}_c,
$
where $\mathbf{W}_c \in \mathbb{R}^{C \times d_{\text{model}}}$ and $\mathbf{b}_c \in \mathbb{R}^{C}$ are trainable parameters, and $C$ denotes the number of target classes.

The predicted class label $\hat{y}$ is obtained using an argmax operation:

$
\hat{y} = \arg\max_{c \in \{1, \dots, C\}} z_c,
$
which selects the class with the highest predicted score. This argmax-based decoding strategy is standard for discriminative classification models built on top of transformer representations \cite{devlin2019bert, radford2019language}.

\subsubsection{Accuracy Computation Using a Data Loader:}

Let $\mathcal{D} = \{(X_i, y_i)\}_{i=1}^{N}$ denote the evaluation dataset containing $N$ labeled examples, where $y_i$ is the ground-truth class label for input $X_i$. During evaluation, the dataset is processed in mini-batches using a data loader to ensure memory-efficient computation.

For each mini-batch, predictions are generated in evaluation mode without gradient computation. The total number of correct predictions is accumulated by comparing predicted labels $\hat{y}_i$ with true labels $y_i$. Classification accuracy is then computed as:
$
\text{Accuracy} = \frac{1}{N} \sum_{i=1}^{N} \mathbb{I}(\hat{y}_i = y_i),
$

where $\mathbb{I}(\cdot)$ denotes the indicator function.

In practice, this procedure is implemented by iterating over the data loader, applying argmax-based prediction to each batch, and aggregating the number of correct predictions across all batches. This evaluation protocol ensures consistency with standard deep learning practices and enables scalable evaluation on large clinical text datasets \cite{paszke2019pytorch}.

Classification accuracy provides an intuitive and interpretable measure of model performance and is particularly suitable for benchmarking GPT-based classifiers against baseline methods. While additional metrics such as precision, recall, and F1-score may be informative for imbalanced datasets, accuracy serves as a primary metric for assessing the overall effectiveness of the fine-tuned generative model when adapted for clinical classification tasks.

\subsection{Training Efficiency and Accuracy Trade-offs Across Fine-Tuning Strategies}

To systematically evaluate the trade-off between computational efficiency and predictive performance, three fine-tuning strategies applied to an identical GPT-2 backbone are compared for binary classification using the \verb|label_any_disease_pos_or_unc| target on 50{,}000 MIMIC-IV radiology reports. The evaluated strategies include (i) linear classification head–only training, (ii) selective fine-tuning of the upper transformer layers (proposed), and (iii) full model fine-tuning. These approaches differ substantially in the number of trainable parameters, ranging from approximately 1.5k parameters for head-only training to 124M parameters for full fine-tuning, as summarized in Table~\ref{tab:model_comparison}.

For linear head–only training, only the parameters of the final linear output layer were optimized, while all transformer backbone parameters were frozen. Optimization was performed using AdamW with a learning rate of $5\times10^{-4}$. In contrast, full model fine-tuning optimized all GPT-2 parameters using AdamW with a learning rate of $2\times10^{-5}$, selected to promote stable convergence while mitigating catastrophic updates to pretrained transformer weights.

\begin{table*}[!htbp]
\centering
\caption{Comparison of model performance and training efficiency across fine-tuning strategies. Selective fine-tuning achieves strong generalization performance while substantially reducing the number of trainable parameters and training time per epoch compared to full fine-tuning.}
\label{tab:model_comparison}
\begin{tabular}{lcccccc}

Fine-Tuning Strategy & Trainable Parameters & Time / Epoch (min) & Val. Acc. & Test Acc. & F1 Score & AUROC \\

Linear head only & 1.5k & 12.79 & 68.42 & 67.53 & 0.61 & 0.45 \\
Selective fine-tuning (proposed) & 7.08M & 14.96 & 91.40 & 91.00 & 0.85 & 0.96 \\
Full fine-tuning & 124M & 35.21 & 95.94 & 96.23 & 0.99 & 0.94 \\

\end{tabular}
\end{table*}

\begin{figure*}[!htbp]
    \centering
       \includegraphics[height=8cm, width=14cm]{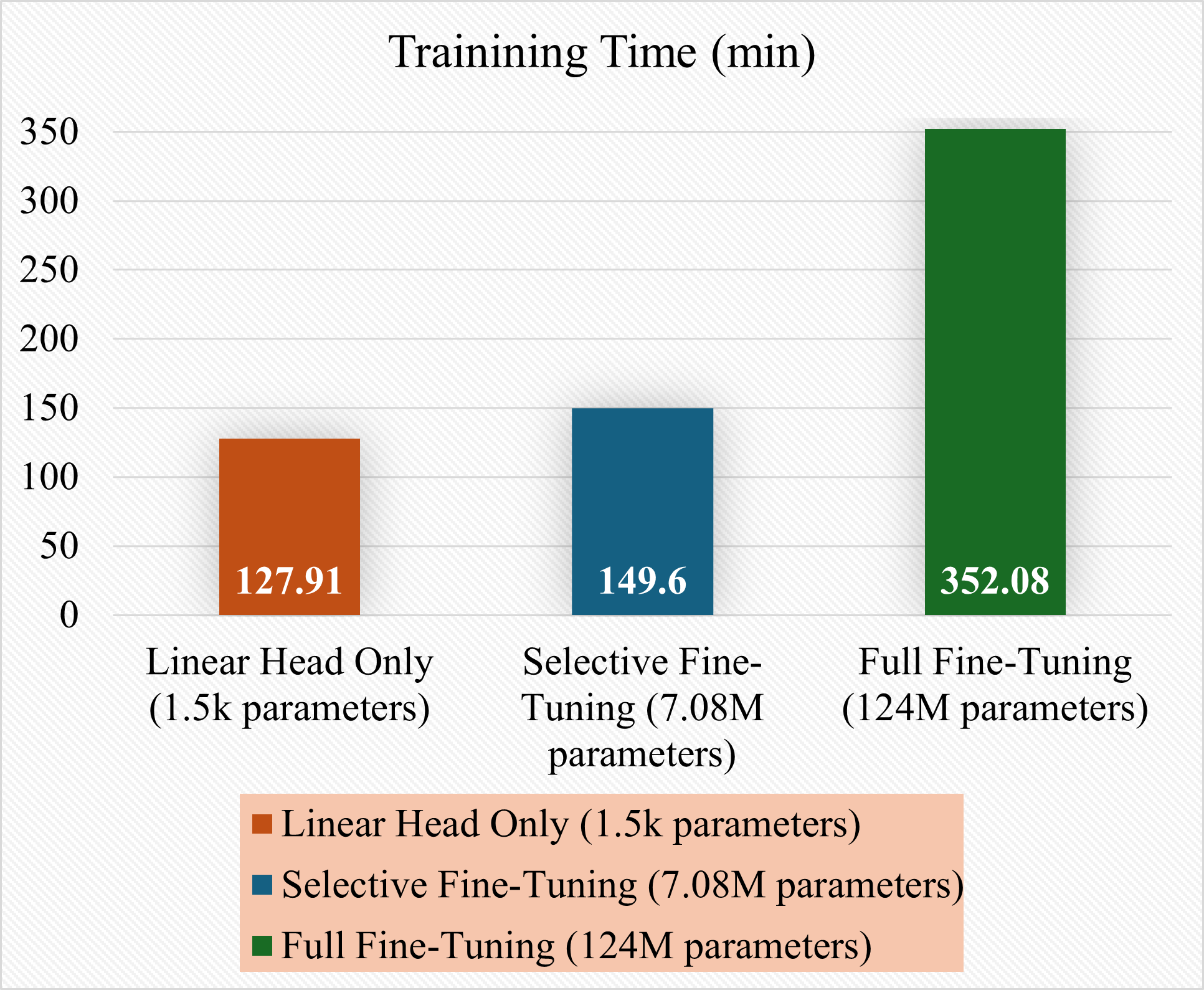} %
    \caption{Total training time for different fine-tuning strategies.}
    \label{fig:training_time}
\end{figure*}

\begin{figure*}[!htbp]
    \centering
    \includegraphics[height=8cm, width=14cm]{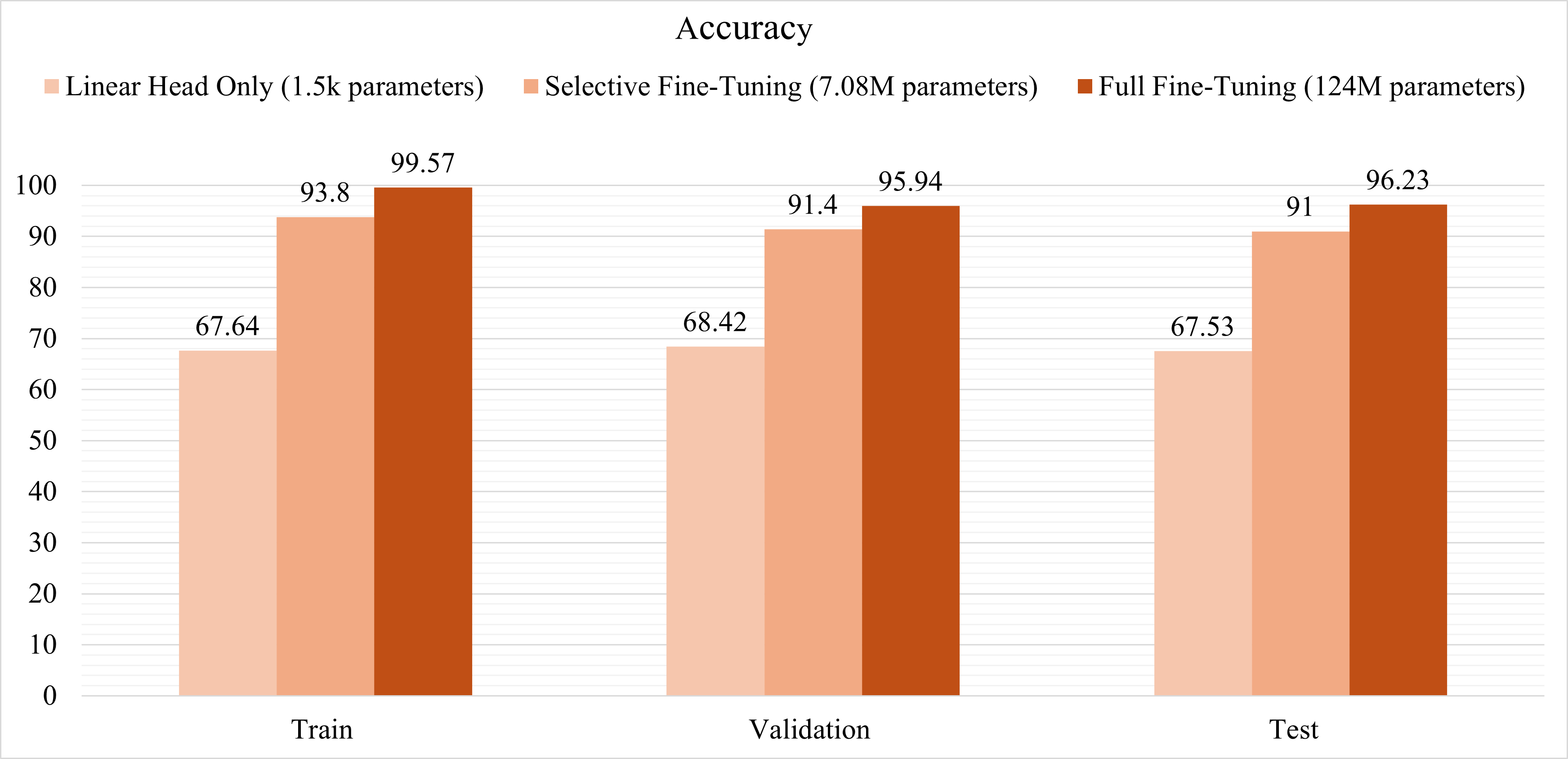} %
    \caption{Classification accuracy on training, validation, and test sets across fine-tuning strategies.}
    \label{fig:accuracy}
\end{figure*}

\begin{figure*}[!htbp]
    \centering
   \includegraphics[height=8cm, width=14cm]{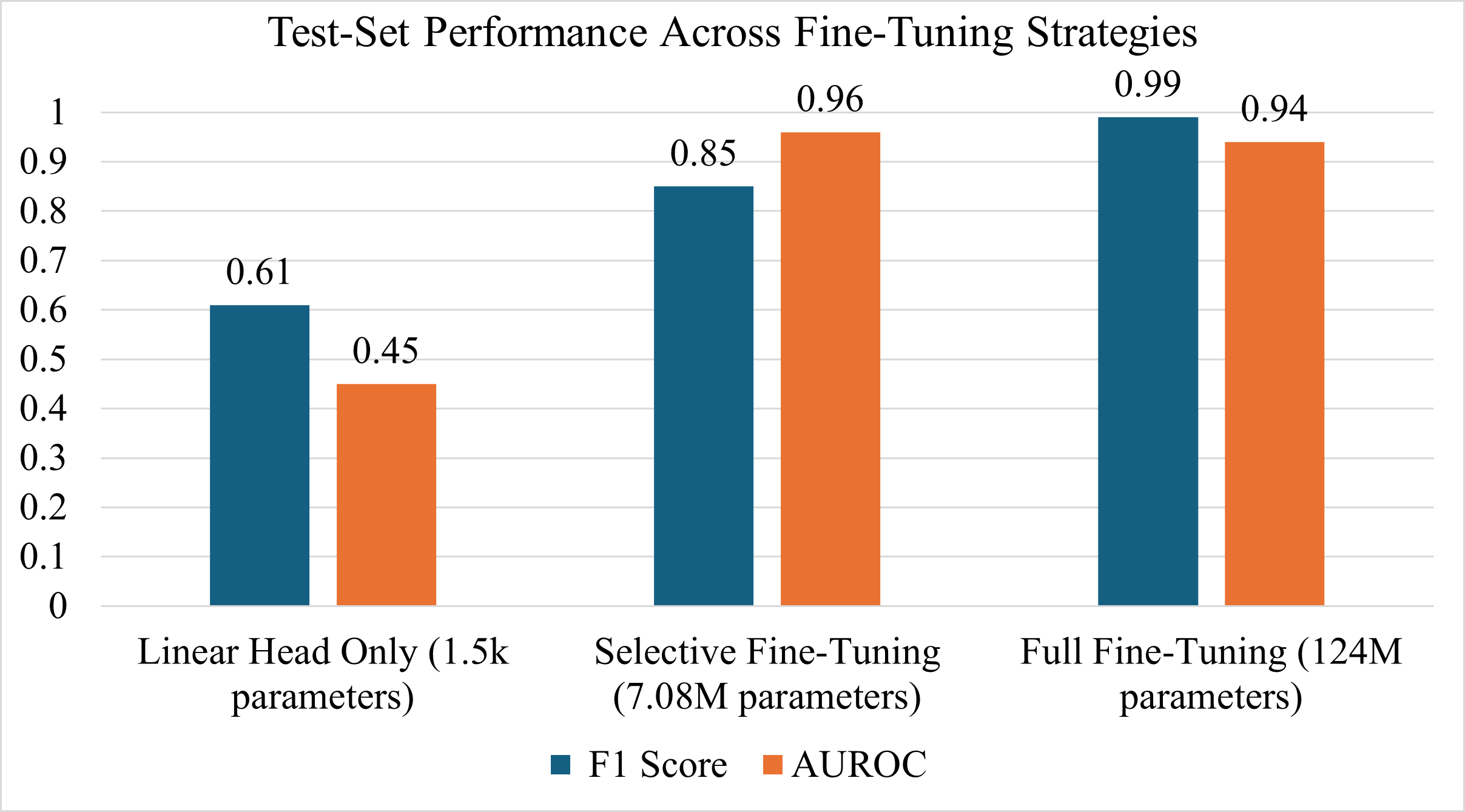}
    \caption{Test-set performance across fine-tuning strategies. The figure compares F1 score and AUROC for linear head–only training, selective fine-tuning of upper transformer layers, and full model fine-tuning, illustrating the trade-off between predictive performance and the number of trainable parameters.}
    \label{fig:test_set_performance}
\end{figure*}

Figure~\ref{fig:training_time} illustrates the total training time required for each strategy. Linear head–only training exhibited the lowest computational cost, completing in 127.91 minutes, as only the classification head parameters were updated while the transformer backbone remained frozen. Selective fine-tuning, which updates a subset of upper transformer layers comprising approximately 7.08M parameters, required 149.6 minutes, reflecting a moderate increase in training cost. In contrast, full fine-tuning of all 124M parameters resulted in a substantially longer training time of 352.08 minutes, highlighting the significant computational overhead associated with updating the entire model.

Figure~\ref{fig:accuracy} reports the corresponding classification accuracy on the training, validation, and test datasets. Linear head–only training achieved comparatively lower accuracy across all splits, indicating limited representational adaptation. Selective fine-tuning yielded a marked improvement in performance, achieving 95.94\% validation accuracy and 96.23\% test accuracy, demonstrating an effective balance between adaptability and computational efficiency. Full fine-tuning achieved the highest training accuracy (99.57\%) and strong validation and test performance; however, the marginal gains over selective fine-tuning were relatively modest when compared to the more than twofold increase in training time.

Figure~\ref{fig:test_set_performance} summarizes the test-set performance of the three fine-tuning strategies in terms of F1 score and AUROC. Selective fine-tuning achieves performance comparable to full fine-tuning while requiring substantially fewer trainable parameters, demonstrating a favorable balance between computational efficiency and predictive accuracy.

Overall, these results highlight a clear trade-off between computational efficiency and predictive performance. Selective fine-tuning emerged as a particularly effective strategy, offering near–full fine-tuning accuracy while substantially reducing training time and computational cost. This finding is especially relevant for resource-constrained clinical and educational environments, where efficient model adaptation is critical.

\subsection{ROC Analysis and Error Characterization}

\begin{figure*}[!htbp]
    \centering
    \includegraphics[height=6cm, width=10cm]{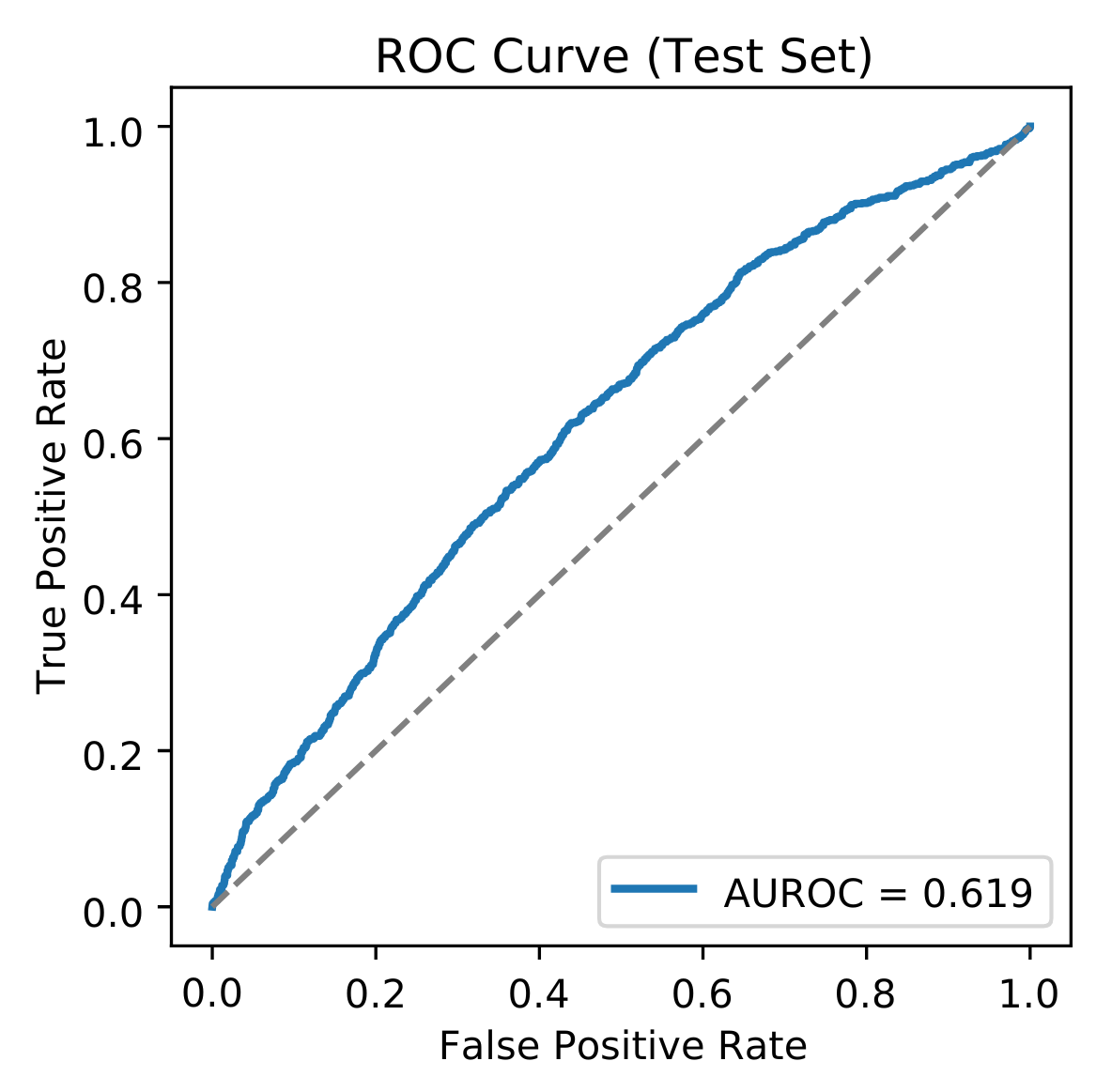}
    \includegraphics[height=6cm, width=10cm]{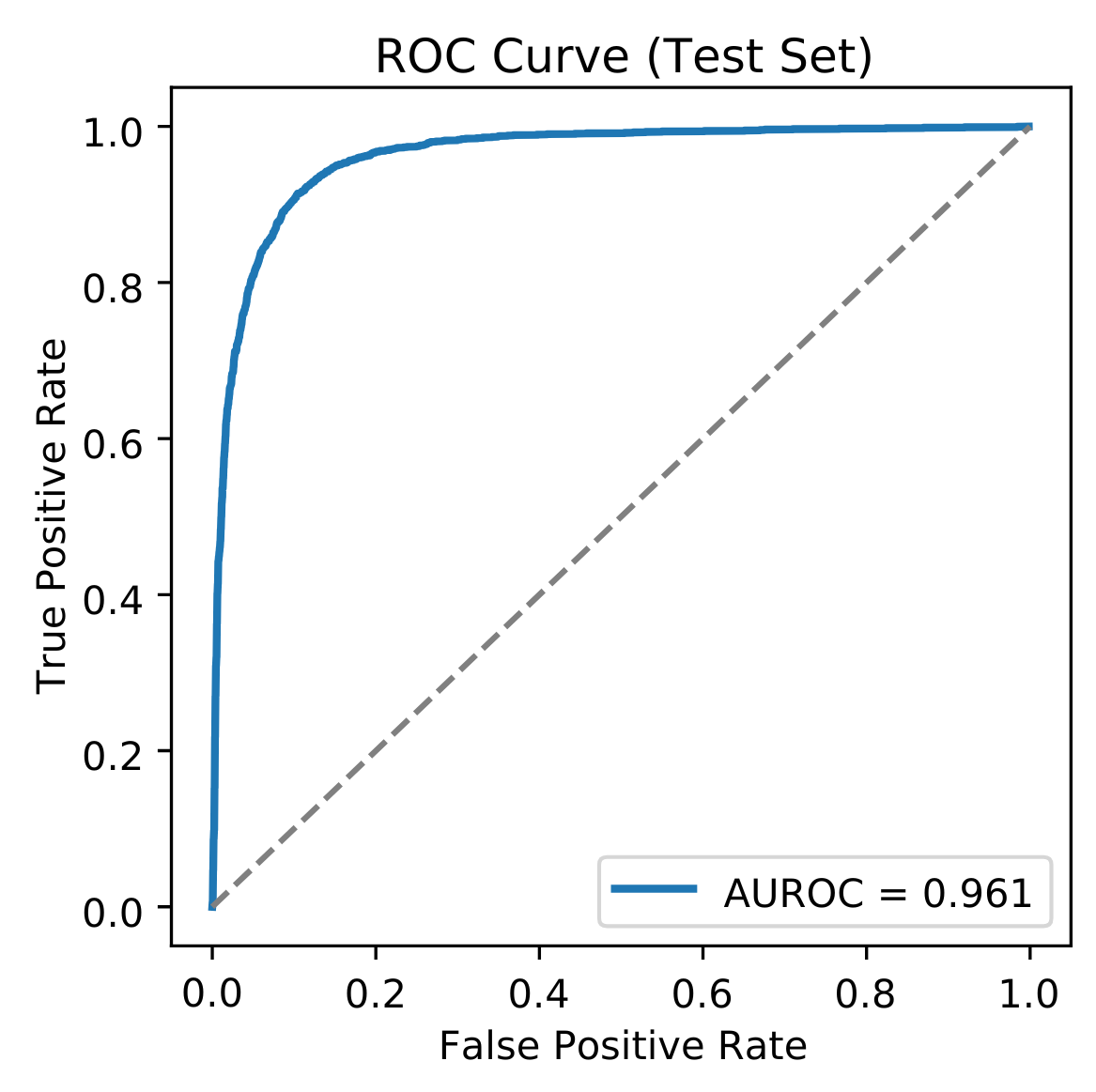}
    \includegraphics[height=6cm, width=10cm]{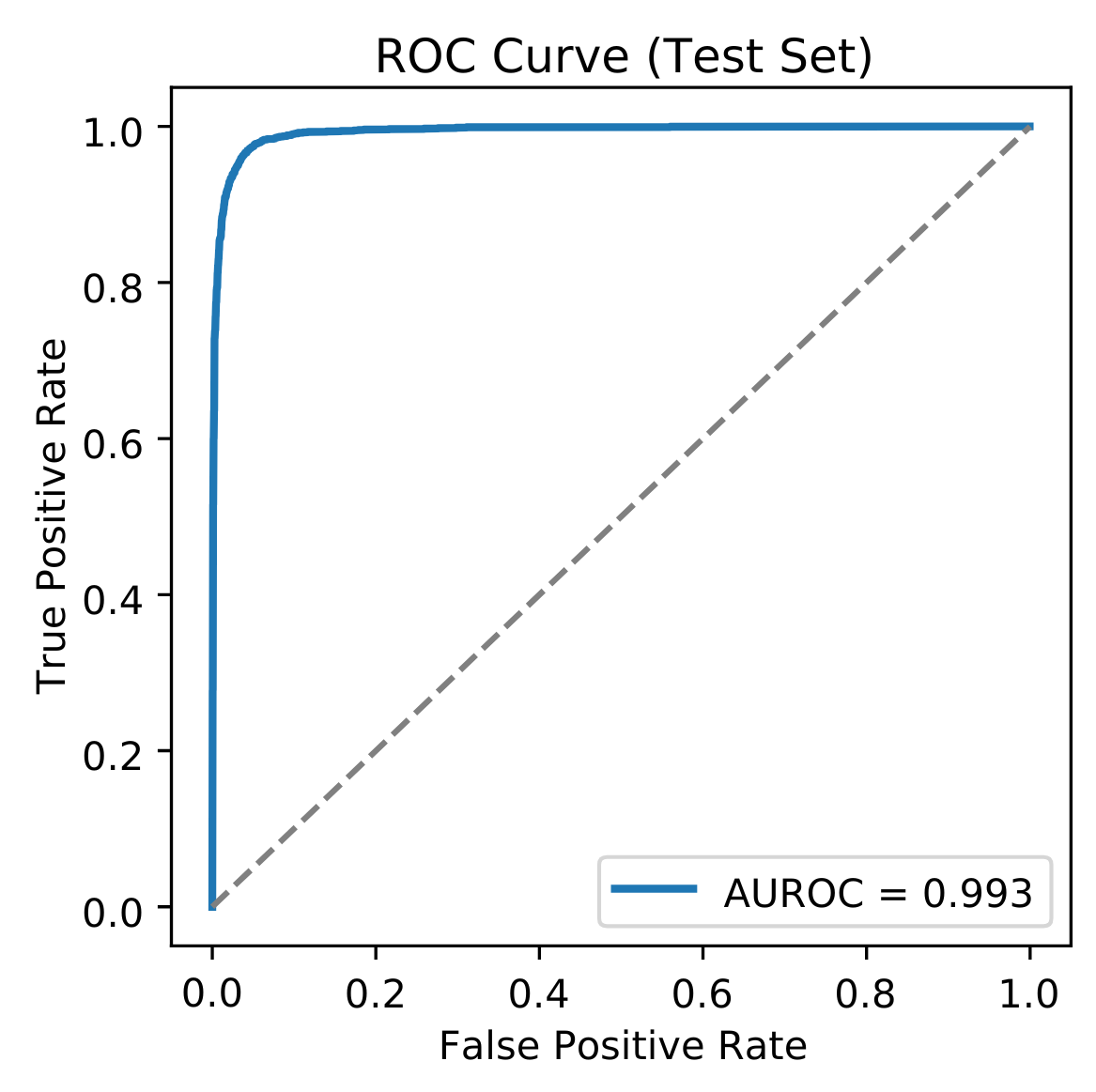}
    \caption{Test-set ROC curves for binary abnormality classification using linear head–only training, selective fine-tuning, and full fine-tuning. Selective fine-tuning achieves strong discriminative performance (AUROC = 0.96), closely matching full fine-tuning (AUROC = 0.99) while substantially outperforming head-only training (AUROC = 0.62).}
    \label{fig:roc_curves}
\end{figure*}

\begin{figure*}[!htbp]
    \centering
    \includegraphics[height=6cm, width=10cm]{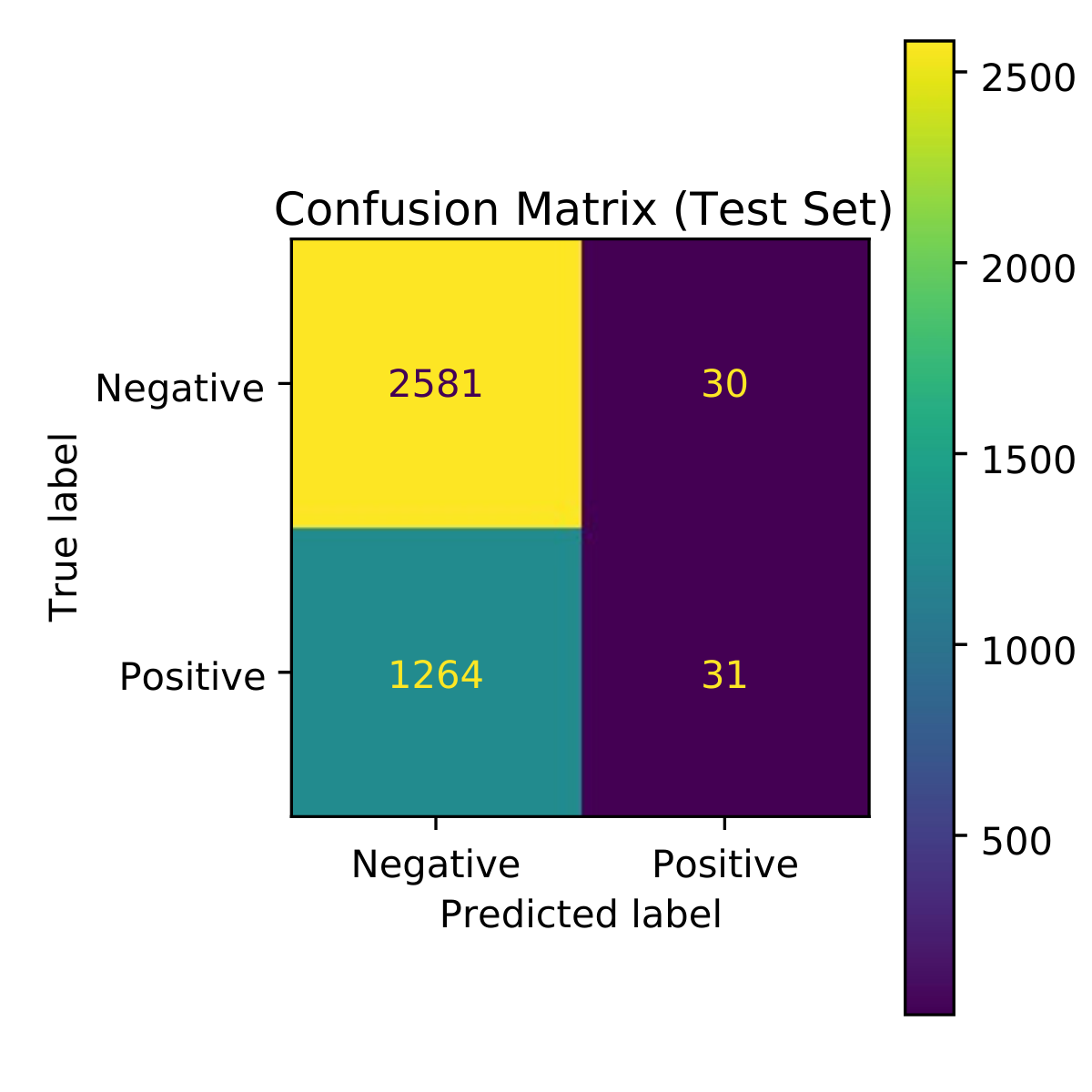}
    \includegraphics[height=6cm, width=10cm]{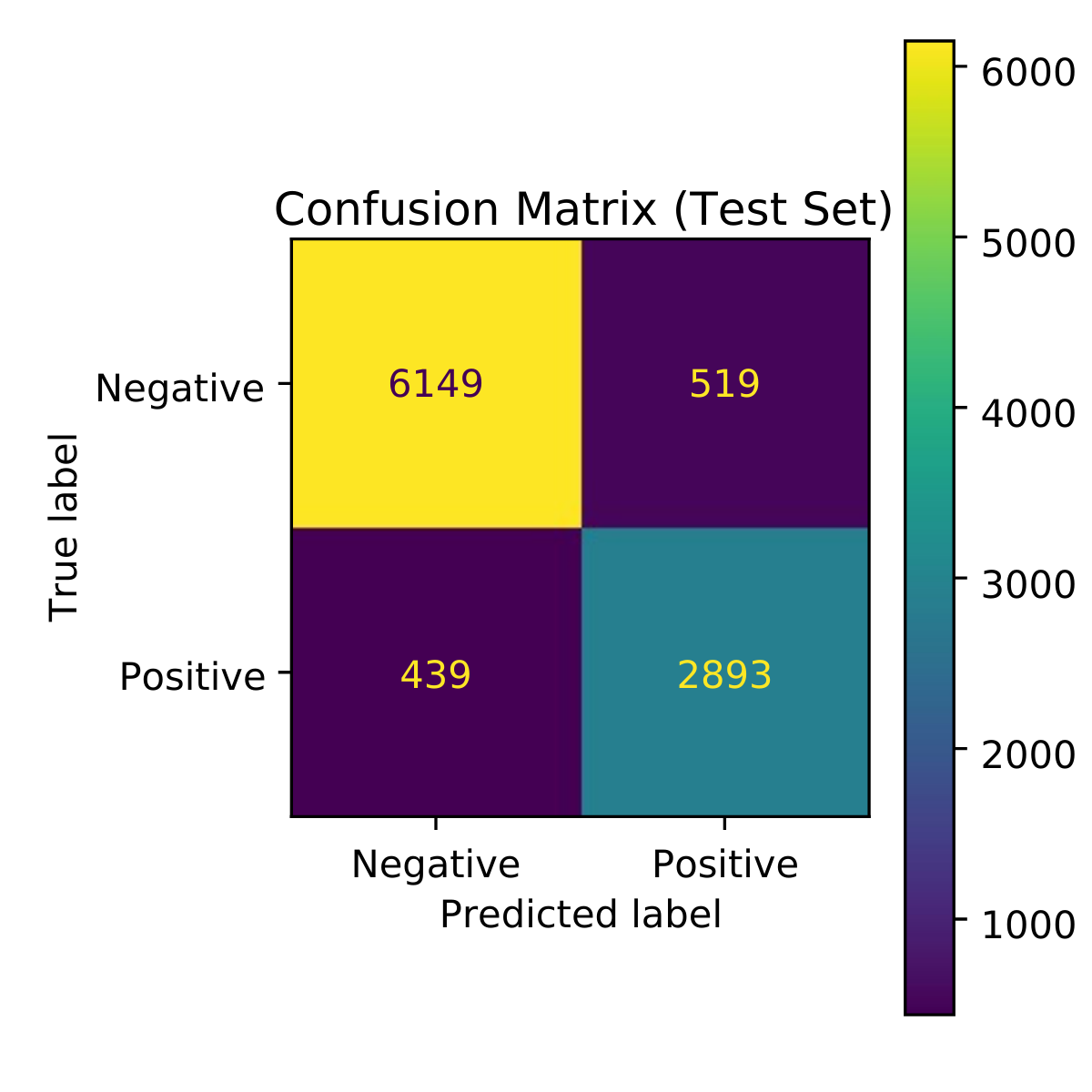}
    \includegraphics[height=6cm, width=10cm]{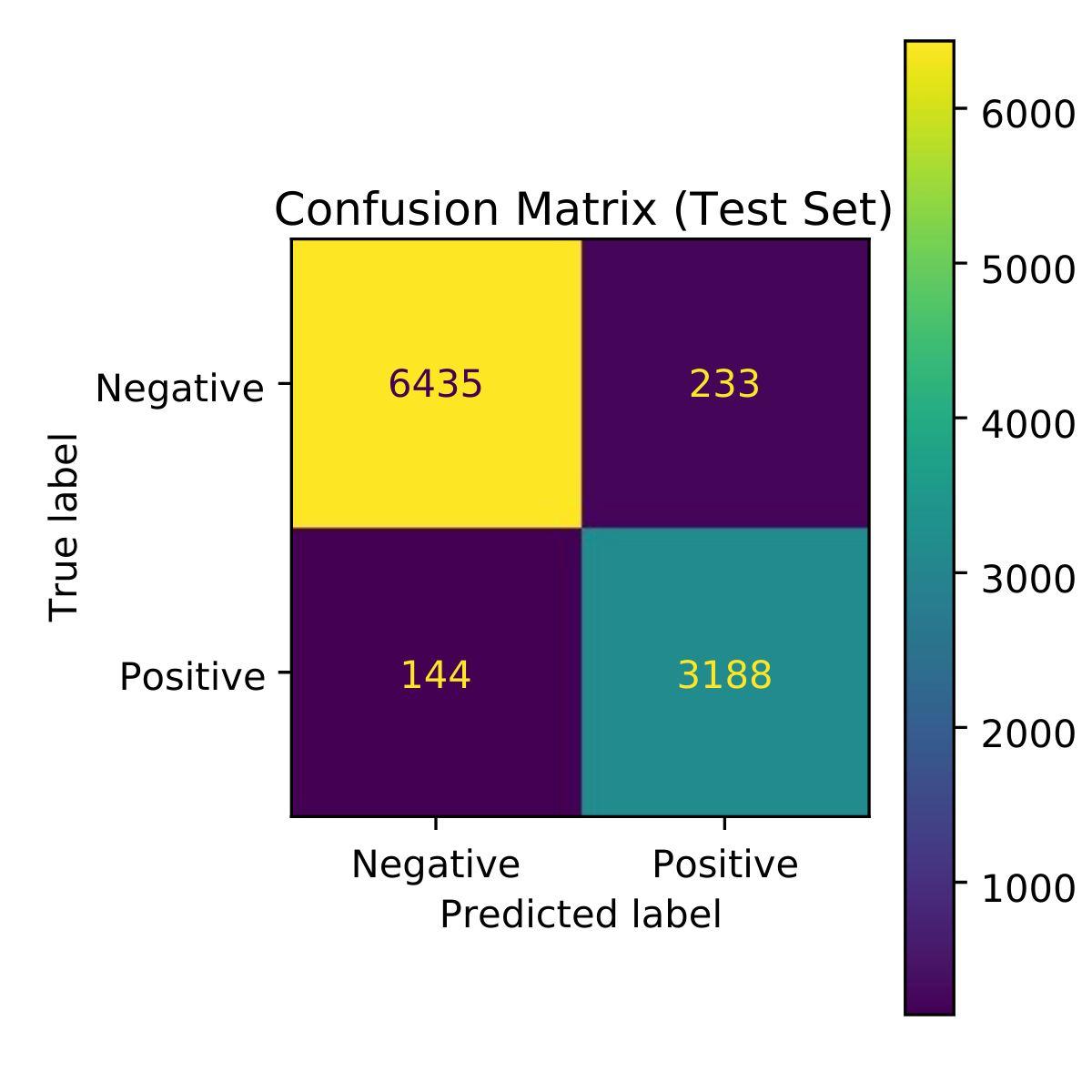}
    \caption{Test-set confusion matrices for binary abnormality classification under different fine-tuning strategies. Linear head–only training exhibits a high false-negative rate, whereas selective fine-tuning and full fine-tuning substantially reduce false negatives while maintaining low false-positive rates.}
    \label{fig:confusion_matrices}
\end{figure*}

To further assess classifier behavior beyond aggregate performance metrics, test-set discrimination is evaluated using receiver operating characteristic (ROC) curves, and error patterns are analyzed using confusion matrices. ROC curves provide a threshold-independent measure of separability between positive and negative classes, while confusion matrices offer insight into false positive and false negative distributions at the operating threshold.

Figure~\ref{fig:roc_curves} presents test-set ROC curves for linear head–only training, selective fine-tuning, and full fine-tuning. The selective fine-tuning strategy achieves strong discriminative performance with an AUROC of 0.96, closely matching full fine-tuning (AUROC = 0.99) and substantially outperforming linear head–only training (AUROC = 0.62). These results indicate that selective fine-tuning preserves most of the discriminative capacity of full fine-tuning while significantly reducing the number of trainable parameters.

To complement the ROC analysis, Figure~\ref{fig:confusion_matrices} shows the corresponding test-set confusion matrices. Linear head–only training exhibits a high false-negative rate, indicating limited capacity to detect abnormal findings. In contrast, both selective and full fine-tuning substantially reduce false negatives while maintaining low false-positive rates. Notably, selective fine-tuning achieves a favorable balance between sensitivity and specificity, yielding an error profile comparable to full fine-tuning.

\subsection{Training Dynamics and Convergence Behavior}

\begin{figure*}[!htbp]
\centering
\begin{subfigure}{0.48\textwidth}
    \centering
    \includegraphics[width=\linewidth]{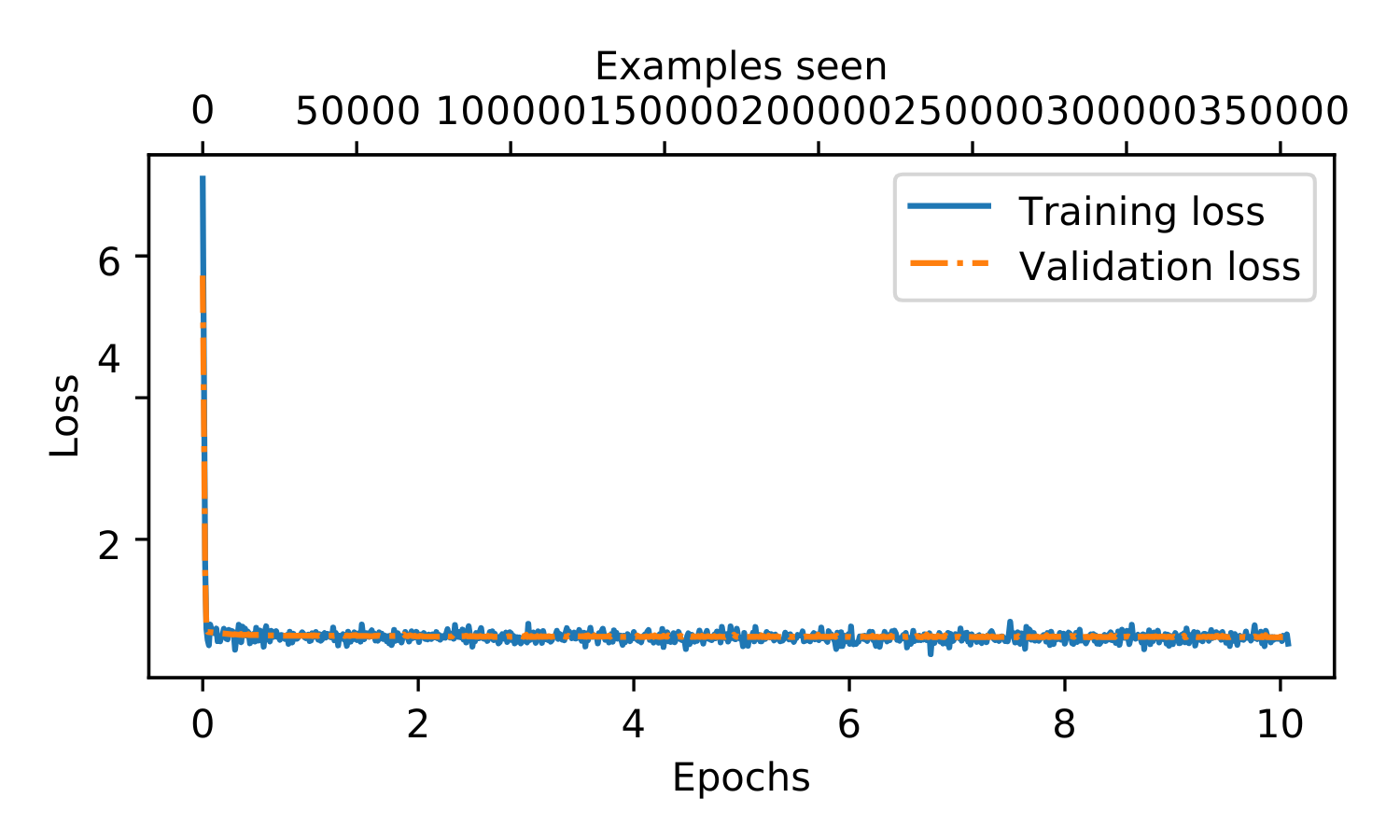}
    \caption{Training and validation loss}
    \label{fig:loss_head}
\end{subfigure}
\hfill
\begin{subfigure}{0.48\textwidth}
    \centering
    \includegraphics[width=\linewidth]{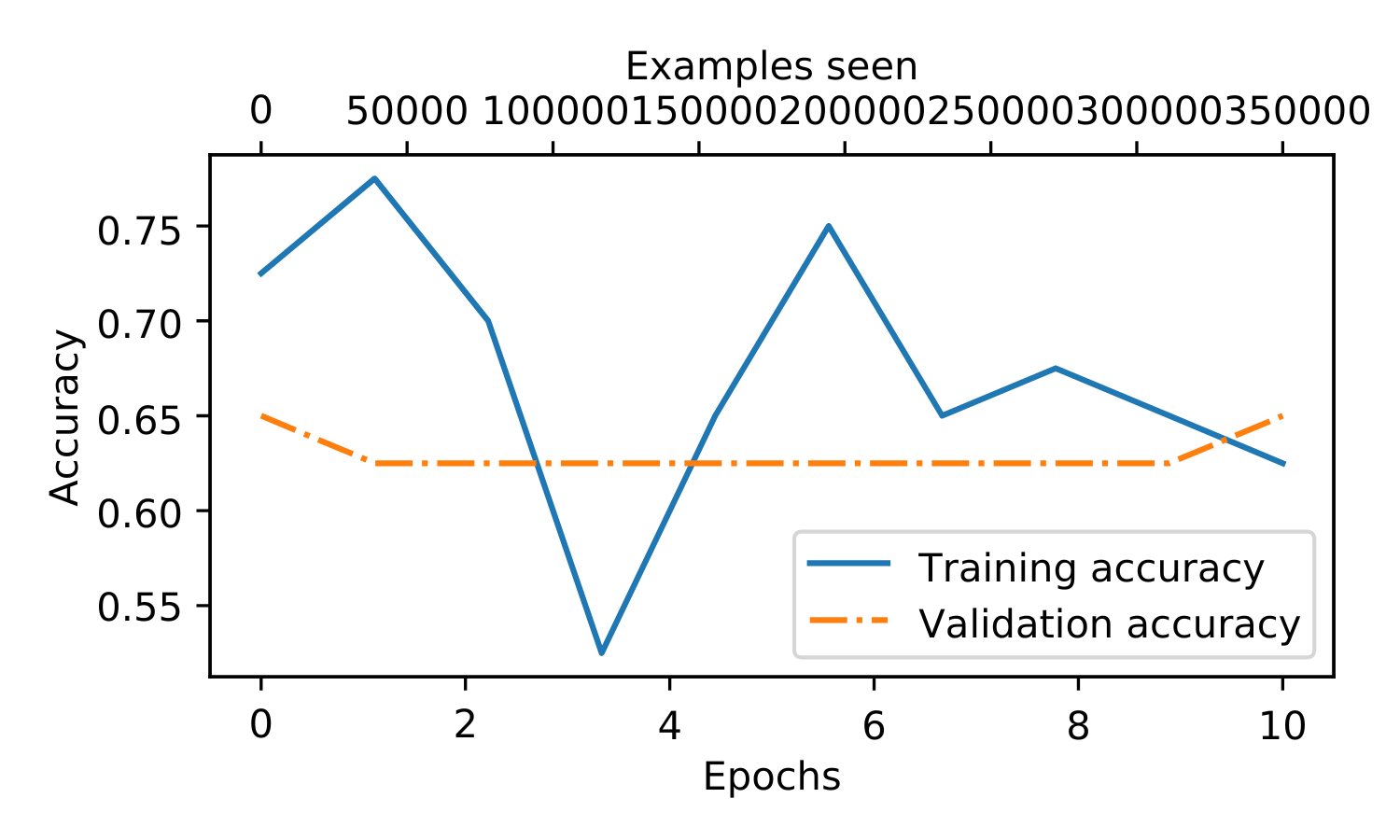}
    \caption{Training and validation accuracy}
    \label{fig:acc_head}
\end{subfigure}
\caption{Learning curves for classification-head-only fine-tuning on 500,00 radiology reports.}
\label{fig:learning_curves_classification_head}
\end{figure*}

\begin{figure*}[!htbp]
\centering
\begin{subfigure}{0.48\textwidth}
    \centering
    \includegraphics[width=\linewidth]{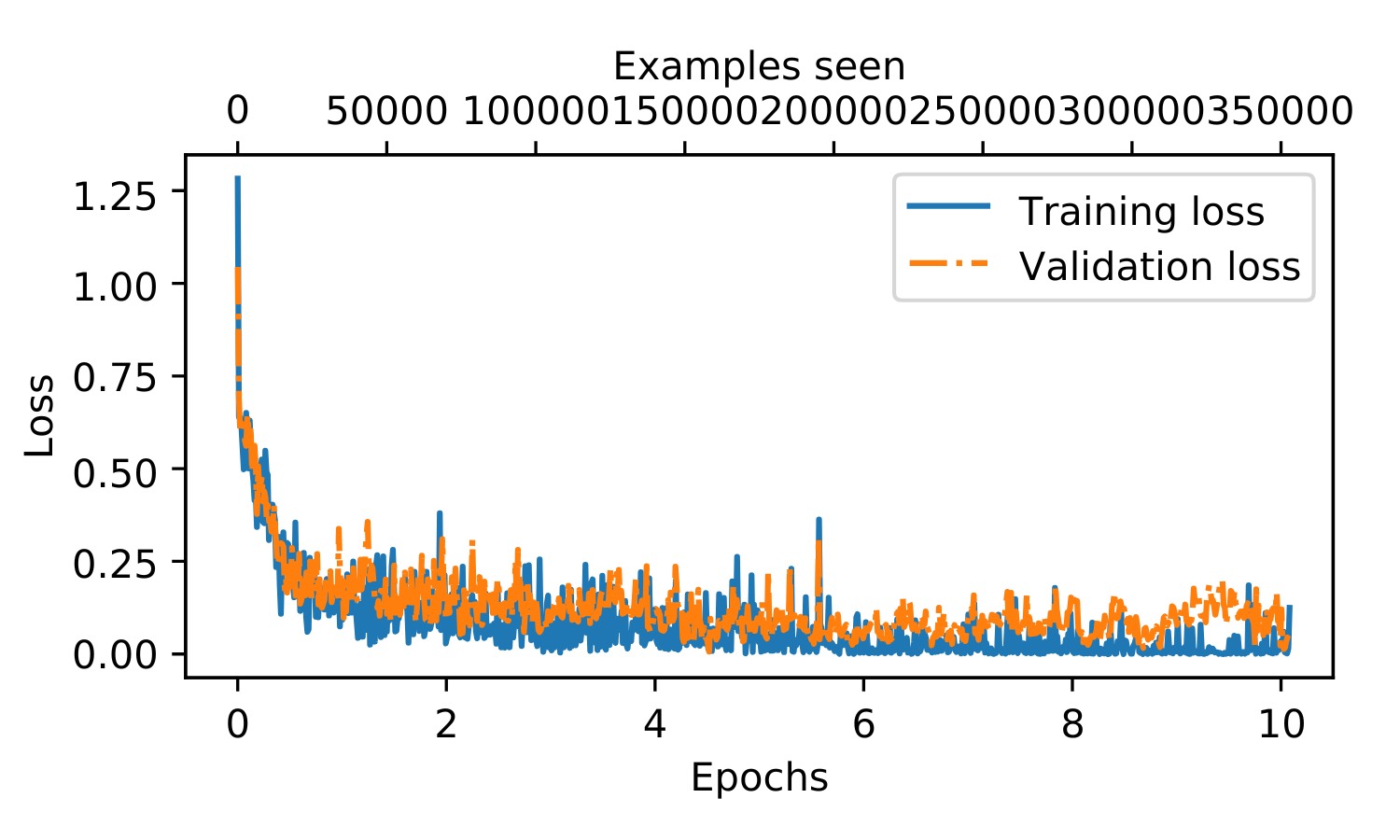}
    \caption{Training and validation loss}
    \label{fig:loss_full}
\end{subfigure}
\hfill
\begin{subfigure}{0.48\textwidth}
    \centering
    \includegraphics[width=\linewidth]{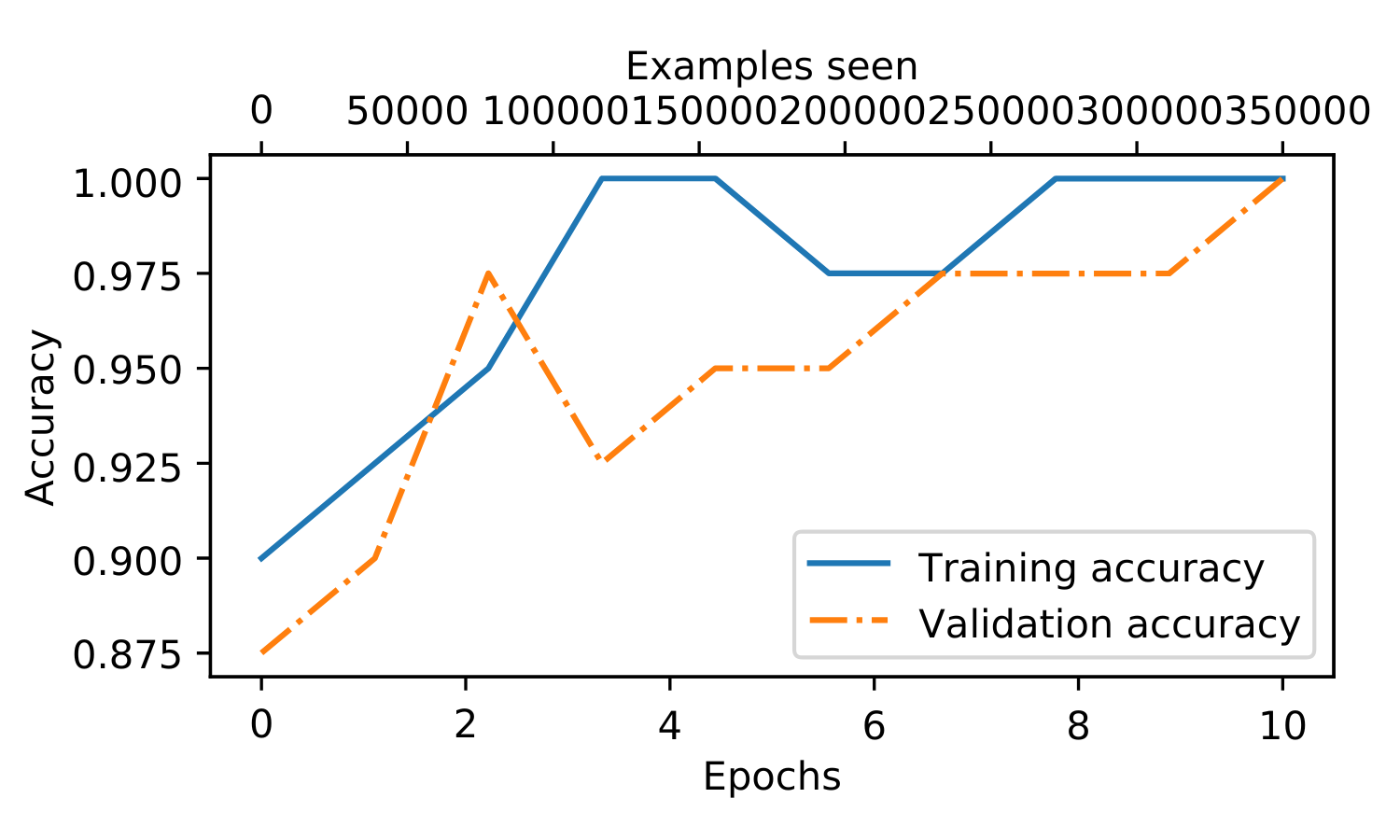}
    \caption{Training and validation accuracy}
    \label{fig:acc_full}
\end{subfigure}
\caption{Learning curves for full model fine-tuning on 500,00 radiology reports.}
\label{fig:learning_curves_full_finetune}
\end{figure*}

To further examine the optimization behavior and training stability of the proposed fine-tuning strategies, the evolution of training and validation loss, along with classification accuracy, is analyzed across successive epochs for the binary \texttt{label\_any\_disease\_pos\_or\_unc} classification task using 50,000 radiology reports.
Figures~\ref{fig:loss_head} and~\ref{fig:loss_full} present the loss trajectories for classification head–only optimization and full model fine-tuning, respectively, while Figures~\ref{fig:acc_head} and~\ref{fig:acc_full} depict the corresponding accuracy trends.

For classification head–only training, the loss decreases sharply during the initial epochs and reaches an early plateau, as shown in Figure~\ref{fig:loss_head}. Following convergence, training and validation loss curves closely align, indicating stable optimization but limited capacity for further improvement due to the frozen transformer backbone. This behavior is reflected in the accuracy trends in Figure~\ref{fig:acc_head}, where training accuracy exhibits moderate variability and validation accuracy remains largely constant across epochs, suggesting constrained representational adaptation.

In contrast, full model fine-tuning exhibits a more gradual yet sustained reduction in loss throughout training, as illustrated in Figure~\ref{fig:loss_full}. Both training and validation loss steadily decrease with increasing exposure to training examples, indicating effective parameter adaptation across the entire network. Correspondingly, the accuracy curves in Figure~\ref{fig:acc_full} demonstrate consistent improvement across epochs, with training and validation accuracy converging toward high values, reflecting enhanced generalization capability.

Overall, these training dynamics underscore the trade-off between optimization efficiency and representational flexibility. While classification head–only training converges rapidly with minimal computational cost, its limited adaptability constrains achievable performance. Full fine-tuning, although computationally more demanding, enables deeper feature refinement and yields more robust convergence behavior, consistent with the superior accuracy observed in the final evaluation results.

\section{Comparison with LoRA and QLoRA for Clinical Text Classification}

Recent work on parameter-efficient fine-tuning (PEFT) has introduced methods such as LoRA and QLoRA as alternatives to full-model fine-tuning for adapting large language models to downstream tasks. While these approaches have demonstrated strong performance for large-scale generative and instruction-tuning settings, their design objectives and optimization characteristics differ from the selective fine-tuning strategy proposed in this work, particularly for sequence-level classification of clinical text.

\subsection{LoRA}
LoRA (Low-Rank Adaptation) introduces trainable low-rank update matrices into selected weight matrices of a pretrained model, most commonly within the query and value projections of the self-attention mechanism \cite{hu2022lora}. During fine-tuning, the original pretrained weights remain frozen, and only the low-rank parameters are optimized. This design significantly reduces the number of trainable parameters while preserving the expressive capacity required for a wide range of downstream tasks.

Although effective, LoRA distributes task adaptation across multiple layers of the Transformer. For discriminative classification tasks, particularly in the clinical domain where labeled data are limited, this distributed adaptation may be unnecessary. Prior studies have shown that higher Transformer layers encode more task-specific representations, whereas lower layers capture general linguistic features \cite{peters2019tune,yosinski2014transferable}. By injecting trainable parameters throughout the network, LoRA may adapt representations beyond what is required for sequence-level classification, increasing optimization complexity and the risk of overfitting.

\subsection{QLoRA}
QLoRA extends LoRA by combining low-rank adaptation with aggressive weight quantization, typically using 4-bit quantized pretrained weights while maintaining performance comparable to full-precision fine-tuning \cite{dettmers2023qlora}. This approach enables fine-tuning of very large language models on memory-constrained hardware and has proven particularly effective for instruction tuning and conversational generation tasks.

However, the benefits of QLoRA are less pronounced for classification-focused applications. Quantization introduces approximation noise that can disproportionately affect discriminative decision boundaries, where small representation differences may significantly influence classification outcomes. Moreover, QLoRA is primarily motivated by scalability to extremely large models, whereas many clinical NLP applications prioritize stability, interpretability, and controlled adaptation over maximum model scale.

\subsection{Advantages of the Proposed Selective Fine-Tuning Strategy}
In contrast to LoRA and QLoRA, the approach proposed in this paper adopts a structurally minimal and classification-oriented fine-tuning strategy. Rather than introducing additional parameters or modifying internal attention projections, optimization is restricted to the final Transformer block, the final layer normalization, and a lightweight task-specific classification head, while all earlier layers are frozen.

This design offers several advantages for clinical text classification. First, it aligns directly with the discriminative objective of sequence-level classification. In decoder-only architectures, the hidden state of the final token implicitly aggregates contextual information from the entire input sequence under causal self-attention \cite{radford2019language}. Adapting only the highest-level representations therefore targets the layers most relevant to classification decisions.

Second, by limiting the number of trainable parameters, the proposed approach substantially reduces the risk of overfitting in data-limited clinical settings. At the same time, freezing lower layers preserves the general linguistic knowledge acquired during large-scale pretraining, mitigating catastrophic forgetting \cite{peters2019tune}.

Third, the proposed strategy enables transparent parameter accounting and computational complexity analysis. Unlike LoRA and QLoRA, which introduce distributed low-rank updates across multiple layers, the selective fine-tuning approach allows explicit quantification of trainable parameters and training-time cost. This transparency is particularly important in clinical research contexts, where reproducibility and methodological clarity are essential.

Finally, the proposed method avoids architectural modification, custom parameter injection, or quantization-aware training, resulting in a simpler optimization pipeline that is easier to implement and deploy in academic and clinical environments. While LoRA and QLoRA are designed to preserve generative flexibility across diverse tasks, the proposed approach intentionally prioritizes stable and efficient discriminative performance, which is the primary requirement for clinical phenotyping, outcome prediction, and diagnostic classification.

Overall, while LoRA and QLoRA represent powerful and general-purpose PEFT methods, their strengths are most evident in large-scale generative and instruction-tuned settings. For clinical text classification tasks characterized by long documents, limited labeled data, and strict interpretability requirements, the selective fine-tuning strategy proposed in this work provides a simpler, more transparent, and computationally efficient alternative.

\subsection*{Data Availability Statement}

This research used the MIMIC-IV clinical database, which is accessible to credentialed researchers through PhysioNet. In accordance with the data use agreement, patient-level clinical text and derived datasets are not shared by the authors. Reproducibility is supported through publicly available code and Google Colab notebooks at:
\url{https://drive.google.com/drive/folders/1GI21lZZiRAiI8Y-C5FEU1PsTklN0487X}

%% file: Conclusion.tex
\section{Conclusion}

This study investigated the adaptation of a pretrained GPT-based language model for clinical text classification using a selective fine-tuning strategy designed to balance performance, interpretability, and computational efficiency. By freezing the majority of the pretrained GPT-2 parameters and updating only the final Transformer block, the final layer normalization, and a task-specific classification head, training complexity is substantially reduced while preserving the expressive power of the underlying language model.

Through experiments on radiology reports from the MIMIC-IV-Note dataset, the proposed approach is shown to perform effectively across multiple clinical prediction settings, including multi-label classification of CheXpert-style findings, binary per-label classification, and aggregate disease outcome prediction. The observed training and validation loss trajectories indicate stable optimization behavior, and accuracy results improve consistently with increasing sample size. Notably, multi-label classification experiments exhibit high overall accuracy, which is attributable in part to the prevalence of null or negative findings in real-world radiology reports, highlighting the importance of careful interpretation of performance metrics in imbalanced clinical datasets.

Beyond empirical performance, a key contribution of this work is a transparent and detailed analysis of model parameterization and training-time complexity. By explicitly quantifying frozen and trainable parameters, this study provides practical guidance for researchers seeking to apply large language models in data-limited or resource-constrained clinical environments. The accompanying notebooks and documentation further support reproducibility and extensibility.

Future work will explore extending this framework to additional clinical note types, incorporating alternative uncertainty-handling strategies, and evaluating more parameter-efficient adaptation methods in comparison to selective fine-tuning. Overall, the results demonstrate that generative pretrained language models can be effectively and efficiently repurposed for clinical text classification without full-model fine-tuning, offering a scalable and accessible approach for clinical NLP research.

%% file: cogsci_bibliography_template.bib
@inproceedings{paszke2019pytorch,
  title     = {PyTorch: An Imperative Style, High-Performance Deep Learning Library},
  author    = {Paszke, Adam and Gross, Sam and Massa, Francisco and 
               Lerer, Adam and Bradbury, James and Chanan, Gregory and 
               others},
  booktitle = {Advances in Neural Information Processing Systems},
  volume    = {32},
  year      = {2019}
}

@article{gage1994bpe,
  title={A New Algorithm for Data Compression},
  author={Gage, Philip},
  journal={C Users Journal},
  volume={12},
  number={2},
  year={1994}
}

@inproceedings{sennrich2016bpe,
  title={Neural Machine Translation of Rare Words with Subword Units},
  author={Sennrich, Rico and Haddow, Barry and Birch, Alexandra},
  booktitle={Proceedings of the 54th Annual Meeting of the Association for Computational Linguistics},
  year={2016}
}

@inproceedings{vaswani2017attention,
  title     = {Attention Is All You Need},
  author    = {Vaswani, Ashish and Shazeer, Noam and Parmar, Niki and Uszkoreit, Jakob and Jones, Llion and Gomez, Aidan N. and Kaiser, Lukasz and Polosukhin, Illia},
  booktitle = {Advances in Neural Information Processing Systems},
  volume    = {30},
  year      = {2017},
  url       = {https://papers.neurips.cc/paper/7181-attention-is-all-you-need.pdf}
}

@article{radford2018improving,
  title     = {Improving Language Understanding by Generative Pre-Training},
  author    = {Radford, Alec and Narasimhan, Karthik and Salimans, Tim and Sutskever, Ilya},
  journal   = {OpenAI Technical Report},
  year      = {2018},
  url       = {https://cdn.openai.com/research-covers/language-unsupervised/language_understanding_paper.pdf}
}

@article{radford2019language,
  title     = {Language Models are Unsupervised Multitask Learners},
  author    = {Radford, Alec and Wu, Jeffrey and Child, Rewon and Luan, David and Amodei, Dario and Sutskever, Ilya},
  journal   = {OpenAI Technical Report},
  year      = {2019},
  url       = {https://openai.com/research/language-models}
}

@article{srivastava2014dropout,
  title     = {Dropout: A Simple Way to Prevent Neural Networks from Overfitting},
  author    = {Srivastava, Nitish and Hinton, Geoffrey and Krizhevsky, Alex and Sutskever, Ilya and Salakhutdinov, Ruslan},
  journal   = {Journal of Machine Learning Research},
  volume    = {15},
  number    = {1},
  pages     = {1929--1958},
  year      = {2014},
  url       = {https://jmlr.org/papers/v15/srivastava14a.html}
}

@article{ba2016layernorm,
  title     = {Layer Normalization},
  author    = {Ba, Jimmy Lei and Kiros, Jamie Ryan and Hinton, Geoffrey},
  journal   = {arXiv preprint arXiv:1607.06450},
  year      = {2016},
  url       = {https://arxiv.org/abs/1607.06450}
}

@article{he2016deep,
  title={Deep Residual Learning for Image Recognition},
  author={He, Kaiming and others},
  journal={Proceedings of the IEEE Conference on Computer Vision and Pattern Recognition},
  year={2016}
}

@article{xu2020understanding,
  title={Understanding and Improving Transformer From a Multi-Particle Dynamic System Point of View},
  author={Xu, Jingjing and others},
  journal={arXiv preprint arXiv:2002.04745},
  year={2020}
}

@article{hendrycks2016gelu,
  title={Gaussian Error Linear Units (GELUs)},
  author={Hendrycks, Dan and Gimpel, Kevin},
  journal={arXiv preprint arXiv:1606.08415},
  year={2016}
}

@book{goodfellow2016deep,
  title={Deep Learning},
  author={Goodfellow, Ian and Bengio, Yoshua and Courville, Aaron},
  publisher={MIT Press},
  year={2016},
  url={https://www.deeplearningbook.org/}
}

@book{bishop2006pattern,
  title={Pattern Recognition and Machine Learning},
  author={Bishop, Christopher M.},
  publisher={Springer},
  year={2006},
  url={https://link.springer.com/book/10.1007/978-0-387-45528-0}
}

@article{loshchilov2019adamw,
  title={Decoupled Weight Decay Regularization},
  author={Loshchilov, Ilya and Hutter, Frank},
  journal={ICLR},
  year={2019},
  url={https://arxiv.org/abs/1711.05101}
}

@article{peters2019tune,
  title={To Tune or Not to Tune? Adapting Pretrained Representations to Diverse Tasks},
  author={Peters, Matthew E. and Ruder, Sebastian and Smith, Noah A.},
  journal={ACL Workshop},
  year={2019},
  url={https://arxiv.org/abs/1903.05987}
}

@article{yosinski2014transfer,
  title={How Transferable Are Features in Deep Neural Networks?},
  author={Yosinski, Jason and Clune, Jeff and Bengio, Yoshua and Lipson, Hod},
  journal={NeurIPS},
  year={2014},
  url={https://arxiv.org/abs/1411.1792}
}

@online{openai2019gpt2,
  title        = {GPT-2: 1.5B Release},
  author       = {{OpenAI}},
  year         = {2019},
  url          = {https://openai.com/index/gpt-2-1-5b-release/}
}

@inproceedings{brown2020language,
  title     = {Language Models are Few-Shot Learners},
  author    = {Brown, Tom B. and Mann, Benjamin and Ryder, Nick and Subbiah, Melanie and Kaplan, Jared and Dhariwal, Prafulla and others},
  booktitle = {Advances in Neural Information Processing Systems},
  year      = {2020},
  url       = {https://proceedings.neurips.cc/paper/2020/file/1457c0d6bfcb4967418bfb8ac142f64a-Paper.pdf}
}

@inproceedings{devlin2019bert,
  title     = {BERT: Pre-training of Deep Bidirectional Transformers for Language Understanding},
  author    = {Devlin, Jacob and Chang, Ming-Wei and Lee, Kenton and Toutanova, Kristina},
  booktitle = {Proceedings of the 2019 Conference of the North American Chapter of the Association for Computational Linguistics: Human Language Technologies (NAACL-HLT)},
  year      = {2019},
  url       = {https://aclanthology.org/N19-1423/}
}

@article{peters2018elmo,
  title={Deep contextualized word representations},
  author={Peters, Matthew E and Neumann, Mark and Iyyer, Mohit and Gardner, Matt and Clark, Christopher and Lee, Kenton and Zettlemoyer, Luke},
  journal={NAACL},
  year={2018},
  url={https://arxiv.org/abs/1802.05365}
}

@article{howard2018universal,
  title={Universal Language Model Fine-tuning for Text Classification},
  author={Howard, Jeremy and Ruder, Sebastian},
  journal={ACL},
  year={2018},
  url={https://arxiv.org/abs/1801.06146}
}

@article{liu2019roberta,
  title={RoBERTa: A Robustly Optimized BERT Pretraining Approach},
  author={Liu, Yinhan and Ott, Myle and Goyal, Naman and Du, Jingfei and Joshi, Mandar and Chen, Danqi and Levy, Omer and Lewis, Mike and Zettlemoyer, Luke and Stoyanov, Veselin},
  journal={arXiv preprint arXiv:1907.11692},
  year={2019},
  url={https://arxiv.org/abs/1907.11692}
}

@article{lan2019albert,
  title={ALBERT: A Lite BERT for Self-supervised Learning of Language Representations},
  author={Lan, Zhenzhong and Chen, Mingda and Goodman, Sebastian and Gimpel, Kevin and Sharma, Piyush and Soricut, Radu},
  journal={ICLR},
  year={2020},
  url={https://arxiv.org/abs/1909.11942}
}

@article{yang2019xlnet,
  title={XLNet: Generalized Autoregressive Pretraining for Language Understanding},
  author={Yang, Zhilin and Dai, Zihang and Yang, Yiming and Carbonell, Jaime and Salakhutdinov, Ruslan and Le, Quoc V},
  journal={NeurIPS},
  year={2019},
  url={https://arxiv.org/abs/1906.08237}
}

@article{raffel2020t5,
  title={Exploring the Limits of Transfer Learning with a Unified Text-to-Text Transformer},
  author={Raffel, Colin and Shazeer, Noam and Roberts, Adam and Lee, Katherine and Narang, Sharan and Matena, Michael and Zhou, Yanqi and Li, Wei and Liu, Peter J},
  journal={JMLR},
  year={2020},
  url={https://arxiv.org/abs/1910.10683}
}

@article{clark2020electra,
  title={ELECTRA: Pre-training Text Encoders as Discriminators Rather Than Generators},
  author={Clark, Kevin and Luong, Minh-Thang and Le, Quoc V and Manning, Christopher D},
  journal={ICLR},
  year={2020},
  url={https://arxiv.org/abs/2003.10555}
}

@article{dai2019transformerxl,
  title={Transformer-XL: Attentive Language Models Beyond a Fixed-Length Context},
  author={Dai, Zihang and Yang, Zhilin and Yang, Yiming and Carbonell, Jaime and Le, Quoc and Salakhutdinov, Ruslan},
  journal={ACL},
  year={2019},
  url={https://arxiv.org/abs/1901.02860}
}

@article{beltagy2020longformer,
  title={Longformer: The Long-Document Transformer},
  author={Beltagy, Iz and Peters, Matthew E and Cohan, Arman},
  journal={arXiv preprint arXiv:2004.05150},
  year={2020},
  url={https://arxiv.org/abs/2004.05150}
}

@inproceedings{zaheer2020bigbird,
  title     = {Big Bird: Transformers for Longer Sequences},
  author    = {Zaheer, Manzil and Guruganesh, Guru and Dubey, Avinava and Ainslie, Joshua and Alberti, Chris and Ontanon, Santiago and Pham, Philip and Ravula, Anirudh and Wang, Qifan and Yang, Li and Ahmed, Amr},
  booktitle = {Advances in Neural Information Processing Systems},
  year      = {2020},
  url       = {https://papers.neurips.cc/paper_files/paper/2020/file/c8512d142a2d849725f31a9a7a361ab9-Paper.pdf}
}

@article{kitaev2020reformer,
  title={Reformer: The Efficient Transformer},
  author={Kitaev, Nikita and Kaiser, Lukasz and Levskaya, Anselm},
  journal={ICLR},
  year={2020},
  url={https://arxiv.org/abs/2001.04451}
}

@article{wang2020linformer,
  title={Linformer: Self-Attention with Linear Complexity},
  author={Wang, Sinong and Li, Belinda Z and Khabsa, Madian and Fang, Han and Ma, Hao},
  journal={arXiv preprint arXiv:2006.04768},
  year={2020},
  url={https://arxiv.org/abs/2006.04768}
}

@article{choromanski2021performer,
  title={Rethinking Attention with Performers},
  author={Choromanski, Krzysztof and Likhosherstov, Valerii and Dohan, David and Song, Xingyou and Gane, Andreea and Sarlos, Tamas and Hawkins, Peter and Davis, Jared and Mohiuddin, Afroz and Kaiser, Lukasz and Belanger, David and Colwell, Lucy and Weller, Adrian},
  journal={ICLR},
  year={2021},
  url={https://arxiv.org/abs/2009.14794}
}

@inproceedings{gururangan2020don,
  title     = {Don't Stop Pretraining: Adapt Language Models to Domains and Tasks},
  author    = {Gururangan, Suchin and Marasovic, Ana and Swayamdipta, Swabha and Lo, Kyle and Beltagy, Iz and Downey, Doug and Smith, Noah A.},
  booktitle = {Proceedings of the 58th Annual Meeting of the Association for Computational Linguistics (ACL)},
  year      = {2020},
  url       = {https://aclanthology.org/2020.acl-main.740/}
}

@article{beltagy2019scibert,
  title={SciBERT: A Pretrained Language Model for Scientific Text},
  author={Beltagy, Iz and Lo, Kyle and Cohan, Arman},
  journal={EMNLP},
  year={2019},
  url={https://arxiv.org/abs/1903.10676}
}

@article{lee2020biobert,
  title={BioBERT: a pre-trained biomedical language representation model for biomedical text mining},
  author={Lee, Jinhyuk and Yoon, Wonjin and Kim, Sungdong and Kim, Donghyeon and Kim, Sunkyu and So, Chan and Kang, Jaewoo},
  journal={Bioinformatics},
  year={2020},
  url={https://academic.oup.com/bioinformatics/article/36/4/1234/5566506}
}

@article{gu2021pubmedbert,
  title={Domain-Specific Language Model Pretraining for Biomedical Natural Language Processing},
  author={Gu, Yu and Tinn, Robert and Cheng, Hao and Lucas, Michael and Usuyama, Naoto and Liu, Xiaodong and Naumann, Tristan and Gao, Jianfeng and Poon, Hoifung},
  journal={ACL},
  year={2021},
  url={https://arxiv.org/abs/2007.15779}
}

@article{shin2020biomegatron,
  title={BioMegatron: Larger Biomedical Domain Language Model},
  author={Shin, Hoo-Chang and others},
  journal={arXiv preprint arXiv:2010.06060},
  year={2020},
  url={https://arxiv.org/abs/2010.06060}
}

@article{huang2019clinicalbert,
  title={ClinicalBERT: Modeling Clinical Notes and Predicting Hospital Readmission},
  author={Huang, Kexin and Altosaar, Jaan and Ranganath, Rajesh},
  journal={arXiv preprint arXiv:1904.05342},
  year={2019},
  url={https://arxiv.org/abs/1904.05342}
}

@inproceedings{houlsby2019adapters,
  title     = {Parameter-Efficient Transfer Learning for NLP},
  author    = {Houlsby, Neil and Giurgiu, Andrei and Jastrzebski, Stanislaw and Morrone, Bruna and de Laroussilhe, Quentin and Gesmundo, Andrea and Attariyan, Mona and Gelly, Sylvain},
  booktitle = {Proceedings of the 36th International Conference on Machine Learning (ICML)},
  series    = {Proceedings of Machine Learning Research},
  volume    = {97},
  pages     = {2790--2799},
  year      = {2019},
  url       = {https://proceedings.mlr.press/v97/houlsby19a.html}
}

@article{pfeiffer2020adapterfusion,
  title={AdapterFusion: Non-Destructive Task Composition for Transfer Learning},
  author={Pfeiffer, Jonas and others},
  journal={EACL},
  year={2021},
  url={https://arxiv.org/abs/2005.00247}
}

@article{hu2022lora,
  title={LoRA: Low-Rank Adaptation of Large Language Models},
  author={Hu, Edward J and Shen, Yelong and Wallis, Phillip and Allen-Zhu, Zeyuan and Li, Yuanzhi and Wang, Shean and Chen, Weizhu},
  journal={ICLR},
  year={2022},
  url={https://arxiv.org/abs/2106.09685}
}

@inproceedings{li2021prefixtuning,
  title     = {Prefix-Tuning: Optimizing Continuous Prompts for Generation},
  author    = {Li, Xiang Lisa and Liang, Percy},
  booktitle = {Proceedings of the 59th Annual Meeting of the Association for Computational Linguistics (ACL)},
  year      = {2021},
  url       = {https://aclanthology.org/2021.acl-long.353/}
}

@article{lester2021prompttuning,
  title={The Power of Scale for Parameter-Efficient Prompt Tuning},
  author={Lester, Brian and Al-Rfou, Rami and Constant, Noah},
  journal={EMNLP},
  year={2021},
  url={https://arxiv.org/abs/2104.08691}
}

@article{liu2021ptuning,
  title={P-Tuning: Prompt Tuning Can Be Comparable to Fine-tuning Across Scales and Tasks},
  author={Liu, Xiao and Zheng, Yanan and Du, Zhengxiao and Ding, Ming and Qian, Yujie and Yang, Zhilin and Tang, Jie},
  journal={arXiv preprint arXiv:2103.10385},
  year={2021},
  url={https://arxiv.org/abs/2103.10385}
}

@article{zaken2022bitfit,
  title={BitFit: Simple Parameter-efficient Fine-tuning for Transformer-based Masked Language-models},
  author={Zaken, Elad Ben and Ravfogel, Shauli and Goldberg, Yoav},
  journal={ACL Findings},
  year={2022},
  url={https://arxiv.org/abs/2106.10199}
}

@article{liu2022ia3,
  title={Few-shot Parameter-efficient Fine-tuning is Better and Cheaper than In-context Learning},
  author={Liu, Haokun and Tam, Derek and Muqeeth, Mohammed and Mohta, Jay and Huang, Tenghao and Bansal, Mohit and Raffel, Colin},
  journal={NeurIPS},
  year={2022},
  url={https://arxiv.org/abs/2205.05638}
}

@article{ding2023peftsurvey,
  title={Parameter-Efficient Fine-Tuning of Large-Scale Pre-trained Language Models: A Survey},
  author={Ding, Ning and others},
  journal={arXiv preprint arXiv:2303.15647},
  year={2023},
  url={https://arxiv.org/abs/2303.15647}
}

@article{goldberger2000physionet,
  title   = {PhysioBank, PhysioToolkit, and PhysioNet: Components of a new research resource for complex physiologic signals},
  author  = {Goldberger, Ary L. and Amaral, Luis A. N. and Glass, Leon and Hausdorff, Jeffrey M. and Ivanov, Plamen Ch. and Mark, Roger G. and Mietus, Joseph E. and Moody, George B. and Peng, Chung-Kang and Stanley, H. Eugene},
  journal = {Circulation},
  volume  = {101},
  number  = {23},
  pages   = {e215--e220},
  year    = {2000},
  url     = {https://physionet.org/about/publications/},
  note    = {RRID:SCR\_007345}
}

@article{press2017using,
  title={Using the Output Embedding to Improve Language Models},
  author={Press, Ofir and Wolf, Lior},
  journal={EACL},
  year={2017},
  url={https://arxiv.org/abs/1608.05859}
}

@article{johnson2016mimic,
  title={MIMIC-III, a freely accessible critical care database},
  author={Johnson, Alistair E W and Pollard, Tom J and Shen, Lu and Li-Wei, H Lehman and Feng, Mengling and Ghassemi, Mohammad and Moody, Benjamin and Szolovits, Peter and Celi, Leo Anthony and Mark, Roger G},
  journal={Scientific Data},
  year={2016},
  url={https://www.nature.com/articles/sdata201635}
}

@article{uzuner20112010i2b2,
  title={Evaluating the state of the art in coreference resolution for electronic medical records},
  author={Uzuner, Ozlem and others},
  journal={Journal of the American Medical Informatics Association},
  year={2011},
  url={https://academic.oup.com/jamia/article/19/5/786/683705}
}

@inproceedings{mullenbach2018icd,
  title     = {Explainable Prediction of Medical Codes from Clinical Text},
  author    = {Mullenbach, James and Wiegreffe, Sarah and Duke, Jonathon and Sun, Jimeng and Eisenstein, Jacob},
  booktitle = {Proceedings of NAACL-HLT 2018},
  year      = {2018},
  doi       = {10.18653/v1/N18-1100},
  url       = {https://aclanthology.org/N18-1100/}
}

@article{li2019caml,
  title={CAML: Convolutional Attention-based Multi-label learning for ICD coding},
  author={Li, Fei and Yu, Hong},
  journal={AAAI},
  year={2019},
  url={https://ojs.aaai.org/index.php/AAAI/article/view/3806}
}

@article{singhal2023medpalm,
  title={Large Language Models Encode Clinical Knowledge},
  author={Singhal, Karan and others},
  journal={Nature},
  year={2023},
  url={https://www.nature.com/articles/s41586-023-06291-2}
}

@article{jiang2023healthsystemreview,
  title={A Survey of Large Language Models for Healthcare},
  author={Jiang, Zhe and others},
  journal={arXiv preprint arXiv:2303.01246},
  year={2023},
  url={https://arxiv.org/abs/2303.01246}
}

@article{johnson2023mimiciv,
  title={MIMIC-IV, a freely accessible electronic health record dataset},
  author={Johnson, Alistair E. W. and Bulgarelli, Lucas and Pollard, Tom J. and Horng, Steven and Celi, Leo Anthony and Mark, Roger G.},
  journal={Scientific Data},
  volume={10},
  number={1},
  pages={1--10},
  year={2023},
  url={https://www.nature.com/articles/s41597-022-01899-x}
}

@article{xiong2020layernorm,
  title={On layer normalization in the Transformer architecture},
  journal={ICLR},
  year={2020},
  url={https://arxiv.org/abs/2002.04745}
}

@article{dettmers2023qlora,
  title={QLoRA: Efficient Finetuning of Quantized LLMs},
  author={Dettmers, Tim and Pagnoni, Artidoro and Holtzman, Ari and Zettlemoyer, Luke},
  journal={Advances in Neural Information Processing Systems},
  year={2023},
  url={https://arxiv.org/abs/2305.14314}
}

@article{yosinski2014transferable,
  title={How transferable are features in deep neural networks?},
  author={Yosinski, Jason and Clune, Jeff and Bengio, Yoshua and Lipson, Hod},
  journal={Advances in Neural Information Processing Systems},
  year={2014},
  url={https://arxiv.org/abs/1411.1792}
}

@inproceedings{wolf2020transformers,
  title     = {Transformers: State-of-the-Art Natural Language Processing},
  author    = {Wolf, Thomas and Debut, Lysandre and Sanh, Victor and Chaumond, Julien and Delangue, Cl{\'e}ment and Moi, Anthony and Cistac, Pierric and Rault, Tim and Louf, R{\'e}mi and Funtowicz, Morgan and Brew, Jamie},
  booktitle = {Proceedings of the 2020 Conference on Empirical Methods in Natural Language Processing: System Demonstrations},
  year      = {2020},
  pages     = {38--45},
  url       = {https://aclanthology.org/2020.emnlp-demos.6/}
}

@article{ba2016layer,
  title   = {Layer Normalization},
  author  = {Ba, Jimmy Lei and Kiros, Jamie Ryan and Hinton, Geoffrey E.},
  journal = {arXiv preprint arXiv:1607.06450},
  year    = {2016},
  url     = {https://arxiv.org/abs/1607.06450}
}

@article{irvin2019chexpert,
  title={CheXpert: A Large Chest Radiograph Dataset with Uncertainty Labels and Expert Comparison},
  author={Irvin, Jeremy and Rajpurkar, Pranav and Ko, Michael and Yu, Yifan and Ciurea-Ilinca, Florentina and Chute, Chris and Marklund, Henrik and Haghgoo, Babak and Ball, Robyn and Shpanskaya, Katie and Seekins, Jenna and Mong, David A and Halabi, Safwan S and Sandberg, Jesse K and Jones, Ricky and Larson, David B and Langlotz, Curtis P and Patel, Bhavik N and Lungren, Matthew P and Ng, Andrew Y},
  journal={arXiv preprint arXiv:1901.07031},
  year={2019},
  url={https://arxiv.org/abs/1901.07031}
}

@article{chexpert_irvin2019,
  title={CheXpert: A Large Chest Radiograph Dataset with Uncertainty Labels and Expert Comparison},
  author={Irvin, Jeremy and Rajpurkar, Pranav and Ko, Michael and Yu, Yifan and Ciurea-Ilcus, Silviana and Chute, Chris and Marklund, Henrik and Haghgoo, Behzad and Ball, Robyn and Shpanskaya, Katie and Seekins, James and Mong, David A and Halabi, Safwan S and Sandberg, Jesse K and Jones, Ricky and Larson, David B and Langlotz, Curtis P and Patel, Bhavik N and Lungren, Matthew P and Ng, Andrew Y},
  journal={arXiv preprint arXiv:1901.07031},
  year={2019},
  url={https://arxiv.org/abs/1901.07031}
}

@article{johnson2019mimiccxr,
  title={MIMIC-CXR: A large publicly available database of labeled chest radiographs},
  author={Johnson, Alistair E. W. and Pollard, Tom J. and Berkowitz, Seth and Greenbaum, Nathaniel R. and Lungren, Matthew P. and Deng, Chengyuan and Mark, Roger G. and Horng, Steven},
  journal={arXiv preprint arXiv:1901.07042},
  year={2019},
  url={https://physionet.org/content/mimic-cxr/2.0.0/}
}

@article{irany2024large,
  title={Large Scale Data Analysis with Application to Computational Epidemiology and Network Science},
  author={Irany, Fariba Afrin},
  journal={PhD Dissertation, University of North Texas},
  year={2024},
  url={https://digital.library.unt.edu/ark:/67531/metadc2415893/}
}

@article{santra2023efficient,
  title={Efficient community detection in multilayer networks using boolean compositions},
  author={Santra, Abhishek and Irany, Fariba Afrin and Madduri, Kamesh and Chakravarthy, Sharma and Bhowmick, Sanjukta},
  journal={Frontiers in Big Data},
  volume={6},
  pages={1144793},
  year={2023},
  publisher={Frontiers Media SA}
}

@article{irany2024bias,
  title={Algorithms to Reduce Biases in Disease Rate Estimates Caused by Data Suppression.},
  author={Irany, Fariba Afrin and Al Subhi, Sundos and Flores, Rubenia Borge and Tiwari, Chetan},
  journal={IAENG International Journal of Applied Mathematics},
  volume={54},
  number={5},
  year={2024}
}
